\documentclass[lettersize,journal]{IEEEtran}

\usepackage[colorlinks=true,
            citecolor=steelblue,
            linkcolor=steelblue,
            urlcolor=steelblue]{hyperref}
            
\usepackage{amsmath,amsfonts}
\usepackage{algpseudocode}
\usepackage{algorithm}
\usepackage{array}

\usepackage{textcomp}
\usepackage{stfloats}
\usepackage{url}
\usepackage{verbatim}
\usepackage{graphicx}
\usepackage{cite}

\usepackage{etoolbox}
\patchcmd{\abstract}{\bfseries}{}{}{}
\newcommand{\ours}{\texttt{EPD-Solver}}
\newcommand{\oursplugin}{\texttt{EPD-Plugin}}

\newcommand{\tableCellHeight}{1}
\newcommand{\tabstyle}[1]{
  \setlength{\tabcolsep}{#1}
  \renewcommand{\arraystretch}{\tableCellHeight}
  \centering
  \small
}
\usepackage{amsthm}
\theoremstyle{plain}

\newtheorem{theorem}{Theorem}

\usepackage{amsfonts,bm}
\usepackage{graphicx}	
\usepackage{amsmath}
\usepackage{amssymb}	
\usepackage{booktabs}
\usepackage{times}
\usepackage{microtype}
\usepackage{epsfig}
\usepackage{caption}
\usepackage{float}
\usepackage{placeins}
\usepackage{color, colortbl}
\usepackage{stfloats}
\usepackage{enumitem}
\usepackage{tabularx}
\usepackage{xstring}
\usepackage{multirow}
\usepackage{xspace}
\usepackage{url}
\usepackage{subcaption}
\usepackage{xcolor}
\definecolor{tabhighlight}{HTML}{e5e5e5}
\definecolor{lightCyan}{rgb}{0.925,1,1}
\usepackage[hang,flushmargin]{footmisc}
\usepackage{algorithm}
\usepackage{wrapfig}
\usepackage{cleveref}

\definecolor{steelblue}{RGB}{45, 125, 189}

% ...

\def\onedot{.}

\newcommand{\R}[1]{{%
    \textbf{%
        \ifstrequal{#1}{1}{\textcolor{red}{R#1}}{%
        \ifstrequal{#1}{2}{\textcolor{blue}{R#1}}{%
        \ifstrequal{#1}{3}{\textcolor{magenta}{R#1}}{%
        \ifstrequal{#1}{4}{\textcolor{teal}{R#1}}{%
                           \textcolor{cyan}{R#1}%
        }}}}%
    }%
}}

\def\eg{\emph{e.g}\onedot} 
\def\ie{\emph{i.e}\onedot}

\def\eqref#1{equation~\ref{#1}}
\def\1{\bm{1}}

\def\rd{{\mathrm{d}}}

\def\rvd{{\mathbf{d}}}

\def\rvw{{\mathbf{w}}}
\def\rvx{{\mathbf{x}}}
\def\rvy{{\mathbf{y}}}

\DeclareMathAlphabet{\mathsfit}{\encodingdefault}{\sfdefault}{m}{sl}
\SetMathAlphabet{\mathsfit}{bold}{\encodingdefault}{\sfdefault}{bx}{n}

\def\gN{{\mathcal{N}}}

\def\gT{{\mathcal{T}}}

\def\sR{{\mathbb{R}}}

\begin{document}

\title{Parallel Diffusion Solver via \\Residual Dirichlet Policy Optimization}

\author{
    Ruoyu Wang$^{1}$$^*$\thanks{$^*$Equal contribution. $^\dagger$Corresponding author.} \quad
    Ziyu Li$^{1,2}$$^*$ \quad
    Beier Zhu$^{3}$$^*$ \quad
    Liangyu Yuan$^{1,4}$ \\
    Hanwang Zhang$^{3}$ \quad Xun Yang$^{5}$\quad Xiaojun Chang$^{5}$\quad Chi
     Zhang$^{1}$$^\dagger$\\
    $^1$AGI lab, Westlake University \quad $^2$University of Illinois Urbana-Champaign \\
    $^3$Nanyang Technological University \quad
    $^4$Shanghai Jiao Tong University \\
    $^5$University of Science and Technology of China\\
    {\tt\small  wangruoyu71@westlake.edu.cn}

        % <-this % stops a space
}


\maketitle

\begin{abstract}

\textit{Diffusion models (DMs) have achieved state-of-the-art generative performance but suffer from high sampling latency due to their sequential denoising nature. Existing solver-based acceleration methods often face significant image quality degradation under a low-latency budget, primarily due to accumulated truncation errors arising from the inability to capture high-curvature trajectory segments. In this paper, we propose the Ensemble Parallel Direction solver (dubbed as \ours), a novel ODE solver that mitigates these errors by incorporating multiple parallel gradient evaluations in each step. Motivated by the geometric insight that sampling trajectories are largely confined to a low-dimensional manifold, \ours~leverages the Mean Value Theorem for vector-valued functions to approximate the integral solution more accurately. Importantly, since the additional gradient computations are independent, they can be fully parallelized, preserving low-latency sampling nature. We introduce a two-stage optimization framework. Initially, \ours~optimizes a small set of learnable parameters via a distillation-based approach. We further propose a parameter-efficient Reinforcement Learning (RL) fine-tuning scheme that reformulates the solver as a stochastic Dirichlet policy. Unlike traditional methods that fine-tune the massive backbone, our RL approach operates strictly within the low-dimensional solver space, effectively mitigating reward hacking while enhancing performance in complex text-to-image (T2I) generation tasks. In addition, our method is flexible and can serve as a plugin (\oursplugin) to improve existing ODE samplers. Extensive experiments demonstrate the effectiveness of \ours. On validation benchmarks, at the same latency level of 5 NFE, the distilled \ours~achieves state-of-the-art FID scores of 4.47 on CIFAR-10, 7.97 on FFHQ, 8.17 on ImageNet, and 8.26 on LSUN Bedroom, surpassing existing learning-based solvers by a significant margin. On T2I benchmarks, our RL-tuned \ours~significantly improves human preference scores on both Stable Diffusion v1.5 and SD3-Medium. Notably, it outperforms the official 28-step baseline of SD3-Medium with only 20 steps, effectively bridging the gap between inference efficiency and high-fidelity generation.}

\end{abstract}

% \begin{IEEEkeywords}
% Diffusion Models, Diffusion Sampler, Parallel Sampling, Dirichlet reparameterization
% \end{IEEEkeywords}

\begin{figure}[t]
    \centering
\includegraphics[width=0.45\textwidth]{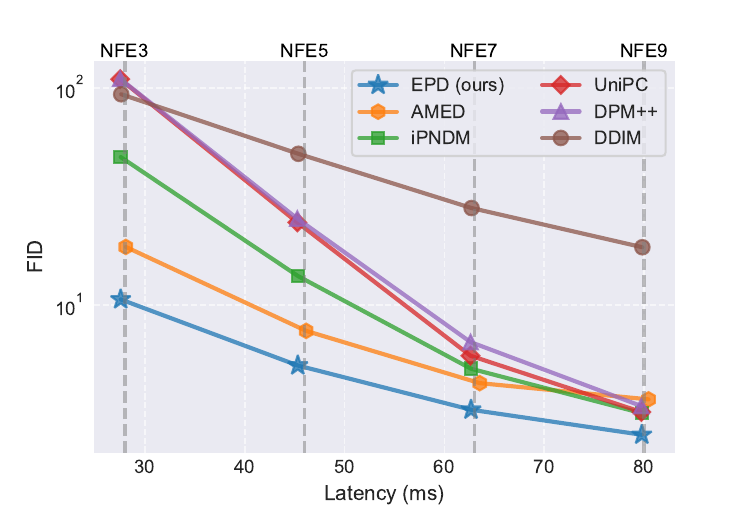}
    \caption{Comparison of various solvers on diffusion models. We compare the FID versus latency (ms) across different NFE settings on a NVIDIA 4090. Our proposed $\ours$ shows superior image quality without increasing latency.} 
    \label{fig:fidvslatency}
        \vspace{-4mm}
\end{figure}
\section{Introduction}
\label{sec:intro}

Diffusion models (DMs)~\cite{sohl2015deep,ho2020denoising,rombach2022high} have emerged as a leading paradigm in generative modeling, delivering state-of-the-art performance across image synthesis~\cite{rombach2022high,saharia2022photorealistic,Lei_2025_CVPR} and video generation~\cite{blattmann2023align,ho2022video,zhou2025streamingdragorientedinteractivevideo,zhao2025realtimemotioncontrollableautoregressivevideo}. These models generate data by iteratively refining noisy inputs through a sequential denoising process. While this mechanism produces high-fidelity outputs, the requirement for multi-step sequential evaluation introduces substantial latency, rendering real-time sampling inefficient.

\begin{figure*}[t]
    \centering
\includegraphics[width=0.95\textwidth]{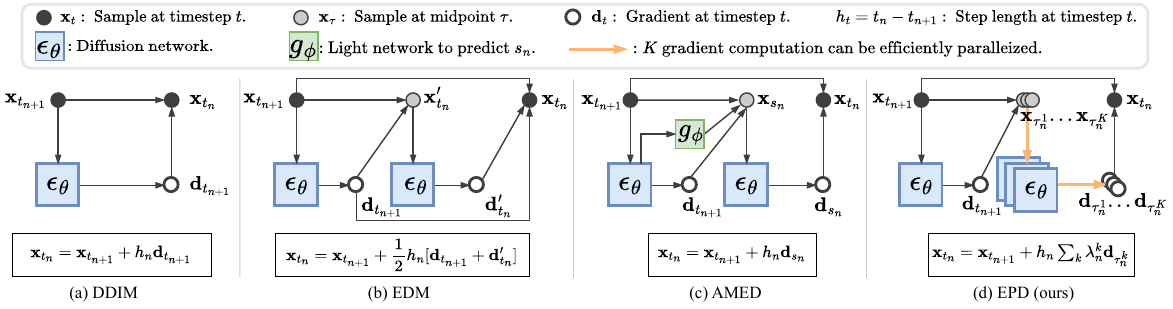}
    \caption{Computation graphs of various ODE solvers. \textbf{(a)} DDIM solver~\cite{songdenoising} (Euler's method) adopts the rectangle rule that uses the gradient at the start point: $\rvd_{t_{n+1}}=\bm{\epsilon}_\theta(\rvx_{t_{n+1}},t_{n+1})$. disclose EDM solver~\cite{karras2022elucidating} (Heun's method) uses the trapezoidal rule that averages the gradients of both the start and the end timesteps, \ie, $\rvd_{t_{n+1}}=\bm{\epsilon}_\theta(\rvx_{t_{n+1}},t_{n+1})$ and $\rvd_{t_{n}}'=\bm{\epsilon}_\theta(\rvx_{t_{n}}',t_{n})$, where $\rvx_{t_n}'$ is the additional evaluation given by Euler's method. \textbf{(c)} AMED solver~\cite{zhou2024fast} optimizes a small network $g_\phi(\cdot)$ to output an intermediate timestep $s_n \in (t_n,t_{n+1})$ to compute the gradient: $\rvd_{s_n}=\bm{\epsilon}_\theta(\rvx_{s_{n}},s_{n})$. Since AMED introduces a network in sequential computation, its latency is slightly higher than that of other solvers, as shown in~\cref{fig:fidvslatency}. \textbf{(d)} Our $\ours$ leverage $K$ parallel gradients to achieve more accurate integral approximation. We optimize $K$ intermediate timesteps $\tau_n^1, \dots, \tau_n^K$, compute their gradients $\rvd_{\tau_{n}^1}, \dots, \rvd_{\tau_{n}^K}$, and combine them via a simplex-weighted sum.} 
    \label{fig:teaser}
    \vspace{-2mm}
\end{figure*}

To mitigate this inefficiency, recent research has focused on accelerating the sampling process through various approaches, such as solver-based, distillation-based and parallelism-based methods. Solver-based methods aim to develop fast numerical solvers to reduce the number of sampling steps~\cite{songdenoising,karras2022elucidating,lu2022dpm,lu2022dpm_plus,liupseudo,zhangfast,zhao2024unipc,zhou2024fast,kim2024distilling,watson2021learning}. However, aggressive step reduction often leads to significant truncation errors, causing quality degradation at low function evaluations (NFEs). Distillation-based methods establish a direct mapping between noise and data~\cite{zhou2025simple,luhman2021knowledge,liuflow,berthelot2023tract,salimansprogressive,meng2023distillation,song2023consistency,luo2023latent,kimconsistency}, achieving extreme acceleration (e.g., one-step generation). Yet, they incur high training costs and lack the flexibility to trade speed for quality. Parallelism-based methods~\cite{shih2023parallel,li2024faster,li2024distrifusion,chenasyncdiff} attempt to trade computation for speed, but this direction remains under-explored for quality enhancement.

In this paper, we seek to combine the strengths of these approaches by investigating solver-based methods under low-latency constraints. We propose the \texttt{E}nsemble \texttt{P}arallel \texttt{D}irection (\texttt{EPD}) solver, a novel approach that leverages extra parallel computation to minimize truncation errors in each ODE step without increasing wall-clock time.
Unlike standard solvers that rely on a single gradient evaluation (e.g., DDIM~\cite{songdenoising}) or sequential multi-point estimates (e.g., EDM~\cite{karras2022elucidating} and AMED~\cite{zhangfast}), our method concurrently evaluates gradients at multiple learned intermediate timesteps within a single integration interval (see~\Cref{fig:teaser}).
By aggregating these parallel gradient estimates via a weighted combination, we achieve a significantly more accurate approximation of the integral direction, which is \textit{theoretically grounded} in the mean value theorem for vector-valued functions~\cite{mcleod1965mean}.
Crucially, since these gradient computations are independent of each other, they can be efficiently parallelized on modern hardware. This allows $\ours$ to enhance sampling fidelity with negligible latency overhead.
 As in \Cref{fig:fidvslatency}, $\ours$ consistently achieves better FID scores than existing ODE solvers at comparable inference latencies on CIFAR~\cite{krizhevsky2009learning}.

We adopt a two-stage optimization framework to decide the intermediate evaluation timesteps and their combining weights.
\textit{(1) In the first stage,} we distill a few-step \texttt{EPD} sampler by optimizing its learnable parameters to approximate the trajectories generated by a high-NFE teacher solver.
However, in the extremely low-step regime, distillation alone is insufficient. It not only struggles to learn an accurate mapping from noise to teacher trajectories~\cite{tong2024learning} but also falls short in aligning with human perceptual preferences.
For large-scale text-to-image (T2I) diffusion models, human preference is better characterized by semantic and perceptual alignment rather than strict trajectory consistency.
Motivated by this, \textit{(2) in the second stage}, we perform \textbf{Residual Dirichlet Policy Optimization} \textbf{(RDPO)}, where the solver is reparameterized as a stochastic policy initialized using the parameters distilled in Stage~1.
This parameterization induces a structured, simplex-constrained policy space that supports stable and efficient optimization of human-aligned rewards via a  PPO~\cite{schulman2017proximalpolicyoptimizationalgorithms} variant.
Moreover, our framework is lightweight and plug-and-play: it optimizes only a few solver parameters while freezing the backbone, thereby reducing tuning cost, improving RL stability, and preserving generation robustness; the resulting solver can also be seamlessly integrated into existing ODE samplers as \texttt{EPD-Plugin}.

We evaluate \ours~across a diverse set of image generation models spanning resolutions from 32 to 1024, including standard unconditional and class-conditional benchmarks (CIFAR-10~\cite{krizhevsky2009learning}, FFHQ~\cite{karras2019style}, ImageNet~\cite{russakovsky2015imagenet}, LSUN Bedroom~\cite{yu2015lsun}) and large-scale T2I models (Stable Diffusion v1.5~\cite{rombach2022high}, SD3-Medium~\cite{sd3}). Empirical results confirm that incorporating parallel gradients significantly reduces truncation errors and consistently outperforms prior learning-based solvers. At 5 NFE, $\ours$ achieves FIDs of 4.47 (CIFAR-10), 7.97 (FFHQ), 8.17 (ImageNet), and 8.26 (LSUN Bedroom), notably surpassing AMED-Solver~\cite{zhou2024fast}, which yields 13.20 on LSUN Bedroom. For T2I models, our residual Dirichlet policy based solver achieves strong alignment with high efficiency: at just 20 NFEs, $\ours$ attains an HPSv2.1 score of 0.2482 and an ImageReward of 0.0121, surpassing 50-step baselines like iPNDM while cutting inference cost by 60\%.
Our contributions are summarized as follows:

\begin{itemize}

    \item We propose $\ours$, a novel ODE solver that exploits parallel gradient evaluations to reduce truncation errors with minimal latency overhead, and introduce $\oursplugin$, a flexible plugin to existing samplers.

    \item  We develop a parameter-efficient RL training scheme that optimizes a residual Dirichlet policy, significantly improving large-scale text-to-image generation.

    \item   We provide both theoretical justification and strong empirical evidence that $\ours$ consistently improves sample quality across diverse models and datasets, outperforming prior solvers under tight latency budgets.

\end{itemize}

Our preliminary results was published in \textbf{ICCV 2025}~\cite{zhu2025distilling}. The source code and checkpoints are available in \texttt{\url{https://github.com/BeierZhu/EPD}}. 
\section{Related Work}
\label{sec:related}

\subsection{Sampling Acceleration Methods}
High latency in the sampling process is a major drawback of DMs compared to other generative models~\cite{goodfellow2014generative,kingma2013auto}. Prior acceleration efforts mainly fall into the following categories:

\noindent\textbf{Distillation-based methods.}
These methods accelerate diffusion models by re-training or fine-tuning the entire DM. One category is trajectory distillation, which trains a student model to imitate the teacher's trajectory with fewer steps~\cite{zhou2025simple}. This process can be achieved through offline distillation~\cite{luhman2021knowledge,liuflow}, which requires constructing a dataset sampled from teacher models, or online distillation, which progressively reduces sampling steps in a multi-stage manner~\cite{berthelot2023tract,salimansprogressive,meng2023distillation}. Another line of research is consistency distillation, where the denoising outputs along the sampling trajectory are enforced to remain consistent~\cite{song2023consistency,luo2023latent,kimconsistency}. Apart from distilling noise-image pairs, distribution matching methods match real and reconstructed samples at the distribution level~\cite{pooledreamfusion,wang2023prolificdreamer,sauer2024adversarial,yin2024one}. Despite significantly enhancing quality, these approaches incur high training costs and require carefully designed training procedures.

\noindent\textbf{Solver-based methods.}  
Beyond fine-tuning DMs, fast ODE solvers have been extensively studied. Training-free methods include Euler's method~\cite{songdenoising}, Heun’s method~\cite{karras2022elucidating}, Taylor expansion-based solvers (DPM-Solver~\cite{lu2022dpm}, DPM-Solver++~\cite{lu2022dpm_plus}), multi-step methods (PNDM~\cite{liupseudo}, iPNDM~\cite{zhangfast}), and predictor-corrector frameworks (UniPC~\cite{zhao2024unipc}). Some solvers require additional training, \eg, AMED-Solver~\cite{zhou2024fast}  
, D-ODE~\cite{kim2024distilling}, and DDSS~\cite{watson2021learning}, AdaSDE~\cite{wang2025adaptivestochasticcoefficientsaccelerating}. Recent work optimizes timestep schedules, with notable studies including LD3~\cite{tong2024learning}, AYS~\cite{sabour2024align}, GITS~\cite{chen2024trajectory}, and DMN~\cite{xue2024accelerating}.  
Though $\ours$ falls into this category, we optimize solver parameters via distillation to achieve high-quality, low-latency generation through parallelism. With minimal learnable parameters, training remains highly efficient.

\noindent\textbf{Parallelism-based methods.}
While promising, parallelism remains an underexplored approach for accelerating diffusion models. ParaDiGMS~\cite{shih2023parallel} leverages Picard iteration for parallel sampling but struggles to maintain consistency with original outputs. Faster Diffusion~\cite{li2024faster} performs decoder computation in parallel by omitting encoder computation at some adjacent timesteps, but this compromises image quality. Distrifusion~\cite{li2024distrifusion} divides high-resolution images into patches and performs parallel inference on each patch. AsyncDiff~\cite{chenasyncdiff} implements model parallelism through asynchronous denoising. Unlike prior methods that focus on reducing latency, our $\ours$ leverages parallel gradients to enhance image quality without incurring notable latency. 

Beyond the above three categories, cache-based acceleration methods~\cite{ma2023deepcache,zhao2024dynamic,zou2024accelerating,Liu_2025_CVPR,liu2024timestep} aim to reduce per-step computational overhead by exploiting temporal redundancy during diffusion sampling.
Specifically, these approaches reuse feature maps or intermediate states from preceding steps to bypass redundant computations.
While effective in reducing wall-clock latency, they do not directly address discretization errors inherent in the numerical integration of the sampling trajectory and are largely orthogonal to our approach.

\subsection{Reinforcement Learning from Human Feedback}
RL-based approaches for aligning pretrained text-to-image (T2I) diffusion models with human preferences can be broadly categorized into supervised-style objectives and policy-gradient–based RL. Supervised-style methods formulate alignment as weighted maximum likelihood or preference-matching, directly shaping the data distribution using scalar rewards or pairwise preference signals without explicit policy gradients \cite{clark2024directlyfinetuningdiffusionmodels,xu2023imagereward,peng2019advantageweightedregressionsimplescalable,lee2023aligningtexttoimagemodelsusing,Wallace_2024_CVPR,dong2023raftrewardrankedfinetuning,zheng2025diffusionnftonlinediffusionreinforcement,zhao2025unsupervised}. In contrast, policy-gradient–based methods treat the T2I model as a stochastic policy and explicitly optimize expected rewards via RL updates, typically in a PPO-style framework \cite{black2024trainingdiffusionmodelsreinforcement,NEURIPS2023_fc65fab8,gupta2025simpleeffectivereinforcementlearning,Miao_2024_CVPR,zhao2025scoreactionfinetuningdiffusion,liu2025flowgrpotrainingflowmatching,li2025mixgrpounlockingflowbasedgrpo,xue2025dancegrpounleashinggrpovisual}. Different from existing RL-based alignment methods that optimize the DM itself, we perform RL at the solver level by learning a residual Dirichlet policy around a distilled base solver, enabling both parameter-efficient and robust preference alignment.

\section{Method}
\label{sec:method}
\begin{figure*}[t]
    \centering
\includegraphics[width=0.9\textwidth]{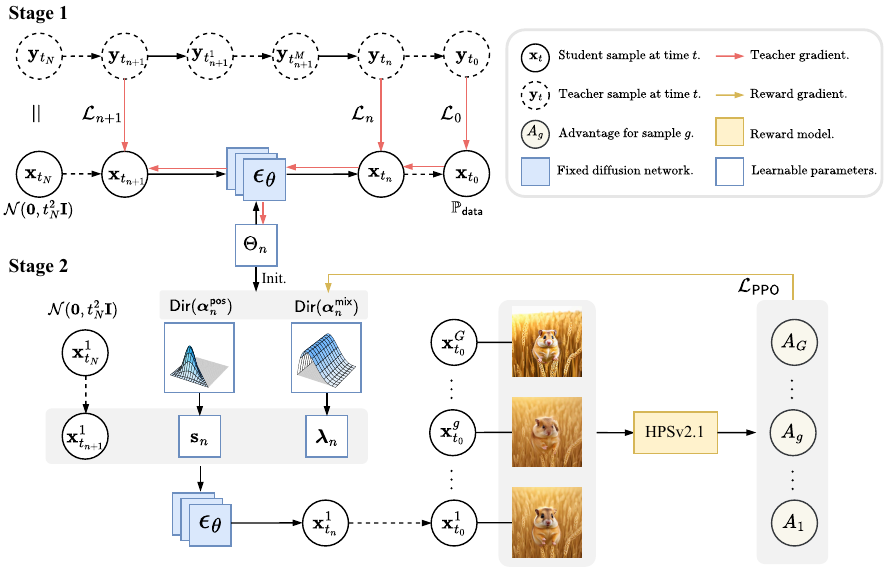}
    \caption{\textbf{Stage 1: Distillation-Based Parameter Optimization (Top).} We optimize the learnable solver parameters $\Theta_n$ by minimizing the trajectory reconstruction error  against a high-precision teacher solver (e.g., DPM-Solver-2), providing a robust initialization for stage 2.
\textbf{Stage 2: Residual Dirichlet Policy Optimization (Bottom).} To align generation with human preferences, we reformulate the solver as a stochastic policy parameterized by Dirichlet distributions (defined by $\bm{\alpha}^\mathsf{pos}_n$ and $\bm{\alpha}^\mathsf{mix}_n$). By sampling multiple parallel trajectories in the low-dimensional solver space and evaluating them with a reward model (\eg, HPSv2.1), we optimize the policy using PPO with a Reward-Leave-One-Out (RLOO) baseline.}
    \label{fig:pipeline}
        \vspace{-1mm}
\end{figure*}

 We begin by reviewing the diffusion sampling process (\Cref{sec:background}) and motivating our $\ours$ approach, which is grounded in the fact that leveraging multiple gradients can effectively reduce truncation error (\Cref{sec:method_motivation}). We then introduce our method, illustrated in \Cref{fig:pipeline}, which consists of two stages: (1) a distillation-based initialization that captures the curvature of the sampling trajectory (\Cref{sec:stage1}), and (2) a parameter-efficient residual Dirchlet policy optimization stage that further fine-tunes the sampler to align with human preferences (\Cref{sec:stage2}).

\subsection{Background}
\label{sec:background}
 Diffusion models gradually inject noise into data via a forward noising process and generate samples by learning a reversed denoising process, initialized with Gaussian noise. Let $\rvx \sim p_\text{data}(\rvx)$ denote the $d$-dimensional data and $p(\rvx;\sigma)$ the data distribution with Gaussian noise of variance $\sigma^2$ injected. The forward process is controlled by a noise schedule defined by the time scaling $s(t)$ and the noise level $\sigma(t)$ at time $t$.
In particular, $\rvx=s(t)\hat{\rvx}_t$, where $\hat{\rvx}_t \sim p(\rvx;\sigma(t))$. Such forward process can be formulated by a SDE~\cite{karras2022elucidating}:
\begin{equation}\label{eq:sde}
    \rd\rvx = \frac{\dot{s}(t)}{s(t)}\rvx + s(t)\sqrt{2\sigma(t)\dot{\sigma}(t)}\rd \rvw_t,
\end{equation}
where $\rvw \in \sR^d$ denotes Wiener process. In this paper, we adopt the framework of         EDM~\cite{karras2022elucidating} by setting $\sigma(t)=t$ and $s(t)=1$.
Generation is then performed with the reverse of~\cref{eq:sde}. Notably, there exists the probability flow ODE: 
\begin{equation}\label{eq:full-ode}
\rd\rvx =  - t\nabla_\rvx \log p(\rvx;t)\rd t 
\end{equation}
We learn a parameterized network $\bm{\epsilon}_\theta(\rvx,t)$ to predict the Gaussian noise added to $\rvx$ at time $t$. The network satisfies: $\bm{\epsilon}_\theta(\rvx,t)=-t\nabla_\rvx\log p(\rvx;t)$ and \Cref{eq:full-ode} simplifies to: 
\begin{equation}\label{eq:ode-simple}
   \rd\rvx=\bm{\epsilon}_\theta(\rvx,t)\rd t
\end{equation}
The noise-prediction model $\bm{\epsilon}_\theta(\rvx,t)$ is trained by minimizing the $\ell_2^2$ loss with a weighting function $\lambda(t)$~\cite{karras2022elucidating,song2021scorebased}:
\begin{equation}
    \mathcal{L}_t(\theta)=\lambda(t)\mathbb{E}_{\rvx\sim p_\text{data},\bm{\epsilon}\sim \mathcal{N}(0,\mathbf{I})}\|\bm{\epsilon}_\theta(\rvx,t)-\bm{\epsilon}\|_2^2
\end{equation}
Given a time schedule $\gT=\{t_0=t_{\min},\cdots,t_N=t_{\max}\}$, data generation involves starting from random noise $\rvx_{t_N} \sim \gN(\bm{0}, t_{\max}^2\mathbf{I})$, then iteratively solving \Cref{eq:ode-simple} to compute the sequence $\{\rvx_{t_{N-1}},...,\rvx_{t_0}\}$. 

\subsection{Motivation and Analysis}
\label{sec:method_motivation}
The solution of \Cref{eq:ode-simple}
at time $t_n$ can be exactly computed in the integral form:
\begin{equation}
    \rvx_{t_n} = \rvx_{t_{n+1}} + \int_{t_{n+1}}^{t_n} \bm{\epsilon}_\theta(\rvx_t,t)\rd t
\end{equation}
Various ODE solvers have been proposed to approximate the integral. At a high level, these solvers leverage one or several points to compute gradients, which are then used to estimate the integral. Let $I$ denote the integral $I=\int_{t_{n+1}}^{t_n}\bm{\epsilon}_\theta(\rvx_t,t)\rd t$ and $h_n$ denote the step length $h_n=t_n-t_{n+1}$. For instance, DDIM~\cite{songdenoising} (Euler's method) adopts the rectangle rule that uses the gradient at the start point:
\begin{equation}
I \approx h_n \underbrace{\bm{\epsilon}_\theta(\rvx_{t_{n+1}},t_{n+1})}_{\text{start point grad.}}. 
\end{equation}
EDM~\cite{karras2022elucidating} considers the trapezoidal rule that averages the gradients of both the start and end points.
\begin{equation}
I \approx \frac{1}{2} h_n \{\underbrace{\bm{\epsilon}_\theta(\rvx_{t_{n+1}},t_{n+1})}_{\text{start point grad.}}+\underbrace{\bm{\epsilon}_\theta(\rvx'_{t_{n}},t_{n})}_{\text{end point grad.}}\}, 
\end{equation}
where $\rvx_{t_n}'$ is the additional evaluation point given by Euler's method, \ie,  $\rvx_{t_n}'=\rvx_{t_{n+1}}+h_n\bm{\epsilon}_\theta(\rvx_{t_{n+1}},t_{n+1})$. AMED-Solver~\cite{zhou2024fast} optimizes a small network to output an intermediate timestep $s_n \in (t_n,t_{n+1})$ to compute the gradient:
\begin{equation}
I \approx  h_n \underbrace{\bm{\epsilon}_\theta(\rvx_{s_n},s_n)}_{\text{midpoint grad.}}, 
\end{equation}
where $\rvx_{s_n}=\rvx_{t_{n+1}}+(s_n-t_{n+1})\bm{\epsilon}_\theta(\rvx_{t_{n+1}},t_{n+1})$. The computational graphs of DDIM, EDM, and AMED-Solver, illustrating their respective integral approximation processes, are shown in~\Cref{fig:teaser}.

Compared to DDIM, EDM and AMED introduce an additional timestep for gradient computation ($t_n$ and $s_n$), leading to improved integral estimation. The key motivation of our method is to leverage multiple timesteps to reduce the truncation errors. Furthermore, since the computations of additional gradients are independent, they can be \textit{efficiently parallelized without increasing inference latency}. In this work, we propose the \texttt{E}nsemble \texttt{P}arallel \texttt{D}irection (\texttt{EPD}) solver, which refines the integral estimation by incorporating multiple intermediate timesteps. Formally, the integral is approximated as:
\begin{equation}\label{eq:epd1}
I \approx h_n \underbrace{\sum_{k=1}^K  \lambda^k_n\bm{\epsilon}_\theta(\rvx_{\tau^k_n},\tau^k_n)}_{\text{ensemble parallel grads.}},
\end{equation}
where $\tau^k_n \in (t_n,t_{n+1})$ are the intermediate timesteps, and the weights form a simplex combination satisfying $\lambda^k_n \geq 0$ and $\sum_{k=1}^K \lambda^k_n = 1$. The state at each intermediate timestep $\tau^k_n$ is computed using Euler’s method as: $\rvx_{\tau^k_n}=\rvx_{t_{n+1}}+(\tau_k-t_{n+1})\bm{\epsilon}_\theta(\rvx_{t_{n+1}},t_{n+1})$. Each gradient computation $\bm{\epsilon}_\theta(\rvx_{\tau^k_n},\tau^k_n)$ is fully parallelizable, preserving efficiency without increasing inference latency. 
In fact, the use of gradients estimated at multiple timesteps for improved integral approximation can be theoretically justified by the following mean value theorem for vector-valued functions.
\begin{theorem}\label{theorem:1}
(\cite{mcleod1965mean})
    When $f$ has values in an $n$-dimensional vector space and is continuous on the closed interval $[a,b]$ and differentiable on the open interval $(a,b)$, we have
    \begin{equation}
        f(b)-f(a)=(b-a)\sum_{k=1}^n \lambda_k f'(c_k),
    \end{equation}
    for some $c_k\in (a,b), \lambda_k \geq 0$, and $\sum_{k=1}^n\lambda_k=1$.
\end{theorem}
In the context of denoising process, the  function outputs an $d$-dimensional vector as $\rvx \in \mathbb{R}^d$. 
According to~\Cref{theorem:1}, the exact integral of $\bm{\epsilon}_\theta(\rvx_t,t)$ over the interval $[t_n,t_{n+1}]$ can be expressed as a simplex-weighted combination of  gradients evaluated at $d$ intermediate points, scaled by the interval length $h_n=t_n-t_{n+1}$, as formulated in~\Cref{eq:epd1}.

\begin{figure}[t]
    \centering
    \includegraphics[width=0.4\textwidth]{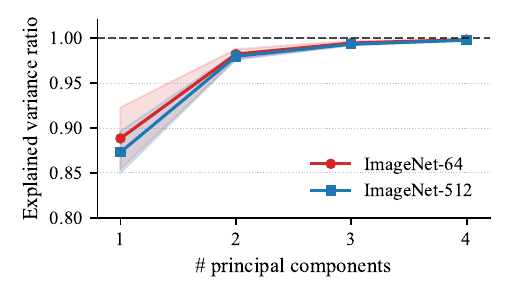}%
    \label{fig:sub_variance}%

\caption{Cumulative explained variance ratio of sampling trajectories using DMs from EDM2~\cite{Karras2024edm2}. We analyze the trajectory's orthogonal complement, \ie, the residuals after removing the  linear component connecting $\mathbf{x}_{t_T}$ and $\mathbf{x}_{t_0}$.  The rapid saturation at the two principle components (capturing $>97\%$ of the residual variance) indicates that the trajectory occurs almost within a single 2D plane.
}
    \label{fig:trajectory_geometry_analysis}
\end{figure}

\subsubsection{Discussion with multi-step solvers}
While multi-step solvers~\cite{lu2022dpm_plus,zhao2024unipc,liupseudo,zhangfast} also combine multiple gradients to approximate the integral, they fundamentally differ in where these gradients are evaluated.
Specifically, multi-step methods rely on Taylor expansion or polynomial extrapolation to linearly combine \emph{historical} gradients evaluated at previous time steps, \ie, outside the current integration interval.
In contrast, \Cref{theorem:1} implies that the exact integral over an interval $[a,b]$ admits a representation as a convex combination of gradients evaluated at points strictly within $(a,b)$.
Motivated by this theoretical result, our method explicitly constructs a simplex-weighted combination of multiple gradients evaluated \emph{within the current time interval}, leading to a more faithful approximation of the integral.

\subsubsection{Discussion with AMED-Solver}
AMED-Solver~\cite{zhou2024fast} estimates the update direction using a \textit{single} intermediate timestep.
However, according to \Cref{theorem:1}, the integral of a vector-valued function \textbf{cannot}, in general, be exactly represented by the derivative at a single timestep; instead, it admits a convex combination of derivatives evaluated at multiple timesteps.
A single timestep may only suffice when the underlying trajectory is effectively one-dimensional.
In the sequel, we empirically analyze the geometric properties of diffusion sampling trajectories and find that they are nearly confined to a two-dimensional manifold, violating this condition.

In \Cref{fig:trajectory_geometry_analysis}, we analyze diffusion sampling trajectories using the models from~\cite{Karras2024edm2}.
Despite the high dimensionality of the ambient space, the cumulative explained variance ratio shows that over 97\% of the residual variance is captured by the first two principal components, indicating that the non-linear trajectory is effectively confined to a two-dimensional manifold.
As a result, single-step or single-intermediate-point methods such as AMED-Solver are generally insufficient to characterize the local curvature within this plane using a single direction.
This motivates the use of multiple intermediate gradients to span the underlying subspace and achieve a more accurate integral approximation.

\subsection{Stage 1: Distillation-Based Parameter Optimization}\label{sec:stage1}

The first stage aims to obtain a strong and stable initialization for RL-based training (Stage 2). We achieve this via a distillation-based optimization that matches low-step student trajectories to high-fidelity teacher trajectories.

\cite{ningelucidating,li2024alleviating} identify exposure bias—\ie, the mismatch between training and sampling inputs—as a key factor contributing to error accumulation and sampling drift. To mitigate this, they propose scaling the network output and shifting the timestep, respectively. Inspired by these insights, we introduce two learnable parameters, $o_n$ and $\delta_n^k$, to perturb the scale of network output’s and the timestep. Our $\ours$ follows the update rule: 
\begin{equation}\label{eq:epdsolverfinal}
\rvx_{t_n} = \rvx_{t_{n+1}} + (1+o_n)h_n\sum_{k=1}^K  \lambda^k_n\bm{\epsilon}_\theta(\rvx_{\tau^k_n},\tau^k_n+\delta^k_n)
\end{equation}
We define the parameters at step $n$ as ${\Theta}_n = \{\tau^k_n, \lambda_n^k, \delta^k_n, o_n \}_{k=1}^K$ and denote the complete set of parameters for an $N$-step sampling process as ${\Theta}_{1:N}$. Consequently, the total number of parameters is given by $N(1+3K)$.

To determine ${\Theta}_{1:N}$, we employ a distillation-based optimization process. 
Specifically, given a student time schedule with $N$ steps $\mathcal{T}_{\mathsf{stu}}=\{t_0=t_{\min},...,t_N=t_{\max}\}$, we insert $M$ intermediate steps between $t_n$ and $t_{n+1}$, \ie, $\mathcal{T}_{\mathsf{tea}}=\{t_0,...,t_n,t_n^1,...,t_n^M,t_{n+1},..,t_N\}$, to yield a more accurate teacher trajectories.
The training process starts with generating teacher trajectories by any ODE solver (\eg, DPM-Solver) and store the reference states as $\{\rvy_{t_n}\}_{n=0}^N$. Afterward, we sample student trajectory with the same initial noise $\rvy_{t_N}$, and optimize the parameters $\{\Theta_n\}_{n=1}^N$ to obtain the student trajectory $\{\rvx_{t_n}\}_{n=0}^N$ that aligns the teacher trajectory w.r.t some distance measurement $\text{dist}(\cdot,\cdot)$. For noisy states $\{\rvx_{t_n}\}_{n=1}^N$, we use the squared $\ell_2$ distance as $\text{dist}(\cdot,\cdot)$. For a generated sample $\rvx_{t_0}$, we compute the squared $\ell_2$ distance in the feature space of the last layer of an ImageNet-pretrained Inception network~\cite{szegedy2015going}.
In particular, to improve the alignment between $\rvx_{t_n}$ and $\rvy_{t_n}$, since the value of $\rvx_{t_n}$ is dependent of the parameters $\Theta_N$ to $\Theta_n$, we aim to optimize them by minimizing
\begin{equation}\label{eq:loss}
    \mathcal{L}_n(\Theta_{N:n})=\mathsf{dist}(\rvx_{t_n}, \rvy_{t_n}).
\end{equation}
In one training loop, we require $N$ backpropagation. The entire training algorithm is listed in~\cref{algo:train} and the inference procedure is provided in~\cref{algo:sample}.
By default, we adopt the analytical first step (AFS) trick~\cite{dockhorn2022genie} in the first step to save one NFE by simply using $\rvx_{t_N}$ as direction. 
\begin{algorithm}[t]
\caption{Stage 1: Distillation-based optimization}
\label{algo:train}
\begin{algorithmic}[1]
\State \textbf{Given:} Time schedules $\mathcal{T}_\mathsf{stu}$ and $\mathcal{T}_\mathsf{tea}$, teacher solver $\mathcal{S}$.
\State \textbf{Return:} $\Theta_{1:N}$, where $\Theta_n=\{\tau_n^k,\lambda_n^k,\delta_n^k,o_n\}_{k=1}^K$
\Repeat 
\State Initialize $\rvx_{t_N}=\rvy_{t_N}\sim \mathcal{N}(\mathbf{0},t^2_N\mathbf{I})$
\State Sample a teacher trajectory $\{\rvy_{t_n}\}_{n=1}^N$ via $\mathcal{S}$
\For{$n = N-1$ \textbf{to} $0$}
    \State Compute $\rvx_{\tau_n^k}$ for all $k$ via an Euler step from $\rvx_{t_{n+1}}$
    \State Compute $\rvx_{t_n}$ using~\Cref{eq:epdsolverfinal}
    \State Update $\Theta_{N:n}$ via $\min\mathcal{L}_n(\Theta_{N:n})$ (\Cref{eq:loss})
\EndFor

\Until{converge}
\end{algorithmic}
\end{algorithm}
\begin{algorithm}[t]
\caption{Stage 2: Residual Dirichlet Policy Optimization}\label{algo:rl_tuning}
\begin{algorithmic}[1]
\State \textbf{Given:} Distilled parameters $\Theta_{1:N}$, reward model $R(\cdot)$, prompts dataset $\mathcal{D}$, policy parameters $\bar{\bm{\alpha}}^{\mathsf{mix}}_n, \bar{\bm{\alpha}}^{\mathsf{pos}}_n$ initialized via \Cref{eq:lambda-init}
\Repeat
    \State Sample batch of prompts $\mathcal{C} \sim \mathcal{D}$
    \For{each prompt $c \in \mathcal{C}$}
        \For{$g = 1$ \textbf{to} $G$}
            \State Sample solver parameter $\mathbf{s}_{1:N}^{g} \sim \mathsf{Dir}(\cdot|\bm{\alpha}^\mathsf{pos}_{1:N})$
            \State Sample solver parameter $\bm{\lambda}_{1:N}^{g} \sim \mathsf{Dir}(\cdot|\bm{\alpha}^\mathsf{mix}_{1:N})$
            \State Override $\Theta_{1:N}$ with $\mathbf{s}_{1:N}^{g},  \bm{\lambda}_{1:N}^{g}$ to obtain $\Theta_{1:N}^{g}$
            \State Generate image $\rvx_{t_0}^{g}$ via $\ours(\Theta_{1:N}^{g})$  
            \State Compute reward $r_{g} \leftarrow R(\rvx_{t_0}^{g}, c)$
        \EndFor

        \For{$g = 1$ \textbf{to} $G$}
            \State $b_{g} \leftarrow \frac{1}{G-1} \sum_{g' \neq g} r_{g}$,\quad $A_{g} \leftarrow r_{g} - b_{g}$
        \EndFor
    \EndFor
    \State Update $\Delta_{1:N}^\mathsf{pos},\Delta_{1:N}^\mathsf{mix}$ via \Cref{eq:ppo,eq:kl}
    \State $\bm{\alpha}_{1:N}^\mathsf{pos} \leftarrow \bar{\bm{\alpha}}_{1:N}^\mathsf{pos}\exp{\Delta_{1:N}^\mathsf{pos}}, \quad \bm{\alpha}_{1:N}^\mathsf{pos} \leftarrow \bar{\bm{\alpha}}_{1:N}^\mathsf{mix}\exp{\Delta_{1:N}^\mathsf{mix}} $
    
\Until{converged}
\end{algorithmic}
\end{algorithm}
\begin{algorithm}[t]
\caption{$\ours$ sampling}\label{algo:sample}
\begin{algorithmic}[1]
\State \textbf{Given:} Time schedule $\mathcal{T}_\mathsf{stu}$, learned parameters $\Theta_{1:N}$.
\State \textbf{Optional:} Compute modes of $\mathsf{Dir}(\cdot\mid \bm{\alpha}_{1:N}^\mathsf{pos})$ and $(\cdot\mid \bm{\alpha}_{1:N}^\mathsf{pos})$, Override $\Theta_{1:N}$ with $\mathbf{s}_{1:N}^\mathsf{mode},  \bm{\lambda}_{1:N}^\mathsf{mode}$
\State \textbf{Return:} $\rvx_{t_0}$
\State Initialize $\rvx_{t_N}\sim \mathcal{N}(\mathbf{0},t^2_N\mathbf{I})$
\For{$n = N-1$ \textbf{to} $0$}
    \State Compute $\rvx_{\tau_n^k}$ for all $k$ via an Euler step from $\rvx_{t_{n+1}}$
    \State $I\leftarrow  (1+o_n)h_n\sum_{k=1}^K  \lambda_n^k\bm{\epsilon}_\theta(\rvx_{\tau_n^k},\tau^k_n+\delta^k_n)$
    \Statex {\color{steelblue} \Comment{implement parallelism for accelerating}}
    \State $\rvx_{t_n} \leftarrow  \rvx_{t_{n+1}} + I$
\EndFor
\end{algorithmic}
\end{algorithm}

\vspace{0.5em}
\noindent\textbf{\oursplugin~to existing solvers.} 
\ours~can be applied to existing solvers to further enhance diffusion sampling. The key idea is to replace their original gradient estimation with multiple parallel branches. As a representative case, we demonstrate this using the multi-step iPNDM sampler~\cite{liupseudo,zhangfast}.
We refer to the modified solver as \oursplugin. Due to space limitations, a detailed description is deferred to \cref{sec:intro-plugin}.

\subsection{Stage 2: Residual Dirichlet Policy Optimization}
\label{sec:stage2}

While existing learnable solvers predominantly rely on trajectory-preserving distillation to compress high-NFE ODE solvers~\cite{wang2025adaptivestochasticcoefficientsaccelerating,zhou2024fast,tong2024learning,chen2024trajectory}, this paradigm often leads to noticeable degradation in the few-step regime.
This limitation arises because, for extremely low-step solvers, learning an exact one-to-one mapping from noise to teacher trajectories is inherently challenging~\cite{tong2024learning}, as fitting errors tend to accumulate along the shortened sampling path.
Fortunately, for large-scale text-to-image diffusion models, human perception does not require strict numerical consistency with the teacher, but rather semantic and perceptual alignment.
Motivated by these observations, we adopt a parameter-efficient reinforcement learning stage to move beyond exact trajectory matching and align the sampling behavior with human preferences.

\vspace{0.5em}
\noindent\textbf{Action space.} 
While the distillation stage offers the full parameter set $\Theta_n=\{\tau_n^k,\lambda_n^k,\delta_n^k,o_n\}_{k=1}^K$, we adopt a simplified and more stable parameterization for the RL fine-tuning stage.
Specifically, we freeze the auxiliary correction terms $o_n$ and $\delta_n^k$.
Although our framework allows optimizing all parameters, jointly learning $o_n$ and $\delta_n^k$  under policy gradients often leads to unstable training due to high-variance reward signals and increases memory consumption (see \Cref{fig:scale_factor}).

\vspace{0.5em}
\noindent\textbf{Dirichlet reparameterization.}
To optimize solver parameters with RL, we require a stochastic policy that can sample valid intermediate timesteps and combination coefficients, together with tractable log-probabilities for likelihood-ratio and KL regularization.
 We interpret the $\ours$ as a policy that, at each interval $(t_{n+1}, t_n)$, decides where to place intermediate evaluations ($K$  \textit{positions} $\{\tau_n^k\}_{k=1}^K$) and how to combine the corresponding gradients ($K$ \textit{mixture} coefficients $\{\lambda_n^k\}_{k=1}^K$).

For \textit{positions}, we introduce a $(K+1)$-dimensional segment vector $\mathbf{s}_n = [s_n^1,\cdots, s_n^{K+1}]^\top \in \mathbb{R}^{K+1}$, whose cumulative sums define ordered intermediate timesteps within $(t_{n+1}, t_n)$:
\begin{equation}
   \tau_n^k = t_{n+1} + r_n^k h_n, \quad r_n^k = \sum_{j=1}^{k} s_n^j, \quad k\in [1,K],
\end{equation} 
where $h_n = t_n - t_{n+1}$. These segments are non-negative and sum to one, \ie, $s_n^j \ge 0$ and $\sum_{j=1}^{K+1} s_n^j = 1$. For \textit{mixture} coefficients, we introduce a $K$-dimensional vector $\bm{\lambda}_n=[\lambda_n^1,\cdots,\lambda_n^K]^\top \in \mathbb{R}^{K}$, which also lies on the simplex with $\lambda_n^j \ge 0$ and $\sum_{j=1}^K \lambda_n^j = 1$.

Since both $\mathbf{s}_n$ and $\bm{\lambda}_n$ lie on simplices, we parameterize them with Dirichlet distributions:
\begin{equation}
    \mathbf{s}_n \sim \mathsf{Dir}(\cdot \mid \bm{\alpha}^{\mathsf{pos}}_n), \quad
    \bm{\lambda}_n \sim \mathsf{Dir}(\cdot \mid \bm{\alpha}^{\mathsf{mix}}_n),
\end{equation}
where a $D$-dimensional Dirichlet distribution with concentration parameters 
$\bm{\alpha}\in\mathbb{R}^D_+$ has density
\begin{equation}
\mathsf{Dir}(\rvx\mid\bm{\alpha})=\frac{1}{B(\bm{\alpha})}\prod_{i=1}^{D} x_i^{\alpha_i - 1},
\  \text{s.t. } \sum_{i=1}^D x_i = 1,\ x_i \ge 0.
\end{equation}
Here, $B(\bm{\alpha})$ is the multivariate Beta function, serving as the normalization constant:
\begin{equation}
    B(\bm{\alpha}) = \frac{\prod_{i=1}^D \Gamma(\alpha_i)}{\Gamma\left(\alpha_0\right)},\quad \text{where} \quad \alpha_0 = \sum_{i=1}^D \alpha_i
\end{equation}
where $\Gamma(\cdot)$ denotes the Gamma function. 
 Importantly, Dirichlet distributions are exactly supported on the simplex and admit closed-form log-densities and KL divergences, which enables stable likelihood-based optimization, and KL control in the RL updates (see \Cref{eq:kl}).

\vspace{0.5em}
\noindent\textbf{Residual Dirichlet Policy around a distilled base solver.}
To make RL both data-efficient and stable, we do not learn $\bm{\alpha}^{\mathsf{pos}}_n$ and $\bm{\alpha}^{\mathsf{mix}}_n$ from scratch. Instead, we leverage the solver distilled in Stage~1 and convert the distilled segments and weights into \emph{base concentration parameters} 
$\bm{\bar{\alpha}}^{\mathsf{pos}}_n$ and $\bm{\bar{\alpha}}^{\mathsf{mix}}_n$, 
whose Dirichlet \textit{modes} recover the distilled solver.
Recall that for a Dirichlet distribution $\mathsf{Dir}(\cdot\mid\bm{\alpha})$ with $\alpha_i>1$, the mode is given by
\begin{equation}\label{eq:mode}
    \operatorname{mode}(\mathsf{Dir}(\cdot\mid\bar{\bm{\alpha}})_i) = \frac{\alpha_i - 1}{\alpha_0 - D}.
\end{equation}

We illustrate the initialization using the \textit{mixture} coefficients as an example.
Let $\bar{\bm{\lambda}}_n$ denote the simplex-valued coefficients obtained in Stage~1.
We initialize the base concentration as
\begin{equation}\label{eq:lambda-init}
    \bar{\bm{\alpha}}^{\mathsf{mix}}_n = \mathbf{1} + \kappa \bar{\bm{\lambda}}_n,
\end{equation}
where $\mathbf{1}$ denotes a $K$-dimensional all-one vector, and $\kappa>0$ is a hyperparameter controlling the global concentration scale. Larger values of $\kappa$ encourage greater exploration; see \Cref{fig:ablation_alpha} for an ablation study.
Substituting \Cref{eq:lambda-init} into \Cref{eq:mode} yields
$\operatorname{mode}(\mathsf{Dir}(\cdot\mid\bar{\bm{\alpha}}^{\mathsf{mix}}_n))=\bar{\bm{\lambda}}_n$.
The base concentrations for the segment variables $\mathbf{s}_n$ are initialized in a  similar manner.

Our policy parameterization outputs \emph{residuals in log-concentration space}:
\begin{align*}
    \log \bm{\alpha}^{\mathsf{pos}}_n = \log \bar{\bm{\alpha}}^{\mathsf{pos}}_n + \Delta^{\mathsf{pos}}_n, \quad
    \log \bm{\alpha}^{\mathsf{mix}}_n = \log \bar{\bm{\alpha}}^{\mathsf{mix}}_n + \Delta^{\mathsf{mix}}_n,
\end{align*}
where $\Delta^{\mathsf{pos}}_n$ and $\Delta^{\mathsf{mix}}_n$ are learnable residuals, which are initialized as zeros, so the resulting Dirichlet policy has the same mode as the distilled solver. Our residual Dirichlet policy is parameter-efficient, easy to optimize, and easy to interpret.

\vspace{0.5em}
\noindent\textbf{Policy optimization.}
In this paper, we adopt a lightweight PPO \cite{schulman2017proximalpolicyoptimizationalgorithms} variant with reward leave-one-out (RLOO) advantages.
Specifically, for each text prompt, we draw $G$ solvers from the policy and generate $G$ candidate images. Let $r_{g}$ denote the scalar reward of the $g$-th candidate, as computed by HPS v2.1 \cite{hpsv2}. The RLOO reward baseline for $r_{g}$ is defined as the average reward of the \emph{other} candidates from the same prompt:
\begin{equation}
    {b}_{g}
    =
    \frac{1}{G-1}
    \sum_{g' \neq g} r_{g'},
\end{equation}
and the corresponding advantage is $A_g =  r_g - b_g$. 
%We further normalize advantages over the batch to have zero mean and unit variance to stabilize optimization.

We adopt the clipped PPO objective to update the policy. 
With advantages $A$ and likelihood ratio
$
    \mathbf{r}=\exp(\log \pi_\theta-\log \pi_{\mathsf{old}}),
$
the surrogate loss is
\begin{equation}
\label{eq:ppo}
    \mathcal{L}_{\mathsf{PPO}}
    =
    - \mathbb{E}\big[
        \min\big(
            \mathbf{r} A,\;
            \operatorname{clip}(\rho, 1-\epsilon, 1+\epsilon)\, A
        \big)
    \big],
\end{equation}
where $\epsilon$ is the clipping range. In addition, we regularize the policy towards the distilled base solver. Thanks to the closed-form KL divergence of Dirichlet distributions, this regularization is analytically tractable:
\begin{equation}\label{eq:kl}
\mathrm{KL}\!\left(\mathsf{Dir}(\bm{\alpha}) \,\|\, \mathsf{Dir}(\bm{\beta})\right)
=
\log\frac{B(\bm{\beta})}{B(\bm{\alpha})}
+ 
\sum_{i=1}^D (\alpha_i-\beta_i)
\big[\psi(\alpha_i)-\psi(\alpha_0)\big] 
\end{equation}
where $\psi(\cdot)$ is the digamma function.

The training pipeline is provided in \Cref{algo:sample}.
Since the DM is frozen and the policy operates on only a small number of Dirichlet parameters, our RL procedure incurs minimal computational overhead while yielding consistent improvements in T2I reward metrics.

\vspace{0.5em}
\noindent\textbf{Inference.} During inference, we do not sample solvers from the residual Dirichlet policy; instead, we use its mode to deterministically instantiate $\bm{\lambda}_n$ and $\mathbf{s}_n$, as defined in \Cref{eq:mode}.

\begin{figure*}[t]
    \centering
\includegraphics[width=0.8\textwidth]{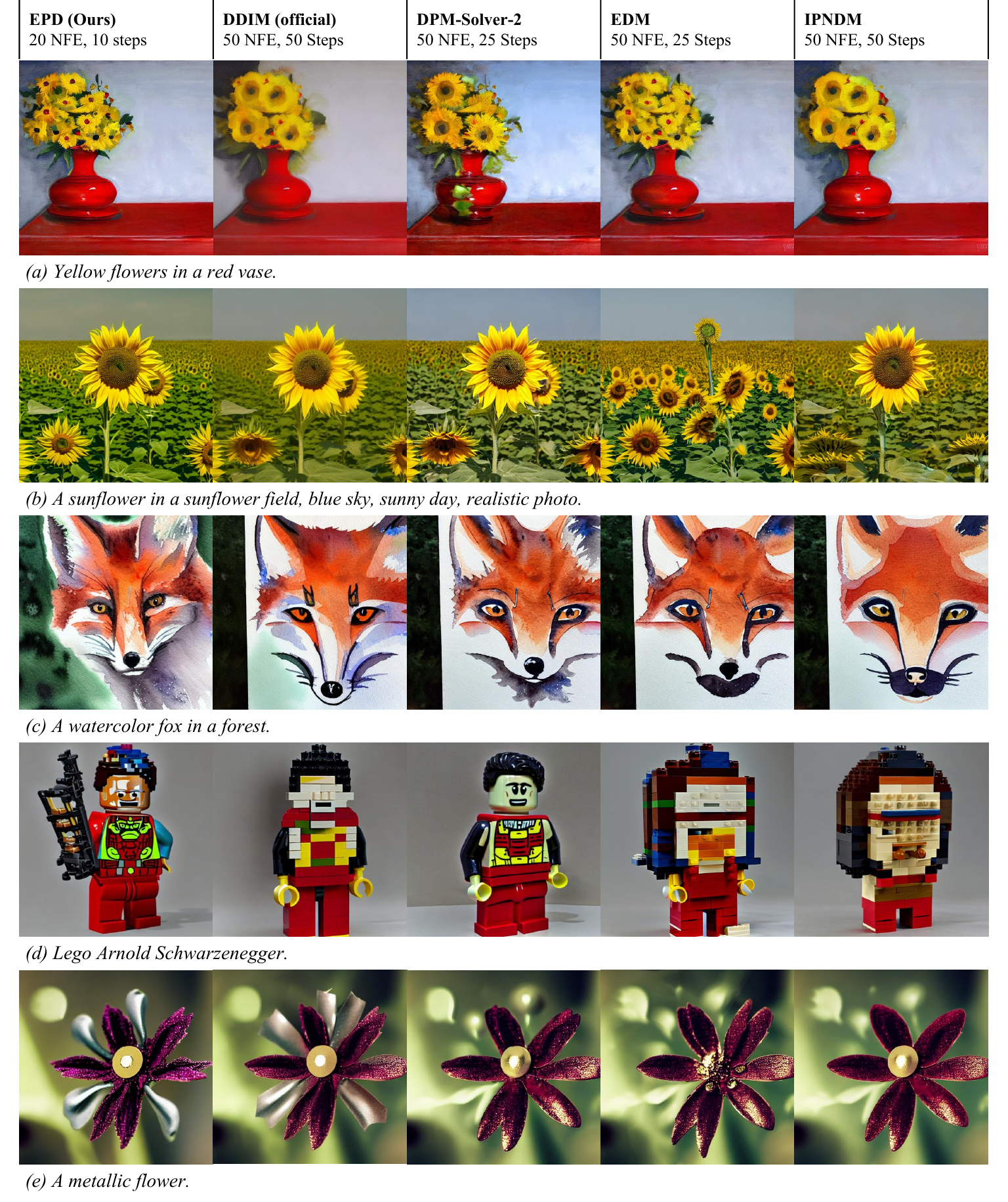}
    \caption{\textbf{Qualitative comparison of T2I generation results using Stable Diffusion v1.5}. We compare our \ours~(20 NFE) against SoTA baselines including DDIM, DPM-Solver-2, EDM, and iPNDM (50 NFE). Our method achieves comparable or superior visual fidelity \textit{with significantly reduced inference steps}. Qualitative results for SD3 are in \Cref{fig:sd3_512,fig:sd3_1024}. }
    \label{fig:sd15}
        \vspace{-1mm}
\end{figure*}
\begin{figure*}[t]
    \centering
    \includegraphics[width=0.82\linewidth]{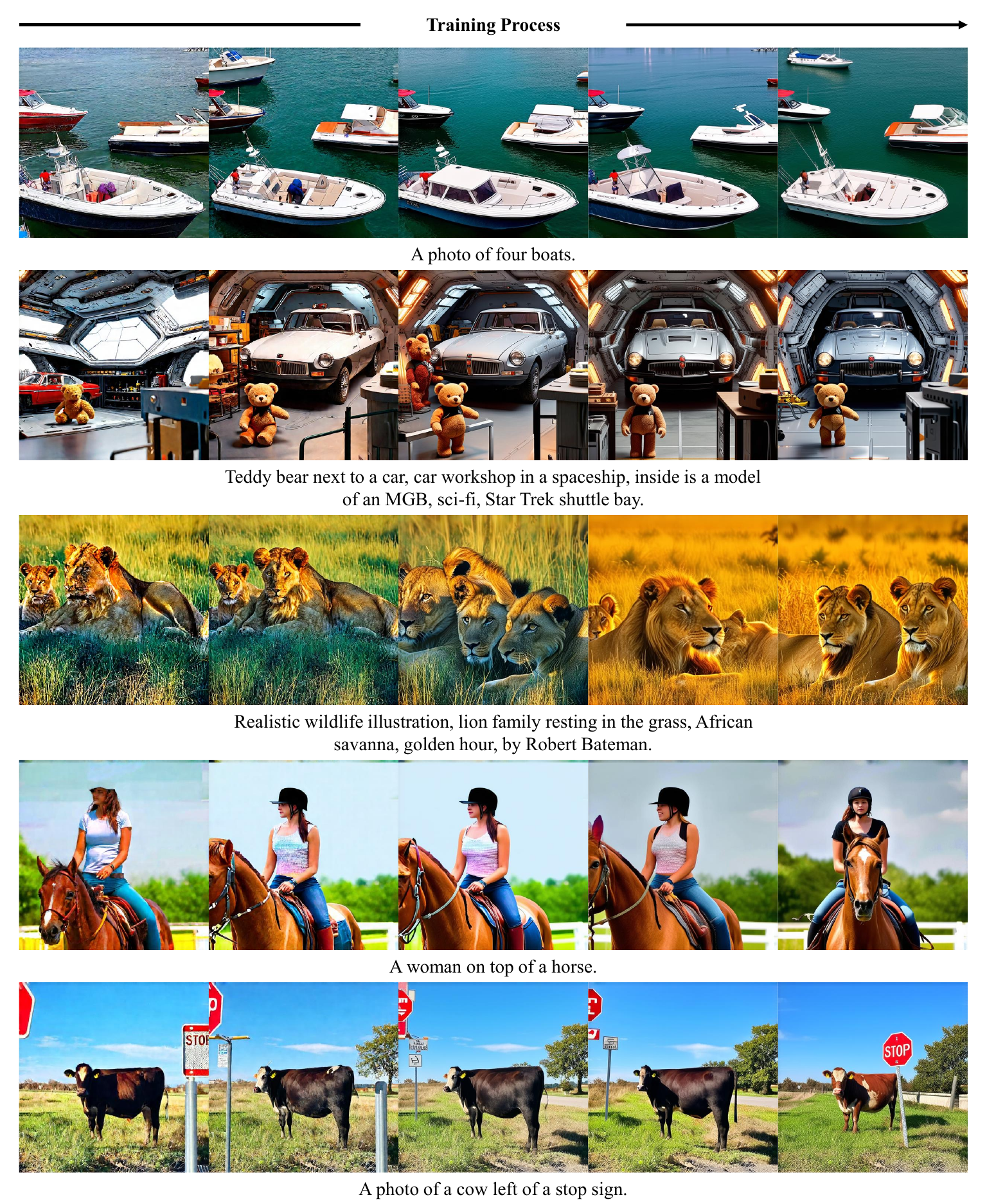}
    \caption{\textbf{Visual evolution of generated samples during training.} 
    We visualize the generation results of \textbf{SD3-Medium (512 $\times$ 512)} utilizing our EPD-Solver at different training checkpoints: Step 0, 1,000, 2,000, 5,000, and the optimal step.}
    \label{fig:training_process}
    \vspace{-1mm}
\end{figure*}
\section{Experiments}
\label{sec:exp}
This section is organized as follows:
\begin{itemize}
    \item \Cref{sec:exp_setup} introduces our experimental setup.
    \item \Cref{sec:main_results,sec:t2i} compares our $\ours$ and $\oursplugin$ with state-of-the-art ODE samplers in both quantitative and qualitative evaluations.
    \item \Cref{sec:numKanalysis} analyzes the impact of the number of parallel directions $K$ on image quality and inference latency.
    \item \Cref{sec:ablations} ablates the main components of $\ours$.
    % \item \Cref{sec:qualitative} showcases qualitative visualizations of the sampling process and generated images.
\end{itemize}

\subsection{Setup}\label{sec:exp_setup}

\vspace{0.5em}
\noindent\textbf{Models.}
We test out ODE solvers on diffusion-based image generation models, covering both pixel-space~\cite{karras2022elucidating} and latent-space models~\cite{rombach2022high}, across image resolutions ranging from 32 to 1024.
For pixel-space models, we evaluate the pretrained models on CIFAR 32$\times$32~\cite{krizhevsky2009learning}, FFHQ 64$\times$64~\cite{karras2019style}, ImageNet 64$\times$64~\cite{russakovsky2015imagenet} from~\cite{karras2022elucidating}. 
For latent-space models, we evaluate pretrained models on LSUN Bedroom 256$\times$256~\cite{yu2015lsun} from~\cite{rombach2022high}, as well as large-scale text-to-image (T2I) models, including \textbf{Stable Diffusion v1.5}~\cite{rombach2022high} at 512$\times$512 and \textbf{SD3-Medium}~\cite{sd3} at both 512$\times$512 and 1024$\times$1024 resolutions.

\vspace{0.5em}
\noindent\textbf{Baseline solvers.}
We compare against representative ODE solvers across three categories:
(1) Single-step solvers: DDIM~\cite{songdenoising}, EDM~\cite{karras2022elucidating}, DPM-Solver-2~\cite{lu2022dpm}, and AMED-Solver~\cite{zhou2024fast};
(2) Multi-step solvers: DPM-Solver++(3M)\cite{lu2022dpm_plus}, UniPC\cite{zhao2024unipc}, iPNDM~\cite{liupseudo,zhangfast}, and AMED-Plugin~\cite{zhou2024fast};
(3) Parallelism-based solver: ParaDiGMS~\cite{shih2023parallel}. 
For a fair comparison, we follow the recommended time schedules from their original papers~\cite{karras2022elucidating,lu2022dpm_plus,zhao2024unipc}. Specifically, we use the logSNR schedule for DPM-Solver-2, DPM-Solver++(3M), and UniPC, the time-uniform schedule for AMED-Solver~\cite{zhou2024fast}, while employing the polynomial time schedule with $\rho=7$ for the remaining baselines. Please refer to~\cref{sec:paradigm_details} for implementation details of baseline solvers.

\vspace{0.5em}
\noindent\textbf{Evaluation.} We evaluate our solvers with different sampling regimes corresponding to model scale. \textit{For validation experiments} (non-T2I models), we operate under low-latency constraints ($\text{NFE}\in \{3,5,7,9\}$) with AFS~\cite{dockhorn2022genie} applied; \textit{for large-scale text-to-image (T2I) experiments}, we adopt a 20-NFE setting. When $K=1$, our method shares the same computational cost as the baseline solvers.
For $K>1$, although each step involves $K-1$ additional function evaluations, these evaluations are fully parallelizable and incur only minimal overhead in inference latency.
We refer to the effective cost under parallel execution as Parallel NFE (Para. NFE).

We assess sample quality using the Fréchet Inception Distance (FID). For unconditional and class-conditional validation, FID is computed over 50k generated images. For T2I models, we evaluate FID by generating 10k images using prompts from the MS-COCO validation set~\cite{lin2014microsoft}. 
To further evaluate T2I alignment and human aesthetic preference, we evaluate on the DrawBench dataset~\cite{saharia2022photorealistic} with 1k generated images and report a series of metrics, including ImageReward~\cite{xu2023imagereward}, HPS v2.1~\cite{hpsv2}, CLIP score~\cite{clip}, and PickScore~\cite{pickscore}.

\noindent\textbf{Implementation details.} 
For \textit{validation experiments}, we adopt the distillation-based approach (stage 1). We optimize the parameters using the Adam optimizer on 10k images with a batch size of 32. To mitigate overfitting, we constrain $o_n$ and $\delta_n^k$ to the range $[-0.05, 0.05]$ using the sigmoid trick. The teacher trajectories are generated using DPM-Solver-2 with $M=6$ intermediate timesteps. Given the small parameter size (ranging from 6 to 45), the training is highly efficient—taking $\sim$3 minutes for CIFAR-10 $32\times32$ on a single NVIDIA 4090 and $\sim$20 minutes for LSUN Bedroom $256\times256$ on four NVIDIA A800 GPUs.

For \textit{T2I experiments}, we further employ the residual Dirichlet policy framework (stage 2) that operates without a teacher model. We utilize both Stable Diffusion v1.5~\cite{rombach2022high} and SD3-Medium~\cite{sd3} as backbones. The training is conducted on the Pick-a-Pic dataset~\cite{pickscore}, employing HPSv2.1~\cite{hpsv2} as the reward model to align with human preferences. 
For classifier-free guidance (CFG), we set the scale to 7.5 for Stable Diffusion v1.5 and 4.5 for SD3-Medium. 
Additional implementation details are provided in \cref{sec:implement-epd}.

\subsection{Validation Experiments Results}\label{sec:main_results}
In~\cref{tab:main_results}, we compare the FID scores of images generated by our \ours~with $K=2$ against baseline solvers across the CIFAR-10, FFHQ, ImageNet, and LSUN Bedroom datasets. The results demonstrate consistent and significant improvements from our learned directions across all datasets and NFE values. Specifically, with 9 (Para.) NFE, we achieve FID scores of 4.27 and 5.01 on the ImageNet and LSUN datasets, respectively, while the second-best baseline counterpart achieves 5.44 and 5.65, showing a notable improvement. Moreover, in the low NFE region, such as 3 NFE on LSUN Bedroom, our $\ours$ achieves a remarkable 13.21 FID, significantly outperforming the second-best baseline solver (AMED-Solver), which achieves 58.21 FID. 
We further evaluate \oursplugin~applied to the iPNDM solver, and observe that it outperforms \ours~when $\text{NFE} > 7$, consistent with our expectation that iPNDM benefits from historical gradients only when the step is sufficiently large. With small NFE, this advantage is less pronounced. We also present qualitative visual comparisons on all four datasets in \Cref{Qualitative_results,fig:sup_grid_cifar10_3,fig:sup_grid_ffhq_3,fig:sup_grid_img_3}.

\makeatletter
\renewcommand\thesubtable{(\alph{subtable})}
\makeatother

\begin{table*}[t!]
\small 
\captionsetup[subfloat]{labelformat=simple, labelsep=space}
\caption{
Validation experiment results across four datasets: 
{(a)} CIFAR10, 
{(b)} FFHQ, 
{(c)} ImageNet, 
{(d)} LSUN Bedroom. 
We compared our \ours~and \oursplugin~with (1) Single-step solvers: DDIM, Heun, DPM-Solver-2 and AMED-Solver, (2) Multi-step solvers: DPM-Solver++(3M), UniPC, iPNDM and AMED-Plugin, (3) Parallelism-based solver: ParaDiGMS. The best results are in \textbf{bold}, the second best are \underline{underlined}. See~\Cref{sec:learnedparameters} for the value of the learned parameters of \ours~and~\oursplugin. Qualitative visual comparisons are in \Cref{Qualitative_results,fig:sup_grid_cifar10_3,fig:sup_grid_ffhq_3,fig:sup_grid_img_3}.
}
\begin{minipage}[t]{0.48\textwidth}
\fontsize{8}{10}\selectfont
\subfloat[Unconditional generation on \textbf{CIFAR10} $32 \times 32$ \cite{krizhevsky2009learning}]{
\begin{tabular}{llcccc}
\toprule
 &\multirow{2}{*}{Method} & \multicolumn{4}{c}{(Para.) NFE} \\
\cmidrule{3-6} & & 3 & 5 & 7 & 9 \\
\midrule
\multirow{4}{*}{\rotatebox{90}{Single-step}} & DDIM~\cite{songdenoising} & 93.36 & 49.66 & 27.93 & 18.43 \\
&Heun~\cite{karras2022elucidating} & 306.2 & 97.67 & 37.28 & 15.76 \\
&DPM-Solver-2~\cite{lu2022dpm} & 155.7 & 57.30 & 10.20 & 4.98 \\
&AMED-Solver~\cite{zhou2024fast} & 18.49 & 7.59 & 4.36 & 3.67 \\ \midrule
\multirow{4}{*}{\rotatebox{90}{Multi-step}} 
& DPM-Solver++(3M)~\cite{lu2022dpm_plus} & 110.0 & 24.97 & 6.74 & 3.42 \\
&UniPC~\cite{zhao2024unipc} & 109.6 & 23.98 & 5.83 & 3.21 \\
&iPNDM~\cite{liupseudo,zhangfast} & 47.98 & 13.59 & 5.08 & 3.17 \\ 
&AMED-Plugin~\cite{zhou2024fast} & 10.81 & 6.61 & 3.65 & 2.63 \\ 
\midrule
\multirow{3}{*}{\rotatebox{90}{Parallel}} &ParaDiGMS~\cite{shih2023parallel} & 51.03 & 18.96 & 7.18 & 6.19 \\
&\cellcolor{lightCyan}\ours~(ours) & \cellcolor{lightCyan}\textbf{10.40}  & \cellcolor{lightCyan}\textbf{4.33}  & \cellcolor{lightCyan}\textbf{2.82} & \cellcolor{lightCyan}\underline{2.49}\\
& \cellcolor{lightCyan}\oursplugin~(ours) & \cellcolor{lightCyan}\underline{10.54} & \cellcolor{lightCyan}\underline{4.47} & \cellcolor{lightCyan}\underline{3.27} & \cellcolor{lightCyan}\textbf{2.42} \\
\bottomrule
\end{tabular}
}
\vspace{0.5em} 
\subfloat[Unconditional generation on \textbf{FFHQ} $64 \times 64$ \cite{karras2019style}]{
\begin{tabular}{llcccc}
\toprule
 &\multirow{2}{*}{Method} & \multicolumn{4}{c}{(Para.) NFE} \\
\cmidrule{3-6} & & 3 & 5 & 7 & 9 \\
\midrule
\multirow{4}{*}{\rotatebox{90}{Single-step}} 
&DDIM~\cite{songdenoising} & 78.21 & 43.93 & 28.86 & 21.01 \\
&Heun~\cite{karras2022elucidating} & 356.5 &	116.7 	&54.51 &	28.86  \\
&DPM-Solver-2~\cite{lu2022dpm} & 266.0 & 87.10 & 22.59 & 9.26 \\
&AMED-Solver~\cite{zhou2024fast}  & 47.31 & 14.80 & 8.82 & 6.31 \\ \midrule
\multirow{4}{*}{\rotatebox{90}{Multi-step}} 
&DPM-Solver++(3M)~\cite{lu2022dpm_plus} & 86.45 & 22.51 & 8.44 & 4.77 \\
&UniPC~\cite{zhao2024unipc} & 86.43 & 21.40 & 7.44 & 4.47 \\ 
&iPNDM~\cite{liupseudo,zhangfast} & 45.98 & 17.17 & 7.79 & 4.58 \\ 
&AMED-Plugin~\cite{zhou2024fast}  & 26.87 & 12.49 & 6.64  & 4.24 \\\midrule
\multirow{3}{*}{\rotatebox{90}{Parallel}} 
&ParaDiGMS~\cite{shih2023parallel} & 43.64 & 20.92 & 16.39 & 8.81 \\
&\cellcolor{lightCyan}\ours~(ours) & \cellcolor{lightCyan}\underline{21.74}  & \cellcolor{lightCyan}\textbf{7.84} & \cellcolor{lightCyan}\textbf{4.81} & \cellcolor{lightCyan}\underline{3.82}\\
&\cellcolor{lightCyan}\oursplugin~(ours) &\cellcolor{lightCyan}\textbf{19.02}&\cellcolor{lightCyan}\underline{7.97}&\cellcolor{lightCyan}\underline{5.09}& \cellcolor{lightCyan}\textbf{3.53} \\
\bottomrule
\end{tabular}
}
\end{minipage}
\hfill
\begin{minipage}[t]{0.48\textwidth}
\centering
  \fontsize{8}{10}\selectfont

\subfloat[Conditional generation on \textbf{ImageNet} $64 \times 64$ \cite{russakovsky2015imagenet}]{
\begin{tabular}{llcccc}
\toprule
 &\multirow{2}{*}{Method} & \multicolumn{4}{c}{(Para.) NFE} \\
\cmidrule{3-6} & & 3 & 5 & 7 & 9 \\
\midrule
\multirow{4}{*}{\rotatebox{90}{Single-step}} 
&DDIM~\cite{songdenoising} & 82.96 & 43.81 & 27.46 & 19.27 \\
&Heun~\cite{karras2022elucidating} & 249.4 & 89.63 & 37.65 & 16.76 \\
&DPM-Solver-2~\cite{lu2022dpm} & 140.2 & 42.41 & 12.03 & 6.64 \\ 
&AMED-Solver~\cite{zhou2024fast} & 38.10 & 10.74 & 6.66 & 5.44 \\ \midrule
\multirow{4}{*}{\rotatebox{90}{Multi-step}} 
&DPM-Solver++(3M)~\cite{lu2022dpm_plus} & 91.52 & 25.49 & 10.14 & 6.48 \\
&UniPC~\cite{zhao2024unipc} & 91.38 & 24.36 & 9.57 & 6.34 \\
&iPNDM~\cite{liupseudo,zhangfast} & 58.53 & 18.99 & 9.17 & 5.91 \\
&AMED-Plugin~\cite{zhou2024fast} & 28.06  & 13.83  & 7.81  & 5.60 \\ \midrule
\multirow{3}{*}{\rotatebox{90}{Parallel}} 
&ParaDiGMS~\cite{shih2023parallel} & 41.11 & 17.27 & 13.67 & 6.38 \\
&\cellcolor{lightCyan}\ours~(ours) & \cellcolor{lightCyan}\textbf{18.28} & \cellcolor{lightCyan}\textbf{6.35} & \cellcolor{lightCyan}\underline{5.26} & \cellcolor{lightCyan}\underline{4.27}\\
&\cellcolor{lightCyan}\oursplugin~(ours) & \cellcolor{lightCyan}\underline{19.89} & \cellcolor{lightCyan}\underline{8.17} & \cellcolor{lightCyan}\textbf{4.81} & \cellcolor{lightCyan}\textbf{4.02} \\
\bottomrule
\end{tabular}
}

\vspace{0.5em}

\subfloat[Unconditional generation on \textbf{LSUN Bedroom} $256 \times 256$ \cite{yu2015lsun}]{
\begin{tabular}{llcccc}
\toprule
 &\multirow{2}{*}{Method} & \multicolumn{4}{c}{(Para.) NFE} \\
\cmidrule{3-6} & & 3 & 5 & 7 & 9 \\
\midrule
\multirow{4}{*}{\rotatebox{90}{Single-step}} 
&DDIM~\cite{songdenoising} & 86.13 & 34.34 & 19.50 & 13.26 \\
&Heun~\cite{karras2022elucidating} &
291.5	 & 175.7	 & 78.67	  & 35.67  \\
&DPM-Solver-2~\cite{lu2022dpm} & 210.6 & 80.60 & 23.25 & 9.61 \\
&AMED-Solver~\cite{zhou2024fast} & 58.21 & 13.20 & 7.10 & 5.65 \\ \midrule
\multirow{4}{*}{\rotatebox{90}{Multi-step}} 
&DPM-Solver++(3M)~\cite{lu2022dpm_plus} & 111.9 & 23.15 & 8.87 & 6.45 \\
&UniPC~\cite{zhao2024unipc} & 112.3 & 23.34 & 8.73 & 6.61 \\
&iPNDM~\cite{liupseudo,zhangfast} & 80.99 & 26.65 & 13.80 & 8.38 \\
&AMED-Plugin~\cite{zhou2024fast} & 101.5 & 25.68 & 8.63 & 7.82 \\ \midrule
\multirow{3}{*}{\rotatebox{90}{Parallel}} 
&ParaDiGMS~\cite{shih2023parallel} & 100.3 & 31.68 & 15.85 & 8.56 \\
&\cellcolor{lightCyan}\ours~(ours) & \cellcolor{lightCyan}\textbf{13.21} & \cellcolor{lightCyan}\textbf{7.52} & \cellcolor{lightCyan}\underline{5.97} & \cellcolor{lightCyan}\underline{5.01} \\
&\cellcolor{lightCyan}\oursplugin~(ours) & \cellcolor{lightCyan}\underline{14.12} & \cellcolor{lightCyan}\underline{8.26} & \cellcolor{lightCyan}\textbf{5.24} & \cellcolor{lightCyan}\textbf{4.51} \\
\bottomrule
\end{tabular}
}
\end{minipage}

\label{tab:main_results}

\end{table*}

\subsection{Text-to-Image Experiment Results}
\label{sec:t2i}
\begin{table*}[t!]
\tabstyle{14pt}
\centering
\fontsize{8}{10}\selectfont 

% 颜色定义 (如果主文档未定义需取消注释)
% \definecolor{lightCyan}{rgb}{0.88,1,1}

\caption{
\textbf{Quantitative comparison of solvers on Stable Diffusion v1.5 (512 $\times$ 512) \cite{rombach2022high}.}
The best results are in \textbf{bold}, and the second best are \underline{underlined}. Qualitative results are in \Cref{fig:sd15}. See \Cref{tab:optimized_parameters_sd} (a) for the value of the learned parameters. 
}
\label{tab:schedule_comparison_grouped}

% 8 columns: Method, Step, Schedule, HPS, Pick, IR, CLIP, Aes
\begin{tabular}{lll ccccc}
\toprule
\textbf{Method} & \textbf{Step} & \textbf{Schedule Type} & \textbf{HPSv2.1} & \textbf{PickScore} & \textbf{ImageReward} & \textbf{CLIP} & \textbf{Aesthetic} \\
\midrule
% ==============================================
% Block 1: High NFE (approx 50)
% ==============================================
\multicolumn{8}{l}{\textit{Standard Setting: (Para.) NFE $=$ 50}} \\
% \midrule
DDIM (Default) & 50 & Uniform & 0.2454 & 21.1705 & -0.0020 & \underline{0.2701} & \textbf{5.2630} \\
iPNDM (3M) & 50 & Uniform & \textbf{0.2474} & \textbf{21.1860} & 0.0234 & 0.2700 & \underline{5.2599} \\
\multirow{2}{*}{Heun} & \multirow{2}{*}{25} & Polynomial & 0.2471 & 21.1498 & 0.0167 & 0.2697 & 5.2385 \\
 &  & Uniform & \underline{0.2473} & \underline{21.1733} & 0.0229 & \underline{0.2701} & 5.2522 \\
\multirow{2}{*}{DPM-Solver-2} & \multirow{2}{*}{25} & Uniform & \underline{0.2473} & 21.1632 & \underline{0.0263} & 0.2699 & 5.2429 \\
 &  & LogSNR & 0.2470 & 21.1632 & \textbf{0.0275} & \textbf{0.2702} & 5.2411 \\

\midrule
% ==============================================
% Block 2: Low NFE (approx 20)
% ==============================================
\multicolumn{8}{l}{\textit{Low-Step Setting: (Para.) NFE $=$ 20}} \\
% \midrule
DDIM & 20 & Uniform & 0.2416 & 21.1275 & -0.0411 & \textbf{0.2705} & \textbf{5.2552} \\
iPNDM (3M)& 20 & Uniform & 0.2454 & \textbf{21.1341} & \underline{-0.0170} & 0.2691 & \underline{5.2546} \\
\multirow{2}{*}{Heun} & \multirow{2}{*}{10} & Polynomial & 0.2387 & 20.8721 & -0.1784 & 0.2680 & 5.1193 \\
 &  & Uniform & 0.2443 & 21.0318 & -0.0244 & 0.2692 & 5.1723 \\
\multirow{2}{*}{DPM-Solver-2} & \multirow{2}{*}{10} & LogSNR & 0.2364 & 20.7718 & -0.1622 & 0.2660 & 5.1278 \\
 &  & Uniform & 0.2435 & 20.9930 & -0.0204 & 0.2689 & 5.1708 \\
% \midrule
% Ours 部分
% \rowcolor{lightCyan}
% \ours~(Stage 1) & 10 & Uniform & 0.2430 & 21.0512 & 3.0405 & 0.2679 & 5.2188 \\
\rowcolor{lightCyan}
\ours & 10 & Uniform & \textbf{0.2482} & \underline{21.1302} & \textbf{0.0121} & \underline{0.2695} & 5.2388 \\
\bottomrule
\end{tabular}
\end{table*}

\vspace{0.5mm}
\begin{table*}[t!]
\tabstyle{14pt}
\centering
\fontsize{8}{10}\selectfont

% 颜色定义
\definecolor{lightCyan}{rgb}{0.88,1,1}

\caption{
\textbf{Quantitative comparison of solvers on SD3-Medium (512 $\times$ 512) \cite{sd3}.}
The best results are in \textbf{bold}, and the second best are \underline{underlined}. Qualitative results are in \Cref{fig:sd3_512}. See \Cref{tab:optimized_parameters_sd} (b) for the value of the learned parameters. 
}
\label{tab:sd3_512}

% 8 columns: Solver, Step, Schedule Type, HPS, Pick, IR, CLIP, Aes
\begin{tabular}{lll ccccc}
\toprule
\textbf{Method} & \textbf{Step} & \textbf{Schedule Type} & \textbf{HPSv2.1} & \textbf{PickScore} & \textbf{ImageReward} & \textbf{CLIP} & \textbf{Aesthetic } \\
\midrule
% ==============================================
% Block 1: NFE = 28
% ==============================================
\multicolumn{8}{l}{\textit{Standard Setting: (Para.) NFE $=$ 28}} \\
% \midrule
DDIM (Default) & 28 & Shift Time & \textbf{0.2734} & \textbf{22.0357} & \textbf{0.7877} & \textbf{0.2820} & \textbf{5.2562} \\
Heun   & 14 & Shift Time & 0.2685 & \underline{21.8897} & 0.7622 & \underline{0.2819} & 5.1382 \\
DPM-Solver-2   & 14 & Shift Time & 0.2656 & 21.8513 & 0.7367 & 0.2810 & 5.1047 \\
iPNDM (3M) & 28 & Shift Time & \underline{0.2700} & 21.8742 & \underline{0.7624} & 0.2800 & \underline{5.1802} \\
\midrule
% ==============================================
% Block 2: NFE = 20
% ==============================================
\multicolumn{8}{l}{\textit{Low-Step Setting: (Para.) NFE $=$ 20}} \\
% \midrule
DDIM & 20 & Shift Time & \underline{0.2707} & \textbf{21.9684} & 0.7585 & 0.2807 & \underline{5.2620} \\
Heun   & 10 & Shift Time & 0.2644 & 21.8026 & 0.7095 & \textbf{0.2814} & 5.0824 \\
DPM-Solver-2   & 10 & Shift Time & 0.2631 & 21.7358 & 0.7081 & 0.2798 & 5.0572 \\
iPNDM (3M) & 20 & Shift Time & 0.2690 & 21.7928 & 0.7467 & 0.2799 & 5.1692 \\
% \midrule
% EPD Row
% \rowcolor{lightCyan}
% \ours~(Stage 1) & 10 & Shift Time & 0.2673 & 21.8272 & 3.7982 & 0.2780 & 5.1997 \\
\rowcolor{lightCyan}
\ours & 10 & Shift Time & \textbf{0.2742} & \underline{21.9514} & \textbf{0.7856} & \underline{0.2813} & \textbf{5.2743} \\
\bottomrule
\end{tabular}
\end{table*}

\begin{table*}[t!]
\tabstyle{14.5pt}
\centering
\fontsize{8}{10}\selectfont

% 颜色定义
\definecolor{lightCyan}{rgb}{0.88,1,1}

\caption{
\textbf{Quantitative comparison of solvers on SD3-Medium (1024 $\times$ 1024) \cite{sd3}.}
The best results are in \textbf{bold}, and the second best are \underline{underlined}. Qualitative results are in \Cref{fig:sd3_1024}. See \Cref{tab:optimized_parameters_sd} (c) for the value of the learned parameters. 
}
\label{tab:sd3_1024}

% 8 columns: Method, Step, Schedule, HPS, Pick, IR, CLIP, Aes
\begin{tabular}{lll ccccc}
\toprule
\textbf{Method} & \textbf{Step} & \textbf{Schedule Type} & \textbf{HPSv2.1} & \textbf{PickScore} & \textbf{ImageReward} & \textbf{CLIP} & \textbf{Aesthetic} \\
\midrule
% ==============================================
% Block 1: NFE = 28
% ==============================================
\multicolumn{8}{l}{\textit{Standard Setting: (Para.) NFE $=$ 28}} \\
% \midrule
DDIM (Default) & 28 & Shift Time & \textbf{0.2820} & \underline{22.4839} & 0.8796 & \underline{0.2854} & \textbf{5.3689} \\
Heun & 14 & Shift Time & 0.2790 & 22.3832 & 0.8622 & 0.2853 & 5.2688 \\
DPM-Solver-2 & 14 & Shift Time & 0.2817 & 22.4119 & \underline{0.9027} & \textbf{0.2866} & 5.2590 \\
iPNDM (3M) & 28 & Shift Time & \underline{0.2818} & \textbf{22.4841} & \textbf{0.9057} & 0.2850 & \underline{5.3565} \\
\midrule
% ==============================================
% Block 2: NFE = 20
% ==============================================
\multicolumn{8}{l}{\textit{Setting: (Para.) NFE $=$ 20}} \\
% \midrule
DDIM & 20 & Shift Time & 0.2769 & \underline{22.3774} & 0.8240 & 0.2850 & \underline{5.3623} \\
Heun & 10 & Shift Time & 0.2707 & 22.2173 & 0.7767 & \underline{0.2871} & 5.1892 \\
DPM-Solver-2 & 10 & Shift Time & 0.2759 & 22.2812 & 0.8323 & \textbf{0.2874} & 5.2458 \\
iPNDM (3M) & 20 & Shift Time & \underline{0.2805} & 22.3740 & \underline{0.8633} & 0.2840 & 5.3460 \\
% EPD Row
% \midrule
% \rowcolor{lightCyan}
% $\ours$ (Stage 1) & 10 & Shift Time & 0.2793 & 22.3633 & 3.9503 & 0.2864 & 5.2564 \\
\rowcolor{lightCyan}
$\ours$ & 10 & Shift Time & \textbf{0.2823} & \textbf{22.3942} & \textbf{0.8765} & 0.2852 & \textbf{5.3995} \\
\bottomrule
\end{tabular}
\end{table*}

\begin{figure*}[t!]
    \centering
    % 调整 width 参数可以控制图片大小，通常设为 \linewidth 或 0.95\linewidth
    \includegraphics[width=1.0\linewidth]{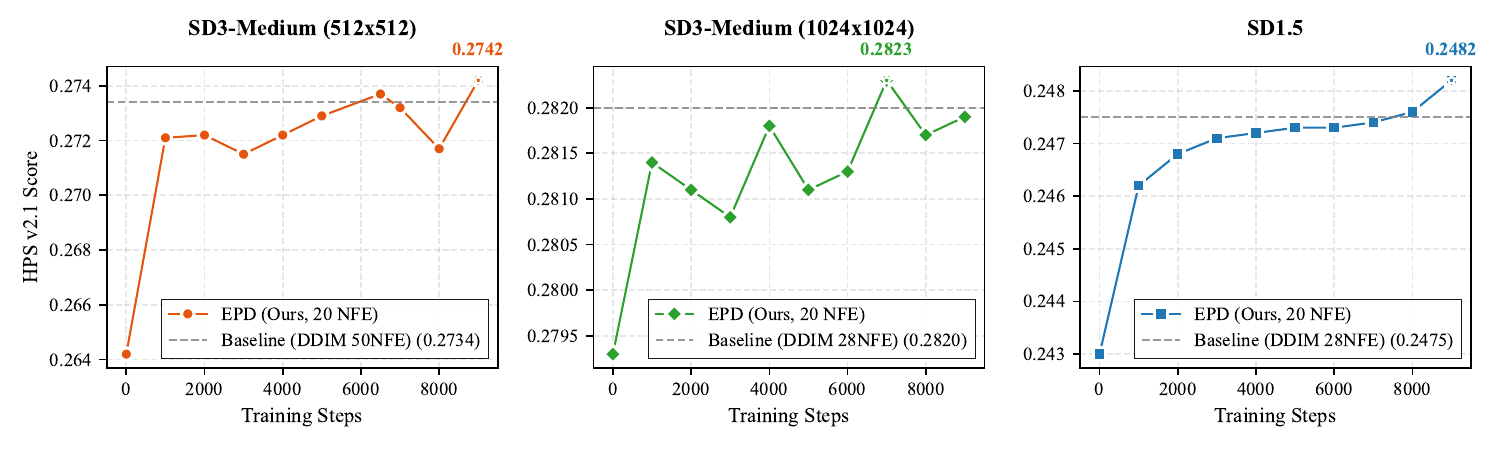}
    \vspace{-2mm} % 可选：微调图片与 Caption 之间的距离
    \caption{
    \textbf{Training dynamics of HPS v2.1 scores across different models and resolutions.} 
    We evaluate the model performance every 1,000 training steps over a total of 9,000 steps. 
    The solid curves represent our EPD method, while the gray dashed lines indicate the baseline performance (DDIM with default settings). 
    The star markers ($\star$) denote the peak performance achieved during training.
    As shown, our method demonstrates rapid convergence and consistently outperforms or matches the strong baselines across \textbf{(Left)} SD3-Medium ($512\times512$), \textbf{(Middle)} SD3-Medium ($1024\times1024$), and \textbf{(Right)} Stable Diffusion v1.5.
    }
    \label{fig:training_dynamics}
    \vspace{-10pt} % 可选：减少图表下方的空白，节省版面
\end{figure*}
% \input{figs/kl_divergence_comparison}

% Analysis of Stable Diffusion v1.5 (Based on Table II)
Table~\ref{tab:schedule_comparison_grouped} compares the quantitative results on Stable Diffusion v1.5 \cite{rombach2022high}. Under the low-latency setting (NFE = 20), $\ours$ demonstrates superior human alignment, achieving an HPSv2.1 score of \textbf{0.2482} \cite{hpsv2}. This result not only surpasses all baselines at the same computational budget but also outperforms the best-performing method at NFE=50 (iPNDM, 0.2474) \cite{zhangfast}. Furthermore, our RL fine-tuning significantly boosts the distilled base model, improving the ImageReward from 3.0405 to \textbf{3.1121} \cite{xu2023imagereward}. Overall, $\ours$ achieves generation quality competitive with 50-step solvers while requiring only 40\% of the inference steps, effectively bridging the gap between efficiency and generation fidelity.

% Analysis of SD3-Medium (Based on Tables III and IV) - Corrected ImageReward comparison
We further validate the scalability of our method on the recent SD3-Medium \cite{sd3} across different resolutions. As shown in Table~\ref{tab:sd3_512}, at $512\times512$ resolution with 20 NFE, our RL-tuned solver achieves an HPSv2.1 score of \textbf{0.2742}, outperforming the offical setting (DDIM at 28 NFE, 0.2734) in terms of human preference alignment. While the ImageReward score (3.8856) is slightly lower than the 28-step baseline (3.8877), it remains highly competitive given the reduced computational cost. Similarly, at $1024\times1024$ resolution (Table~\ref{tab:sd3_1024}), $\ours$ maintains its advantage with an HPSv2.1 score of \textbf{0.2823}, surpassing the 28-step DDIM baseline (0.2820). To examine the training dynamics, \Cref{fig:training_dynamics} plots the HPS v2.1 curves, demonstrating rapid convergence and consistent superiority over baselines across all resolutions. Complementing this, \Cref{fig:training_process} visualizes the qualitative evolution of generated samples. These findings indicate that our  Residual Dirichlet Policy Optimization generalizes effectively to large-scale text-to-image generation.

\begin{figure*}[t]
    \centering
\includegraphics[width=1.0\textwidth]{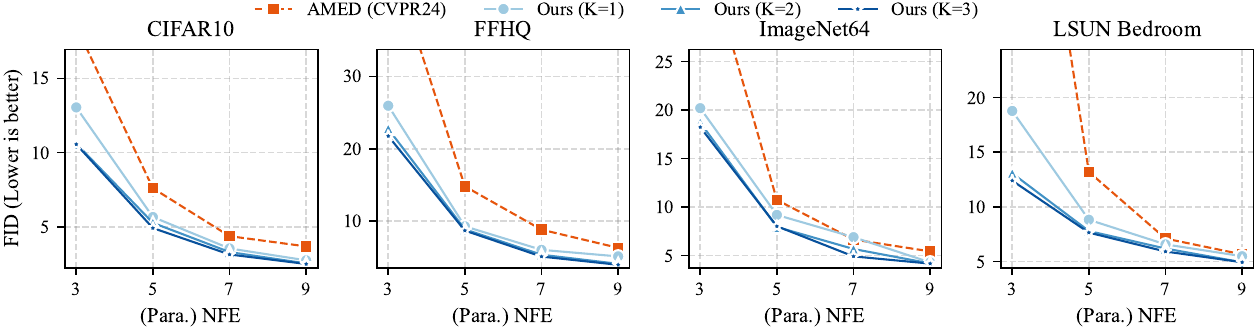}
    \caption{FID curves for different datasets and the number of parallel directions ($K$).} 
    \label{fig:fids_curve}
    % \vspace{-2mm}
\end{figure*}

\begin{table*}[t!]
\small 
\captionsetup[subfloat]{labelformat=simple, labelsep=space}
\caption{Latency (ms) measured across different datasets, Para. NFE values, and the number of parallel directions ($K$). No noticeable latency increase was observed when $K$ increased to 2. The reported values include the 95\% confidence interval. }
\begin{minipage}[t]{0.48\textwidth}
    \fontsize{8}{10}\selectfont
    \subfloat[\textbf{CIFAR10} and \textbf{FFHQ}]{
        \begin{tabular}{lccccc}
\toprule
&  \multirow{2}{*}{$K$}  & \multicolumn{4}{c}{Para. NFE}  \\ \cmidrule{3-6}
                          &     & 3    & 5    & 7   & 9   \\ \midrule
\multirow{3}{*}{\rotatebox{90}{CIFAR}} 
& $1$ &  28.1$\pm$0.84 & 47.2$\pm$0.88 & 63.5$\pm$0.71 & 80.5$\pm$0.73                  \\
& $2$ &  27.6$\pm$0.78 & 45.3$\pm$0.77 & 62.7$\pm$0.76& 79.8$\pm$0.81  \\
 & $3$ &  27.7$\pm$0.85 & 45.7$\pm$0.80 &  63.5$\pm$0.86 & 82.0$\pm$0.94  \\ \midrule   
\multirow{3}{*}{\rotatebox{90}{FFHQ}}
& $1$ & 34.4$\pm$0.79 &  56.1$\pm$0.78  &  77.4$\pm$0.96 & 100.4$\pm$0.74                \\
 & $2$ & 34.4$\pm$0.85  &  56.4$\pm$0.83 &  79.6$\pm$0.92 & 98.6$\pm$0.83    \\
& $3$ & 34.1$\pm$0.92  &   56.0$\pm$0.88 & 78.0$\pm$0.89 & 99.8$\pm$0.94    \\ \bottomrule
\end{tabular}
        }
\end{minipage}\hfill
\begin{minipage}[t]{0.48\textwidth}
    \fontsize{8}{10}\selectfont
        \centering
    \subfloat[ \textbf{ImageNet} and \textbf{LSUN Bedroom}]{
        \begin{tabular}{lccccc}
\toprule
&  \multirow{2}{*}{$K$}  & \multicolumn{4}{c}{Para. NFE}  \\ \cmidrule{3-6}
                          &     & 3    & 5    & 7   & 9   \\ \midrule
\multirow{3}{*}{\rotatebox{90}{IN}} & $1$ & 56.7$\pm$1.09   & 93.3$\pm$1.04  & 128.2$\pm$1.06   &  163.2$\pm$1.08                  \\
& $2$ &  55.7$\pm$1.16  &   92.3$\pm$1.18 & 128.2$\pm$1.14 & 164.4$\pm$1.23   \\
& $3$ & 55.7$\pm$1.20  &   94.7$\pm$1.20     &129.9$\pm$1.21 & 162.8$\pm$1.20    \\ \midrule   
\multirow{3}{*}{\rotatebox{90}{LSUN}} 
& $1$ & 57.5$\pm$1.26  & 78.8$\pm$1.02 & 104.3$\pm$1.15  & 131.1$\pm$1.03                 \\
& $2$ &  56.6$\pm$1.16 & 82.6$\pm$1.12 & 109.6$\pm$1.10 &  138.9$\pm$1.23 \\
& $3$ &  57.9$\pm$1.15 & 86.2$\pm$1.16 & 117.8$\pm$1.10 & 147.8$\pm$1.19  \\ \bottomrule 
\end{tabular}
        }
\end{minipage}

\label{tab:latency}
% \vspace{-5mm}
\end{table*}

\begin{table}[t!]
    \centering
    \caption{Inference latency (s) and peak memory (GB) usage on SD1.5 and SD3-Medium with 20 sampling steps.}
    \label{tab:latency_memory}
    \begin{tabular}{lcccc}
        \toprule
        \textbf{Model} & \textbf{Configuration} & \textbf{latency} & \textbf{Peak Memory} \\
        \midrule
        \multirow{2}{*}{SD1.5} 
            & $K=1$ & $0.6048 \pm 0.0170$ & 6.68 \\
            & $K=2$ & $0.6252 \pm 0.0465$ & 7.71 \\
        \midrule
        \multirow{2}{*}{SD3-Medium} 
            & $K=1$ & $0.5461 \pm 0.0835$ & 25.50 \\
            & $K=2$ & $0.5961 \pm 0.0432$ & 25.50 \\
        \bottomrule
    \end{tabular}
\end{table}
\begin{table*}[t]
    \centering
    
    \begin{minipage}[t]{0.33\textwidth}
            \caption{Effect of scaling factors.}\label{tab:scaling} 
        \centering
        \fontsize{8}{10}\selectfont
         \setlength{\tabcolsep}{5pt} 
        \begin{tabular}{lcccc}
        \toprule
         Para. NFE & 3 & 5 & 7 & 9 \\
        \midrule
        $\ours$ & \textbf{10.40} & \textbf{4.33} & \textbf{2.82} & \textbf{2.49} \\ 
        \quad w.o. $o_n$ &13.25 & 	5.84 &	3.59	& 2.79  \\
        \quad w.o. $\delta_n^k$ & 13.02	&5.47	 & 3.23	& 2.69 \\
        \quad w.o. $o_n$ \& $\delta_n^k$ &16.01	&6.62&	4.24&	3.24 \\
        \bottomrule
        \end{tabular}
  
    \end{minipage}
    \hfill
    \begin{minipage}[t]{0.33\textwidth}
            \caption{Effect of time schedules.}\label{tab:schedule}
        \centering
        \fontsize{8}{10}\selectfont
         \setlength{\tabcolsep}{5pt} 
        \begin{tabular}{lcccc}
        \toprule
        \multirow{2}{*}{Schedule} & \multicolumn{4}{c}{Para. NFE} \\
        % \cmidrule{2-5} 
        & 3 & 5 & 7 & 9 \\
        \midrule
        LogSNR \cite{lu2022dpm} &54.07&8.88&7.95& 3.97\\
        Polynomial~\cite{karras2022elucidating} & 11.10 &8.89  & 4.50& 3.72 \\
        Uniform  & \textbf{10.40} & \textbf{4.33} & \textbf{2.82} & \textbf{2.49} \\
        \bottomrule
        \end{tabular}

    \end{minipage}
    \hfill
    \begin{minipage}[t]{0.33\textwidth}
            \caption{Effect of teacher solvers.}
        \label{tab:teacher}      
         \centering
        \fontsize{8}{10}\selectfont 
         \setlength{\tabcolsep}{5pt} 
        \begin{tabular}{lcccc}
        \toprule
        \multirow{2}{*}{Teacher Solver} & \multicolumn{4}{c}{Para. NFE} \\
        % \cmidrule{2-5}
        & 3 & 5 & 7 & 9 \\
        \midrule
                Heun \cite{karras2022elucidating} & 15.91 & 6.65 & 4.61 & 3.57 \\
               iPNDM \cite{liupseudo,krizhevsky2009learning} & 13.69 & 6.64 & 4.59 & 3.59 \\
                DPM-Solver-2 \cite{lu2022dpm} & \textbf{10.40} & \textbf{4.33} & \textbf{2.82} & \textbf{2.49} \\
        \bottomrule
        \end{tabular}
    \end{minipage}
    % \vspace{-2mm}
\end{table*}

\begin{figure*}[t]
    \centering
\includegraphics[width=1\textwidth]{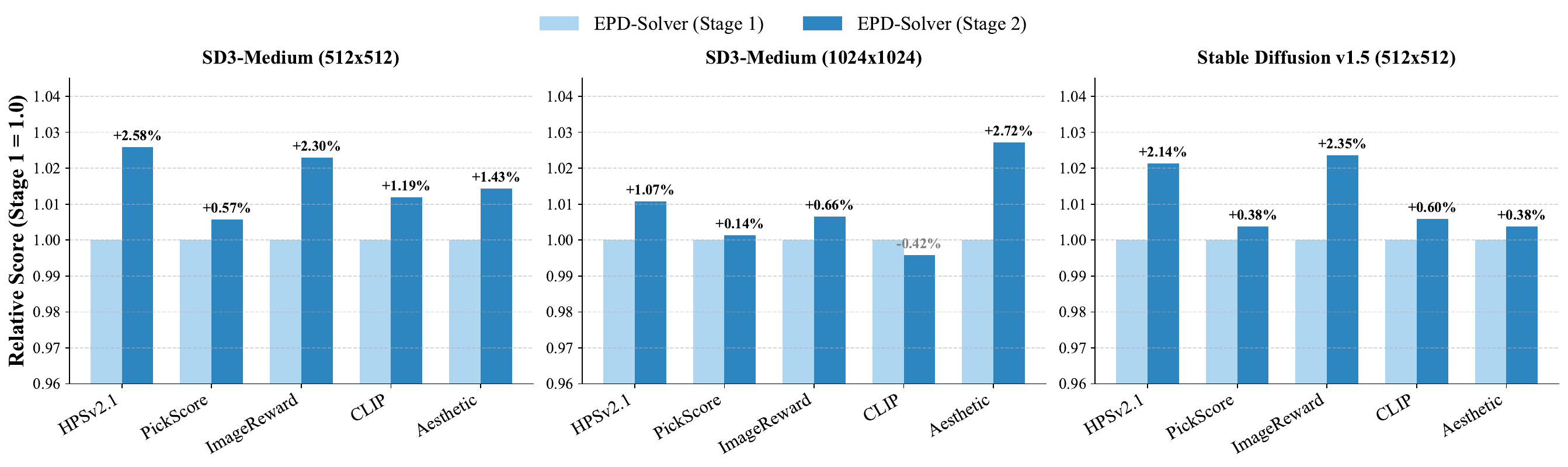}
    \caption{Quantitative improvement of Stage 2 over Stage 1. We report the relative scores of our EPD-Solver (Stage 2) normalized by the results of EPD-Solver (Stage 1) across three settings: SD3-Medium ($512\times512$), SD3-Medium ($1024\times1024$), and Stable Diffusion v1.5. Stage 2 consistently improves human preference metrics (HPSv2.1, PickScore, ImageReward) and Aesthetic scores across all models, demonstrating the effectiveness of our Residual Dirichlet Policy Optimization.}
    \label{fig:rl_ablation}
        \vspace{-1mm}
\end{figure*}
\subsection{On the Number of Parallel Directions}\label{sec:numKanalysis}

\noindent\textbf{Image quality with different values of $K$.}
In~\Cref{fig:fids_curve}, 
we compare the quality of images generated using our $\ours$ with different values of $K$. As expected, increasing the number of intermediate points leads to improved FID scores. For example, on the FFHQ dataset with 3 Para. NFE, the FID score decreases from 26.0 to 22.7 when $K$ increases from 1 to 2. Additionally, the results suggest that increasing the number of points beyond 2 yields diminishing returns. For instance, on ImageNet with 9 Para. NFE, the FID scores for $K=2$ and $K=3$ are 4.20 and 4.18, respectively, showing minimal improvement.

\noindent\textbf{Latency with different values of $K$.} 
Given that each intermediate gradient is fully parallelizable, we examine whether increasing $K$ noticeably impacts latency. 
For validation experiments, 
\Cref{tab:latency} presents inference latency on a single NVIDIA 4090, evaluated over 1000 generated images with a batch size of 1. We report the average inference time along with the 95\% confidence interval.
For CIFAR-10, FFHQ, and ImageNet, increasing $K$ to 3 does not noticeably impact latency. For LSUN Bedroom, we observe a slight increase in latency when $K=3$. However, earlier results show that $K=2$ already yields significant quality improvements. Therefore, setting $K=2$ provides an effective trade-off, achieving high-quality generation while avoiding additional inference cost.

We further evaluate the latency and peak memory footprint of \ours\ on large-scale T2I models.
Table~\ref{tab:latency} reports inference latency and peak GPU memory usage on Stable Diffusion v1.5 and SD3-Medium with 20 sampling steps, measured on a single NVIDIA H800 with batch size 1.
Increasing $K$ from 1 to 2 introduces only a modest increase in inference latency.
On SD1.5, the average latency increases from 0.605s to 0.625s, while on SD3-Medium the increase remains below 0.05s.
Notably, the peak memory usage remains unchanged on SD3-Medium and increases moderately on SD1.5.
These results indicate that the parallel evaluation of multiple intermediate gradients incurs minimal overhead even for large-scale diffusion models, making $K=2$ a practical choice that balances generation quality and inference efficiency.

\subsection{Ablation Studies}\label{sec:ablations}

\noindent\textbf{Effect of scaling factors.} \cite{ningelucidating,li2024alleviating} identify exposure bias—\ie, the input mismatch between training and sampling—as a key factor leading to error accumulation and sampling drift. To mitigate the bias, they propose scaling the gradient and shifting the timestep. Building on these insights, our \ours~introduces two learnable parameters: $o_n$ and $\delta_n^k$. We compare FID scores without these scaling factors to assess their impact. As shown in ~\Cref{tab:scaling}, omitting the scaling factors noticeably reduces image quality. For instance, without $o_n$, FID rises from 4.33 to 5.84 at Para. NFE = 5.

\noindent\textbf{Effect of time schedule.} In ~\Cref{tab:schedule}, we present results on CIFAR-10 using commonly used time schedules: LogSNR, EDM, and Time-uniform. Our solver consistently performs better with the time-uniform schedule.

\noindent\textbf{Effect of teacher ODE solvers.}
We study the impact of different teacher ODE solvers in \Cref{tab:teacher}. The results show that using DPM-Solver-2 to generate teacher trajectories achieves the best performance.
We hypothesize that this is because DPM-Solver-2 also estimates gradients using intermediate points, resulting in a smaller gap to our $\ours$.

\noindent\textbf{Effect of Residual Dirichlet Policy Optimization.} To validate the effectiveness of our parameter-efficient RL fine-tuning, we compare the performance of EPD-Solver before (Stage 1) and after (Stage 2) the Residual Dirichlet Policy Optimization. As illustrated in \Cref{fig:rl_ablation}, Stage 2 yields consistent improvements across multiple human alignment metrics compared to the distilled baseline. For instance, on Stable Diffusion v1.5, the RL fine-tuning significantly boosts the ImageReward score from -0.002 to 0.012. Similarly, on SD3-Medium ($512\times512$) at 20 NFE, our Stage 2 solver achieves an HPSv2.1 score of 0.2742, effectively bridging the gap to high-step baselines. These results confirm that while Stage 1 provides a robust initialization by capturing the trajectory curvature, Stage 2 is crucial for aligning the sampling behavior with human perceptual preferences without increasing inference cost.

\begin{figure}[t]
    \centering
    % width=1.0\linewidth 确保图片填满单栏宽度
    \includegraphics[width=0.95\linewidth]{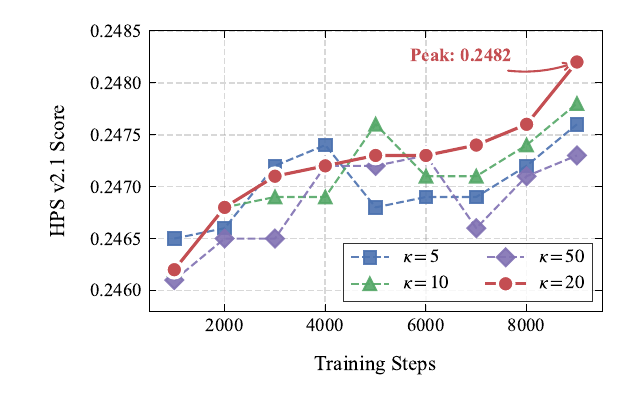}
    \vspace{-2mm} % 可选：微调图片与 Caption 的间距
    \caption{
    Ablation study on the concentration parameter $\kappa$.
    }
    \label{fig:ablation_alpha}
    \vspace{-10pt} % 可选：减少图表下方的空白
\end{figure}

\noindent\textbf{Effect of Dirichlet coefficient.}
We investigate the impact of the concentration parameter $\kappa$ in the Dirichlet distribution, which governs the exploration magnitude of the policy around the distilled solver parameters. We evaluated the training dynamics of HPS v2.1 scores across different coefficient values $\kappa \in \{5, 10, 20, 50\}$.
As illustrated in Fig.~\ref{fig:ablation_alpha}, the choice of $\kappa$ significantly influences the optimization process. The results indicate that the default setting ($\kappa = 20$) strikes the best balance between exploration and stability. Specifically, $\kappa=20$ achieves the highest peak performance with an HPS v2.1 score of \textbf{0.2482} and maintains stable convergence throughout the training steps. In contrast, other values (represented by dashed lines) exhibit either slower convergence or greater instability, failing to reach the peak performance attained by the default setting.

\section{Conclusion}

In this paper, we presented \ours, a novel ODE solver that exploits parallel gradient evaluations to reduce truncation errors, enabling higher-order accuracy at low latency. 
Our method is built upon a two-stage optimization framework.
In the first stage, we perform distillation-based optimization to learn a student EPD solver that accurately approximates high-fidelity sampling trajectories in the few-step regime.
In the second stage, we introduce a RL process based on Residual Dirichlet Policy, which further refines the solver behavior to better align generation with human preferences without modifying the DM itself.
Empirical results confirm that EPD-Solver establishes new state-of-the-art performance on standard benchmarks. Notably, it bridges the gap between efficiency and quality, significantly surpassing existing solvers on large-scale models like Stable Diffusion v1.5 and SD3-Medium with fewer function evaluations.

{
\bibliographystyle{IEEEtran}
\bibliography{cite}
}

\newpage
\clearpage
\appendices

\section{Additional Implementation Details}
\subsection{Implementation Details of \ours}\label{sec:implement-epd}
At each sampling step $n$ (from $t_{n+1}$ to $t_n$) in an $N$-step process, the solver provides a set of learned parameters $\Theta_n = \{\tau_n^k, \lambda_n^k, \delta_n^k, o_n\}_{k=1}^K$, implemented as follows:

\noindent \textbf{Intermediate timesteps} ($\tau_n^k$): These are points within $[t_n, t_{n+1}]$, computed via geometric interpolation. Specifically, the interpolation ratio $r_n^k \in [0, 1]$ is obtained by applying a sigmoid to a learnable scalar parameter, yielding
\begin{equation}
    \tau_n^k = t_{n+1}^{r_n^k} \cdot t_n^{1 - r_n^k}.
\end{equation}

\noindent \textbf{Simplex weights} ($\lambda_n^k$): These non-negative weights form a convex combination of the $K$ parallel gradients, satisfying $\sum_{k=1}^K \lambda_n^k = 1$. They are obtained by applying a softmax over $K$ learnable scalar parameters.

\noindent \textbf{Output scaling} ($o_n$): A learnable scalar that scales the overall update direction by a factor of $(1 + o_n)$ to mitigate exposure bias between training and sampling. To implement this, we introduce a per-branch modulation term $\sigma_n^k \in [-0.05, 0.05]$ that scales the corresponding weight $\lambda_n^k$. Specifically, we constrain $\sigma_n^k$ using a sigmoid-based transformation:
\[
\sigma_n^k = 0.1 \times (\texttt{sigmoid}(\tilde{\sigma}_n^k) - 0.5),
\]
where $\tilde{\sigma}_n^k$ is an unconstrained learnable parameter. The final scaling factor is then given by
\[
o_n = \sum_k \lambda_n^k \sigma_n^k - 1.
\]

\noindent \textbf{Timestep shifting} ($\delta_n^k$): A trainable perturbation applied to the intermediate timestep $\tau_n^k$, producing $\tau_n^k + \delta_n^k$ as input to the denoising network. We implement this by introducing a scaling factor $s_n^k$ that transforms $\tau_n^k$ into $s_n^k \tau_n^k$. The relationship between $s_n^k$ and $\delta_n^k$ is given by
\[
s_n^k \tau_n^k = \tau_n^k + \delta_n^k \quad \Rightarrow \quad \delta_n^k = (s_n^k - 1)\tau_n^k.
\]
To prevent overfitting, $s_n^k$ is constrained to a small range (\eg, $[0.95, 1.05]$) using a sigmoid-based transformation. Specifically, we map an unnormalized parameter $\tilde{s}_n^k$ as follows:
\[
s_n^k = 1 + 0.1 \times (\texttt{sigmoid}(\tilde{s}_n^k) - 0.5).
\]

\subsection{Implementation Details of \oursplugin}\label{sec:intro-plugin}
The \oursplugin~serves as a module integrated in any existing ODE solver. We illustrate this using the multi-step iPNDM~\cite{liupseudo,zhangfast} sampler as a representative implementation.
We begin with a brief review of the iPNDM sampler.

\noindent\textbf{Review of iPNDM.} Let $\mathbf{d}_t$ denote the estimated gradient at time step $t$, \ie, $\mathbf{d}_t=\epsilon_\theta(\mathbf{x}_t,t)$. The update at time step $t_n $ is given by:
\begin{align}
    \mathbf{d}_{t_{n+1}}'&=\tfrac{1}{24}(55\mathbf{d}_{t_{n+1}}-59\mathbf{d}_{t_{n+2}}+37\mathbf{d}_{t_{n+3}}-9\mathbf{d}_{t_{n+4}}) \nonumber \\
    \mathbf{x}_{t_n}&=\mathbf{x}_{t_{n+1}}+h_n \mathbf{d}_{t_{n+1}}'.
\end{align}
This rule applies for $n < N - 3$; for brevity, we present only this case. Other cases can be found in the original paper.

\noindent\textbf{Our EPD plugin for iPNDM.} Our plugin replaces $\mathbf{d}_{t_{n+1}}$ with a weighted combination of $K$ parallel intermediate gradients to reduce truncation error. Similar to \ours, we introduce the parameters at step $n$ as $\Theta_n = \{\tau_n^k, \lambda_n^k, \delta_n^k, o_n\}_{k=1}^K$. The gradient  is now estimated as 
\begin{equation}
    \mathbf{d}_{t_{n+1}}^{\mathsf{EPD}}=(1+o_n)\sum_{k=1}^K  \lambda^k_n\bm{\epsilon}_\theta(\rvx_{\tau^k_n},\tau^k_n+\delta^k_n).
\end{equation}
Accordingly, the update for \oursplugin~becomes:
\begin{align}
    \mathbf{d}_{t_{n+1}}'&=\tfrac{1}{24}(55\mathbf{d}_{t_{n+1}}^{\mathsf{EPD}}-59\mathbf{d}_{t_{n+2}}+37\mathbf{d}_{t_{n+3}}-9\mathbf{d}_{t_{n+4}}) \nonumber \\
    \mathbf{x}_{t_n}&=\mathbf{x}_{t_{n+1}}+h_n \mathbf{d}_{t_{n+1}}'.
\end{align}

\oursplugin~incurs minimal training overhead, in line with the lightweight design of the \ours. Thanks to its limited number of learnable parameters, the optimization process is highly efficient. 

\begin{table}[t]
\centering
\fontsize{8}{10}\selectfont
\begin{tabular}{lcccc}
\toprule
\multirow{2}{*}{Timesteps} & \multicolumn{4}{c}{Para. NFE} \\
\cmidrule{2-5} & 3 & 5 & 7 & 9 \\
\midrule
$t_n,t_{n+1}$ (EDM)& 306.2 & 97.67 & 37.28 & 15.76 \\
$\sqrt{t_nt_{n+1}},t_{n+1}$  & 129.6 & 16.51 & 9.86 & 7.06 \\
$\frac{1}{2}(t_n+t_{n+1}),t_{n+1}$ & 105.8 & 36.14 & 18.08 & 9.85 \\
$t_n,\sqrt{t_nt_{n+1}}$  &225.5&130.8&78.49&44.38\\
$t_n, \frac{1}{2}(t_n+t_{n+1})$ &198.6&119.6&59.23&32.21\\ 
$\sqrt{t_nt_{n+1}}, \frac{1}{2}(t_n+t_{n+1})$  &136.1 & 21.17&10.80&5.83 \\
random, $t_{n+1}$ &90.8&30.01&14.37&9.14 \\
random, random &110.7&57.1&22.86& 11.91\\
$\ours, K=2$ & {10.60}  & {5.26}  & {3.29} & {2.52}\\
\bottomrule
\end{tabular}
\caption{FID results on the choices of two intermediate points. Evaluations are conducted on CIFAR-10 \cite{krizhevsky2009learning}. Start point: $t_{n+1}$, end point: $t_n$, midpoints: $\sqrt{t_nt_{n+1}},\frac{1}{2}(t_n+t_{n+1})$, and `random' denotes  a midpoint randomly chosen from $[t_n,t_{n+1}]$.}\label{tab:midpoints_choice}
\end{table}

\subsection{Implementation Details of ParaDiGMS}\label{sec:paradigm_details}

For direct comparison with \texttt{EDP-\{Solver, Plugin\}}, we re-implemented the ParaDiGMS sampler \cite{shih2023parallel} in the EDM \cite{karras2022elucidating} framework, as its public implementation\footnote{https://github.com/AndyShih12/paradigms} is tailored for Stable Diffusion. To ensure a fair latency comparison with our single-GPU \ours, we run ParaDiGMS on two NVIDIA 4090 GPUs, distributing the workload evenly by matching the Para. NFE/GPU ratio.

Specifically, to align the parallel structure with \ours~($K=2$), we set the batch window size of ParaDiGMS to 2. The core principle was to adjust the tolerance parameter, ranging from $1 \times 10^{-2}$ to $1 \times 10^{-1}$, to calibrate the total Para. NFE. The ratio of Para. NFE / GPUs was set to 3, 5, 7 and 9, which ensures the per-GPU workload and latency level for ParaDiGMS roughly matches the single-GPU \ours. We also observed that the efficiency of ParaDiGMS is reduced in low-NFE regimes, as the substantial error per iteration causes its solver stride to frequently set to 1.

\subsection{Further Details on the Text-to-Image Experimental Setup}\label{sec:t2i_details}

\noindent \textbf{Quality metrics.}
The details of quality metrics are as follows:
\begin{itemize}
    \item \textbf{HPSv2.1}: Human preference model that blends text-image alignment and visual quality to mirror human scoring.
    \item \textbf{PickScore}: Multimodal human-preference scorer emphasizing joint text alignment and visual realism.
    \item \textbf{ImageReward}: General T2I human-preference reward model capturing text consistency, visual fidelity, and safety.
    \item \textbf{CLIP}: Contrastive language-image similarity metric measuring how well generated images match the prompt.
    \item \textbf{Aesthetic}: CLIP-feature linear regressor that predicts an image’s overall aesthetic quality.
\end{itemize}

\noindent \textbf{Hyperparameter specification.}
We provide the detailed hyperparameter settings for our experiments in \Cref{tab:hyperparameter}. We conducted experiments on three model configurations: Stable Diffusion v1.5, SD3-Medium (512 $\times$ 512), and SD3-Medium (1024 $\times$ 1024).

\begin{table}[h]
    \centering
    \caption{Hyperparameter specifications for different model configurations.}
    \label{tab:hyperparameter}
    \vspace{0.2cm}
    \resizebox{\linewidth}{!}{
        \begin{tabular}{lccc}
            \toprule
            \textbf{Parameter} & \textbf{SD3-Medium} & \textbf{SD3-Medium} & \textbf{Stable Diffusion v1.5} \\
            \midrule
            \multicolumn{4}{l}{\textit{Model Settings}} \\
            \midrule
            Resolution & $512 \times 512$ & $1024 \times 1024$ & $512 \times 512$ \\
            Guidance Scale (CFG) & 4.5 & 4.5 & 7.5 \\
            \midrule
            \multicolumn{4}{l}{\textit{RL Optimization}} \\
            \midrule
            Learning Rate & $7 \times 10^{-5}$ & $7 \times 10^{-5}$ & $7 \times 10^{-5}$ \\
            Rollout Batch Size & 16 & 8 & 8 \\
            Mini-batch Size & 4 & 4 & 4 \\
            PPO Epochs & 1 & 1 & 1 \\
            RLOO Samples & 4 & 4 & 4 \\
            Clip Range & 0.2 & 0.2 & 0.2 \\
            Dirichlet Concentration & 10 & 10 & 20 \\
            \midrule
            \multicolumn{4}{l}{\textit{Reward Configuration}} \\
            \midrule
            Reward Model & HPSv2.1 & HPSv2.1 & HPSv2.1 \\
            Reward Weight & 1.0 & 1.0 & 1.0 \\
            \bottomrule
        \end{tabular}
    }
\end{table}

\noindent \textbf{Compute resource specification.}
All experiments were conducted on a single NVIDIA H200 GPU. We report the number of training steps required to reach optimal performance and the corresponding wall-clock training time in \Cref{tab:train_converge}.

\begin{table}[h]
    \centering
    \caption{Training costs and convergence steps for different model configurations on a single NVIDIA H200 GPU.}
    \label{tab:train_converge}
    \vspace{0.2cm}
    \begin{tabular}{lccc}
        \toprule
        \textbf{Model} & \textbf{Optimal Steps}$^*$ & \textbf{Time (GPU Hours)} \\
        \midrule
        SD3-Medium (512 $\times$ 512)  & 9,000 & 24.0 \\
        SD3-Medium (1024 $\times$ 1024) & 7,000 & 34.9 \\
        Stable Diffusion v1.5   & 9,000 & 21.1 \\
        \bottomrule
    \end{tabular}
    
    \vspace{1mm}
    \parbox{\linewidth}{
        \footnotesize 
        \raggedright 
        $^*$The optimal step is determined by evaluating model performance every 1000 training steps.
    }
\end{table}
\begin{figure*}[t]
    \centering
\includegraphics[width=1.0\textwidth]{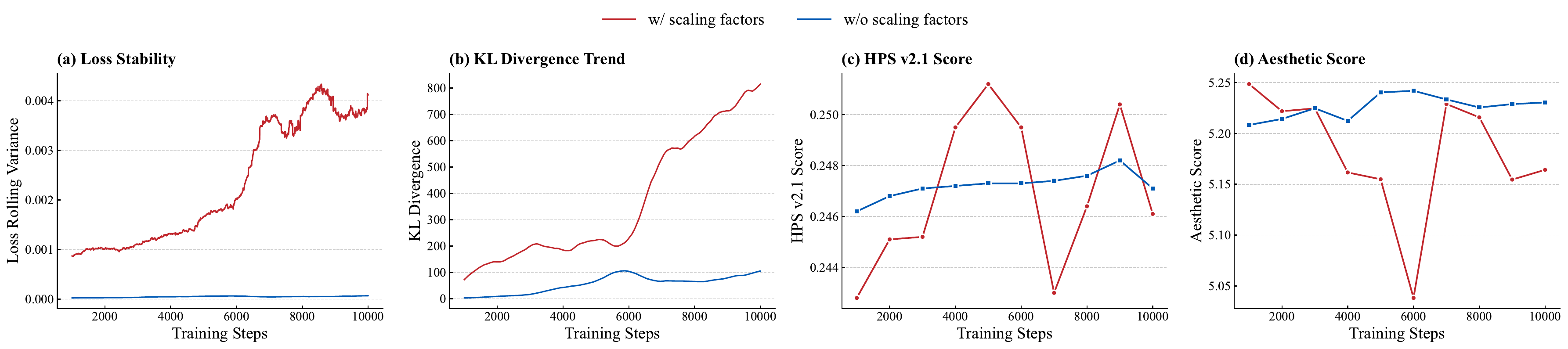}
    \caption{\textbf{Ablation study on the scaling factors during Stage 2.} 
  We compare the training dynamics with and without optimizing the scaling factors ($o_n$ and $\delta_n^k$) in the RL stage. 
  \textbf{(a) Loss Stability:} Jointly learning scaling factors results in exploding loss rolling variance (red line), indicating severe training instability. 
  \textbf{(b) KL Divergence Trend:} The inclusion of scaling factors causes the KL divergence to spike significantly, suggesting the policy drifts uncontrollably from the distilled prior. 
  \textbf{(c) HPS v2.1 Score \& (d) Aesthetic Score:} While freezing the scaling factors (blue line) leads to consistent improvements, optimizing them results in performance collapse and fluctuating reward scores.}
    \label{fig:scale_factor}
        \vspace{-1mm}
\end{figure*}
\begin{figure*}[t]
    \centering
\includegraphics[width=1\textwidth]{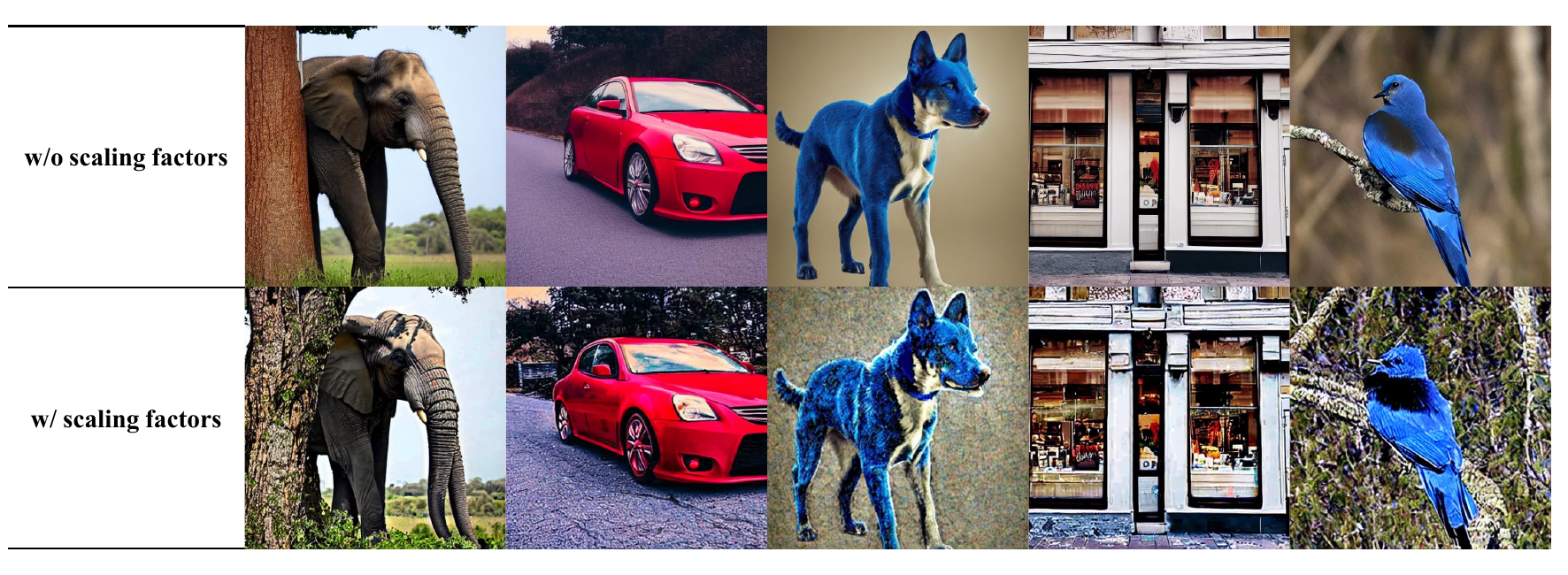}
    \caption{\textbf{Visual Impact of Reward Hacking via Scaling Factors.} We compare the generated image with and without optimizing the scaling factors ($o_n$ and $\delta_n^k$) in the RL stage. \textbf{w/o scaling factors:} Images generated without optimizing scaling factors maintain natural colors, realistic textures, and smooth lighting. \textbf{w/ scaling factors:} Images generated with scaling factor optimization exhibit classic reward hacking artifacts, including extreme color saturation, high-contrast distortion, and unnatural grainy textures.}
    \label{fig:scale}
        \vspace{-1mm}
\end{figure*}
\begin{figure*}[th]
    \centering
\includegraphics[width=1.0\textwidth]{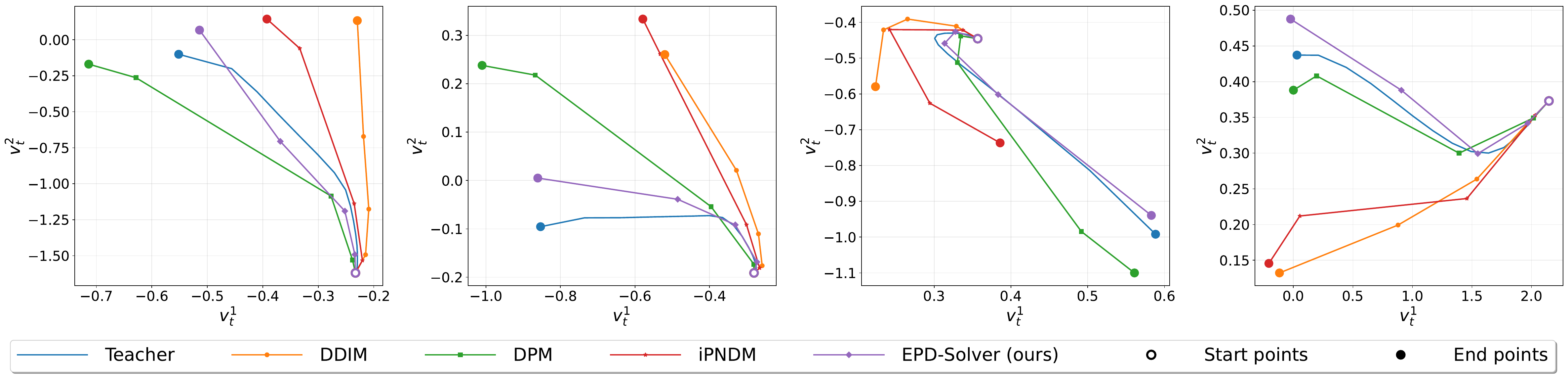}
    \caption{Analysis on local sampling trajectory. The figure shows the generation path of two randomly selected pixels in the images. We employ the  EPD ($\text{Para. NFE}=5, K = 2$) sampler for sampling, using the trajectory of its teacher sampler as the target trajectory. 
     We present the sampling trajectories with $\text{NFE}=5$ of DDIM \cite{songdenoising}, DPM-Solver \cite{lu2022dpm}, and iPNDM \cite{zhangfast} on CIFAR-10 \cite{krizhevsky2009learning}.
    } 
    \label{fig:traj}
    % \vspace{-4mm}
\end{figure*}

\begin{figure*}[t!]
    \centering
\includegraphics[width=1.0\textwidth]{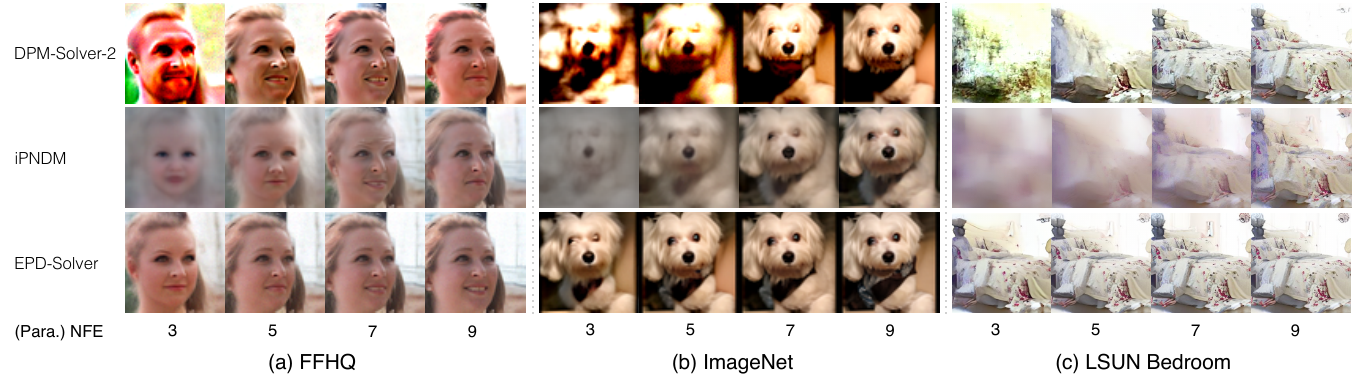}
    \caption{Comparison of generated samples among DPM-Solver-2 \cite{lu2022dpm}, iPNDM \cite{zhangfast} and $\ours$. Compared to other samplers, EPD-Solver achieves high-quality results even at NFE = 3. Additional visualizations are provided in~\cref{sec:visualization}.} 
    \label{Qualitative_results}
\end{figure*}

\noindent \textbf{Instability of learnable scaling factors in Stage 2.}
We freeze the scaling factors in Stage 2 to mitigate training instability. 
As shown in Fig.~\ref{fig:scale_factor}, jointly optimizing these factors leads to exploding loss variance and sharp KL spikes (Fig.~\ref{fig:scale_factor} (a)-(b)). 
Notably, although this setting achieves a higher peak HPS score, it results in severe degradation of aesthetic quality and eventual performance collapse (Fig.~\ref{fig:scale_factor} (c)-(d), Fig.~\ref{fig:scale}). 
Therefore, freezing these factors restricts the optimization to low-dimensional Dirichlet parameters, ensuring stable convergence and robust alignment.

\subsection{Qualitative Analyses of validation experiments}\label{sec:qualitative}

\noindent\textbf{Qualitative results on trajectory.} Since visualizing the trajectories of high-dimensional data is challenging, we adopt the analysis framework in \cite{liupseudo}. Specifically, as shown in ~\cref{fig:traj}, we randomly select two pixels from the images to perform local trajectory visualization, illustrating how their values evolve during the sampling process. Given the sampling $ \rvx_{t_N}, \rvx_{t_{N-1}}, \dots, \rvx_{t_0} $, we track the corresponding values $ v_{t}^1 $ and $ v_{t}^2 $ at two randomly chosen positions $ p_1 $ and $ p_2 $. We then represent \( (v_{t}^1, v_{t}^2) \) as data points and visualize them in \( \mathbb{R}^2 \). We can clearly observe that the pixel value trajectories of \ours~($\text{Para. NFE}=5 ,K=2$) are closer to the target trajectories compared to other samplers. This shows that our \ours~can generate more accurate trajectory, significantly reducing errors in the sampling process.

\section{Additional Experimental Results}

\begin{table*}[t!]
\small 
\captionsetup[subfloat]{labelformat=simple, labelsep=space}
\caption{Optimized Parameters for \ours~($K=2$) on CIFAR10 and FFHQ.}
\begin{minipage}[t]{0.48\textwidth}
    \fontsize{8}{10}\selectfont
     \setlength{\tabcolsep}{4pt} 
    \subfloat[\textbf{CIFAR10} $32 \times 32$ \cite{krizhevsky2009learning}]{
    \centering
        \begin{tabular}{cccccccc}
            \toprule
            Para. NFE &FID& $n$ & $k$ & $r_n^k$ & $s_n^k$ & $\sigma_n^k$ & $\lambda_n^k$ \\
            \midrule
            \multirow{4.5}{*}{3} & \multirow{4.5}{*}{10.40}  
            & \multirow{2}{*}{0} & 0 & 0.01339 & 0.96349 & 0.99731 & 0.85185 \\
            &&  & 1 & 0.67921 & 0.95231 & 0.99754 & 0.14815 \\ \cmidrule{3-8}
            && \multirow{2}{*}{1} & 0 & 0.10020 & 1.03590 & 0.99500 & 0.75008 \\
            &&  & 1 & 0.28855 & 0.95457 & 1.02139 & 0.24992 \\
            \midrule
            \multirow{7}{*}{5} & \multirow{7}{*}{4.33} 
            &\multirow{2}{*}{0} & 0 & 0.03333 & 0.95415 & 0.99735 & 0.86941 \\
            && & 1 & 0.79558 & 0.95376 & 0.98616 & 0.13059 \\ \cmidrule{3-8}
            && \multirow{2}{*}{1} & 0 & 0.07587 & 1.04503 & 0.99400 & 0.41741 \\
            &&  & 1 & 0.63244 & 1.04331 & 1.00711 & 0.58259 \\
            \cmidrule{3-8}
            && \multirow{2}{*}{2} & 0 & 0.38699 & 0.95588 & 1.00299 & 0.22410 \\
            &&  & 1 & 0.09434 & 1.01795 & 0.99999 & 0.77590 \\
            \midrule
            \multirow{9.5}{*}{7}& \multirow{9.5}{*}{2.82} 
            & \multirow{2}{*}{0} & 0 & 0.02511 & 0.96016 & 0.99725 & 0.86908 \\
            &&  & 1 & 0.91820 & 0.95206 & 1.01268 & 0.13092 \\ \cmidrule{3-8}
            && \multirow{2}{*}{1} & 0 & 0.27815 & 0.98792 & 0.98996 & 0.80595 \\
            &&  & 1 & 0.81671 & 0.99280 & 1.01571 & 0.19405 \\
            \cmidrule{3-8}
            &&  \multirow{2}{*}{2} & 0 & 0.34431 & 1.03617 & 0.99038 & 0.17049 \\
            &&  & 1 & 0.60552 & 1.03999 & 0.98517 & 0.82951 \\
            \cmidrule{3-8}
            &&  \multirow{2}{*}{3} & 0 & 0.09416 & 1.01655 & 1.00019 & 0.77621 \\
            &&  & 1 & 0.41999 & 0.96088 & 1.00966 & 0.22379 \\
            \midrule
            \multirow{12}{*}{9} & \multirow{12}{*}{2.49} 
            &\multirow{2}{*}{0} & 0 & 0.28390 & 0.96336 & 0.99459 & 0.74143 \\
            &&  & 1 & 0.08408 & 1.01058 & 0.99785 & 0.25857 \\
            \cmidrule{3-8}
            && \multirow{2}{*}{1} & 0 & 0.33981 & 0.97201 & 0.99713 & 0.31062 \\
            &&  & 1 & 0.47617 & 0.98810 & 1.00195 & 0.68938 \\
            \cmidrule{3-8}
            &&  \multirow{2}{*}{2} & 0 & 0.61703 & 1.03201 & 0.99898 & 0.79387 \\
            &&  & 1 & 0.12204 & 1.01552 & 0.98848 & 0.20613 \\
            \cmidrule{3-8}
            && \multirow{2}{*}{3} & 0 & 0.58062 & 1.02698 & 0.99284 & 0.90470 \\
            &&  & 1 & 0.31738 & 1.02504 & 0.98079 & 0.09530 \\
            \cmidrule{3-8}
            && \multirow{2}{*}{4} & 0 & 0.08719 & 0.98858 & 0.99555 & 0.77554 \\
            &&  & 1 & 0.44045 & 0.97831 & 1.02114 & 0.22446 \\
            \bottomrule
        \end{tabular}
        }
\end{minipage}\hfill
\begin{minipage}[t]{0.48\textwidth}
    \fontsize{8}{10}\selectfont
     \setlength{\tabcolsep}{4pt} 
        \centering
    \subfloat[ \textbf{FFHQ} $64 \times 64$ \cite{karras2019style}]{
        \begin{tabular}{cccccccc}
            \toprule
            Para. NFE &FID& $n$ & $k$ & $r_n^k$ & $s_n^k$ & $\sigma_n^k$ & $\lambda_n^k$ \\
            \midrule
            \multirow{4.5}{*}{3} & \multirow{4.5}{*}{21.74}& \multirow{2}{*}{0} & 0 & 0.00472 & 0.95251 & 0.99909 & 0.85527\\
            &&  & 1 & 0.61291 & 0.95212 & 1.00128 & 0.14473 \\
            \cmidrule{3-8}
            && \multirow{2}{*}{1} & 0 & 0.14636  & 1.00077 & 0.99866 & 0.90603 \\
            &&  & 1 & 0.52375 & 1.03973 &  1.00627 & 0.09397 \\
            \midrule
            \multirow{7}{*}{5} & \multirow{7}{*}{7.84} 
            &\multirow{2}{*}{0} & 0 & 0.00761 & 0.95240 & 0.98863 & 0.85668 \\
            &&  & 1 & 0.68196 & 0.95138 & 1.02573 & 0.14332 \\
            \cmidrule{3-8}
            && \multirow{2}{*}{1} & 0 & 0.48364 & 1.04868 & 1.01419 & 0.98053 \\
            &&  & 1 & 0.19897 & 1.03808 & 1.02313 &  0.01947\\
            \cmidrule{3-8}
            && \multirow{2}{*}{2} & 0 & 0.51289 &  1.01520 &  0.99043 & 0.12838 \\
            &&  & 1 & 0.12570 & 0.96696 & 0.99892 & 0.87162 \\
            \midrule
            \multirow{9.5}{*}{7} & \multirow{9.5}{*}{4.81} 
            &\multirow{2}{*}{0} & 0 & 0.00344 & 0.95175 & 0.99173 & 0.89005 \\
            &&  & 1 & 0.90422 & 0.95040 & 1.01825 & 0.10995 \\
            \cmidrule{3-8}
            && \multirow{2}{*}{1} & 0 & 0.61922 & 1.03974 & 0.99767 & 0.62252 \\
            &&  & 1 & 0.06710 & 1.03036 & 1.00397 & 0.37748 \\
            \cmidrule{3-8}
            && \multirow{2}{*}{2} & 0 & 0.36516 & 1.03981 & 1.01085 & 0.49539\\
            & &  & 1 & 0.71102 & 1.03331 & 1.01083 & 0.50461 \\ 
            \cmidrule{3-8}
            && \multirow{2}{*}{3} & 0 & 0.51302 & 0.99448 & 1.02493 & 0.15205 \\
            &&  & 1 & 0.11444 & 0.96889 & 0.99995 & 0.84795 \\
            \midrule
            \multirow{12}{*}{9} & \multirow{12}{*}{3.82}  
            & \multirow{2}{*}{0}& 0 & 0.07802 & 0.95010 & 0.99990 & 0.16419 \\
            &&  & 1 & 0.08710 & 0.95008 & 0.99990 & 0.83581 \\
            \cmidrule{3-8}
            && \multirow{2}{*}{1} & 0 & 0.85788 & 0.99068 & 0.98106 & 0.00087 \\
            &&  & 1 & 0.51685 & 0.99149 & 0.99980 & 0.99913 \\
            \cmidrule{3-8}
            && \multirow{2}{*}{2} & 0 & 0.5361 & 1.01276 & 0.99527 & 0.68458 \\
            &&  & 1 & 0.49629 & 1.01888 & 0.99385 & 0.31542 \\
            \cmidrule{3-8}
            && \multirow{2}{*}{3} & 0 & 0.55543 & 1.00901 & 1.00370 & 0.83477 \\
            &&  & 1 & 0.95208 & 1.01405& 1.00179 & 0.16523 \\
            \cmidrule{3-8}
            && \multirow{2}{*}{4} & 0 & 0.10233 & 0.95959 & 0.99459 & 0.85282  \\
            &&  & 1 & 0.53488 & 1.03980 & 1.04863 & 0.14718 \\
            \bottomrule
        \end{tabular}
        }
\end{minipage}

\label{tab:optimized_parameters_all_side_by_side}
\end{table*}
\begin{table*}[t!]
\small
\captionsetup[subfloat]{labelformat=simple, labelsep=space}
\caption{Optimized Parameters for \ours~($K=2$) on ImageNet and LSUN Bedroom.}
\begin{minipage}[t]{0.48\textwidth}
    \fontsize{8}{10}\selectfont
         \setlength{\tabcolsep}{4pt} 
        \subfloat[\textbf{ImageNet} $64 \times 64$ \cite{russakovsky2015imagenet}]{
        \centering
        \begin{tabular}{cccccccc}
            \toprule
            Para. NFE &FID& $n$ & $k$ & $r_n^k$ & $s_n^k$ & $\sigma_n^k$ & $\lambda_n^k$ \\
            \midrule
            \multirow{4.5}{*}{3} & \multirow{4.5}{*}{18.28} & \multirow{2}{*}{0} & 0 & 0.03892 & 0.90820 & 0.99810 & 0.78701 \\
            &&  & 1 & 0.58080 & 0.95077 & 1.00097 & 0.21299 \\ 
            \cmidrule{3-8}
            && \multirow{2}{*}{1} & 0 & 0.18326 & 0.99336 & 0.99910 & 0.97757 \\
            &&  & 1 & 0.08246 & 1.01142 & 1.02640 & 0.02243 \\
            \midrule
            \multirow{7}{*}{5}  & \multirow{7}{*}{6.35} & \multirow{2}{*}{0} & 0 &  0.14336 & 0.90835 & 0.99266 & 0.78550 \\
            &&  & 1 & 0.54204 & 0.93916 & 0.99114 &  0.21450 \\
            \cmidrule{3-8}
            && \multirow{2}{*}{1} & 0 & 0.71830 & 1.08078 & 1.00955 & 0.49788 \\
            &&  & 1 & 0.39094 & 1.07179 & 1.01071 & 0.50212 \\
            \cmidrule{3-8}
            && \multirow{2}{*}{2} & 0 & 0.25820 & 0.96964 & 1.00597 & 0.37857 \\
            &&  & 1 & 0.10124 & 1.00380 & 1.00316 & 0.62143 \\
            \midrule
            \multirow{9.5}{*}{7} & \multirow{9.5}{*}{5.26} & \multirow{2}{*}{0} & 0 & 0.11952 & 0.90686 & 0.99347 & 0.91217 \\
            &&  & 1 & 0.95726 & 0.91100 & 1.01887 & 0.08783 \\
            \cmidrule{3-8}
            && \multirow{2}{*}{1} & 0 & 0.41813 & 1.03421 & 0.99877 & 0.83649 \\
            &&  & 1 & 0.76716 & 1.04605 & 1.00396 & 0.16351 \\
            \cmidrule{3-8}
            && \multirow{2}{*}{2} & 0 & 0.86120 & 1.03538 & 1.00931 & 0.02866 \\
            &&  & 1 & 0.52961 & 1.04485 & 1.00040 & 0.97134 \\
            \cmidrule{3-8}
            && \multirow{2}{*}{3} & 0 & 0.19129 & 0.98157 & 1.0024 & 0.99873 \\
            &&  & 1 & 0.17888 & 0.99072 & 1.02263 & 0.00127 \\
            \midrule
            \multirow{12}{*}{9} & \multirow{12}{*}{4.27} & \multirow{2}{*}{0} & 0 & 0.97878 & 0.90410 & 1.01060 & 0.04239 \\
            &&  & 1 & 0.12206 & 0.90047 & 0.99891 & 0.95761 \\
            \cmidrule{3-8}
            && \multirow{2}{*}{1} & 0 & 0.40113 & 0.97924 & 0.99857 & 0.90324 \\
            &&  & 1 & 0.84037 & 1.04647 & 0.99850 & 0.09676 \\
            \cmidrule{3-8}
            && \multirow{2}{*}{2} & 0 & 0.55210 & 1.00744 & 0.99590 & 0.99983 \\
            &&  & 1 & 0.17699 & 0.97798 & 1.01484 & 0.00017 \\
            \cmidrule{3-8}
            && \multirow{2}{*}{3} & 0 & 0.67823 & 0.99619 & 1.01995 & 0.99919 \\
            &&  & 1 & 0.89296 & 1.02559 & 1.02289 & 0.00081 \\
            \cmidrule{3-8}
            && \multirow{2}{*}{4} & 0 & 0.26663 & 0.91395 & 1.01391 & 0.60252 \\
            &&  & 1 &  0.00584 & 1.06452 & 1.00333 & 0.39748 \\
            \bottomrule
        \end{tabular}
        }
\end{minipage}\hfill
\begin{minipage}[t]{0.48\textwidth}
    \fontsize{8}{10}\selectfont
     \setlength{\tabcolsep}{4pt} 
    \centering
        \subfloat[ \textbf{LSUN Bedroom} $256 \times 256$~\cite{yu2015lsun}]{
        \centering
        \begin{tabular}{cccccccc}
            \toprule
            Para. NFE &FID& $n$ & $k$ & $r_n^k$ & $s_n^k$ & $\sigma_n^k$ & $\lambda_n^k$ \\
            \midrule
            \multirow{4.5}{*}{3} & \multirow{4.5}{*}{13.21} & \multirow{2}{*}{0} & 0 & 0.82995 & 0.98769 & 1.01204 & 0.09938 \\
            &&  & 1 & 0.0410 & 1.0101 & 0.9989 & 0.9006 \\
            \cmidrule{3-8}
            && \multirow{2}{*}{1} & 0 & 0.03654 & 1.00350 & 0.98716 & 0.01419 \\
            &&  & 1 & 0.22279 & 0.97061 & 1.00927 & 0.98581 \\
            \midrule
            \multirow{7}{*}{5} & \multirow{7}{*}{7.52} & \multirow{2}{*}{0} & 0 & 0.99712 & 1.00000 & 0.99752 & 0.07831 \\
            &&  & 1 & 0.02895 & 1.00000 & 1.00046 & 0.92169 \\
            \cmidrule{3-8}
            && \multirow{2}{*}{1} & 0 & 0.52144 & 1.00000 & 1.00186 & 0.61657 \\
            &&  & 1 & 0.18287 & 1.00000 & 0.99460 & 0.38343 \\
            \cmidrule{3-8}
            && \multirow{2}{*}{2} & 0 & 0.20350 & 1.00000 & 0.96961 & 0.24707 \\
            &&  & 1 & 0.23099 & 1.00000 & 1.00159 & 0.75293 \\
            \midrule
            \multirow{9.5}{*}{7} & \multirow{9.5}{*}{5.97} & \multirow{2}{*}{0} & 0 & 0.92247 & 1.00000 & 1.00783 & 0.00004 \\
            &&  & 1 & 0.02283 & 1.00000 & 0.99966 & 1.00000 \\
            \cmidrule{3-8}
            && \multirow{2}{*}{1} & 0 & 0.45881 & 1.00000 & 1.00193 & 0.46663 \\
            &&  & 1 & 0.54699 & 1.00000 & 1.00185 & 0.53337 \\
            \cmidrule{3-8}
            && \multirow{2}{*}{2} & 0 & 0.09864 & 1.00000 & 0.98422 & 0.06541 \\
            &&  & 1 & 0.46885 & 1.00000 & 0.99675 & 0.93459 \\
            \cmidrule{3-8}
            && \multirow{2}{*}{3} & 0 & 0.20864 & 1.00000 & 0.96134 & 0.98301 \\
            &&  & 1 & 0.09425 & 1.00000 & 1.02840 & 0.01699 \\
            \midrule
            \multirow{12}{*}{9} & \multirow{12}{*}{5.01}& \multirow{2}{*}{0} & 0 & 0.87854 & 1.00000 & 1.00569 & 0.07317 \\
            &&  & 1 & 0.07964 & 1.00000 & 0.99953 & 0.92683 \\
            \cmidrule{3-8}
            && \multirow{2}{*}{1} & 0 & 0.40848 & 1.00000 & 0.99842 & 0.82916 \\
            &&  & 1 & 0.94301 & 1.00000 & 1.00355 & 0.17084 \\
            \cmidrule{3-8}
            && \multirow{2}{*}{2} & 0 & 0.67654 & 1.00000 & 1.00375 & 0.01636 \\
            &&  & 1 & 0.49911 & 1.00000 & 1.00348 & 0.98364 \\
            \cmidrule{3-8}
            && \multirow{2}{*}{3} & 0 & 0.45169 & 1.00000 & 0.98647 & 0.14504 \\
            &&  & 1 & 0.40655 & 1.00000 & 0.99226 & 0.85496 \\
            \cmidrule{3-8}
            && \multirow{2}{*}{4} & 0 & 0.30053 & 1.00000 & 1.00438 & 0.02853 \\
            &&  & 1 & 0.20058 & 1.00000 & 0.95733 & 0.97147 \\
            \bottomrule
        \end{tabular}
        }
\end{minipage}

\label{tab:optimized_parameters_imagenet_lsun_side}
\end{table*}

\noindent\textbf{Other choice of intermediate points.}
In ~\cref{tab:midpoints_choice}, we compare our $\ours$ with $K=2$, \ie, two learned intermediate points, against two manually selected midpoints and randomly selected ones. In particular, the manually selected midpoints include the start timestep $t_n$, the end timestep $t_{n+1}$ (adopted in EDM), the geometric mean $\sqrt{t_n t_{n+1}}$ (used in DPM-Solver-2), and the arithmetic mean $\frac{1}{2}(t_n+t_{n+1})$. The random midpoints are uniformly sampled from $[t_n, t_{n+1}]$. We note several observations: (1) The combination of start points with mean points (geometric and arithmetic) significantly outperforms combinations that include the end point. For example, using the geometric and arithmetic points achieves an FID of 5.83 with NFE = 9, whereas incorporating the end point leads to much higher FID scores — 44.38 and 32.21 for the geometric and arithmetic points, respectively. (2) The combination that includes random points achieves competitive results. For instance, using a random point together with the start point yields better FID scores than EDM across all NFE values. (3) The gap between the best combination of handcrafted intermediate timesteps and our learned ones remains large, highlighting the necessity of our proposed method.

\subsection{Optimized Parameters for \ours}\label{sec:learnedparameters}

We provide our optimized parameters of \ours~with $K=2$ for CIFAR-10, ImageNet, FFHQ and LSUN Bedroom in \cref{tab:optimized_parameters_all_side_by_side,tab:optimized_parameters_imagenet_lsun_side,tab:optimized_parameters_sd} with different Para. NFEs. 
According to the implementation details in~\cref{sec:implement-epd}, the parameters $\tau_n^k ,\delta_n^k,o_n$ are derived as follows:
\begin{align}
    \tau_n^k&=t_{n+1}^{r_n^k} \cdot t_n^{1 - r_n^k} \\
    \delta_n^k&=(s_n^k-1)\tau_n^k \\ 
    o_n&=\sum_k \lambda_n^k\sigma_n^k - 1
\end{align}

\begin{table*}[t!]
\small 
\captionsetup[subfloat]{labelformat=simple, labelsep=space}
\caption{Optimized Parameters for \oursplugin~($K=2$) on CIFAR10 and FFHQ.}
\begin{minipage}[t]{0.48\textwidth}
    \fontsize{8}{10}\selectfont
     \setlength{\tabcolsep}{4pt} 
    \subfloat[\textbf{CIFAR10} $32 \times 32$ \cite{krizhevsky2009learning}]{
    \centering
        \begin{tabular}{cccccccc}
            \toprule
            Para. NFE &FID& $n$ & $k$ & $r_n^k$ & $s_n^k$ & $\sigma_n^k$ & $\lambda_n^k$ \\
            \midrule
            \multirow{4.5}{*}{3} & \multirow{4.5}{*}{10.54}  
            & \multirow{2}{*}{0} & 0 & 0.06837 & 0.81145 &0.99957  & 0.91271 \\
            &&  & 1 & 0.68803 & 0.85836 & 0.99981 & 0.08729 \\ \cmidrule{3-8}
            && \multirow{2}{*}{1} & 0 & 0.12320 & 0.97533 & 0.99903 & 0.85072 \\
            &&  & 1 & 0.28206 & 0.85043 & 1.00671 & 0.14928 \\
            \midrule
            \multirow{7}{*}{5} & \multirow{7}{*}{4.47} 
            &\multirow{2}{*}{0} & 0 & 0.10548 & 0.80808 & 0.99606 & 0.95656 \\
            && & 1 & 0.96750 & 0.89210 & 1.00082 & 0.04344 \\ \cmidrule{3-8}
            && \multirow{2}{*}{1} & 0 & 0.04114 & 1.03816 & 1.00480 & 0.52907 \\
            &&  & 1 & 0.57891 & 1.02063 & 1.02490 & 0.47093 \\
            \cmidrule{3-8}
            && \multirow{2}{*}{2} & 0 & 0.27989 & 1.00150 & 0.95600 & 0.26331 \\
            &&  & 1 & 0.05394 & 1.02182 & 0.98523 & 0.73669 \\
            \midrule
            \multirow{9.5}{*}{7}& \multirow{9.5}{*}{3.27} 
            & \multirow{2}{*}{0} & 0 & 0.08991 & 0.80504 & 0.99845 & 0.94689 \\
            &&  & 1 & 0.94988 & 0.95487 & 1.01496 & 0.05311 \\ \cmidrule{3-8}
            && \multirow{2}{*}{1} & 0 & 0.04569 & 0.88770 & 0.99774 & 0.75623 \\
            &&  & 1 &  0.80305& 1.04391 & 0.99378 & 0.24377 \\
            \cmidrule{3-8}
            &&  \multirow{2}{*}{2} & 0 & 0.91959 & 1.10578 & 0.99989 & 0.00408 \\
            &&  & 1 & 0.42678 & 1.01745 & 1.00242 & 0.99592 \\
            \cmidrule{3-8}
            &&  \multirow{2}{*}{3} & 0 & 0.36480 & 0.90472 & 1.02327 &0.20787  \\
            &&  & 1 & 0.07649 & 0.96814 & 1.00433 & 0.79213 \\
            \midrule
            \multirow{12}{*}{9} & \multirow{12}{*}{2.42} 
            &\multirow{2}{*}{0} & 0 & 0.08244 & 0.80210 & 0.99483 & 0.08638 \\
            &&  & 1 & 0.25440 & 0.81528 & 0.99964 & 0.91362 \\
            \cmidrule{3-8}
            && \multirow{2}{*}{1} & 0 & 0.02193 & 0.80719 & 0.99517 & 0.99163 \\
            &&  & 1 & 0.02935  & 0.88719 & 0.99437 & 0.00837 \\
            \cmidrule{3-8}
            &&  \multirow{2}{*}{2} & 0 & 0.25227 & 1.08671 & 0.99438 & 0.02010 \\
            &&  & 1 & 0.55490 & 1.03722 & 0.99923 & 0.97990 \\
            \cmidrule{3-8}
            && \multirow{2}{*}{3} & 0 & 0.48861 & 1.01472 & 1.00312 & 0.81266 \\
            &&  & 1 & 0.02553 & 0.98693 &  1.00521& 0.18734 \\
            \cmidrule{3-8}
            && \multirow{2}{*}{4} & 0 & 0.07257 & 0.97384 & 0.99552 & 0.78925 \\
            &&  & 1 & 0.39513  & 0.96933 & 0.99003 & 0.21075 \\
            \bottomrule
        \end{tabular}
        }
\end{minipage}\hfill
\begin{minipage}[t]{0.48\textwidth}
    \fontsize{8}{10}\selectfont
     \setlength{\tabcolsep}{4pt} 
        \centering
    \subfloat[ \textbf{FFHQ} $64 \times 64$ \cite{karras2019style}]{
        \begin{tabular}{cccccccc}
            \toprule
            Para. NFE &FID& $n$ & $k$ & $r_n^k$ & $s_n^k$ & $\sigma_n^k$ & $\lambda_n^k$ \\
            \midrule
            \multirow{4.5}{*}{3} &\multirow{4.5}{*}{19.02}& \multirow{2}{*}{0} & 0 & 0.07642 & 0.84410 & 0.99934 & 0.94986 \\
            &&  & 1 & 0.91510 & 0.97713 & 1.01079 & 0.05014 \\
            \cmidrule{3-8}
            && \multirow{2}{*}{1} & 0 & 0.17864 & 0.97337 & 1.00023 & 0.99041 \\
            &&  & 1 & 0.15293 & 0.90787 & 1.02719 & 0.00959 \\
            \midrule
            \multirow{7}{*}{5} & \multirow{7}{*}{7.97}  &\multirow{2}{*}{0} & 0 & 0.00858 & 0.82007 & 0.99986 & 0.87461 \\
            &&  & 1 & 0.65658 & 0.86946 & 0.99954 & 0.12539 \\
            \cmidrule{3-8}
            && \multirow{2}{*}{1} & 0 & 0.39945 & 0.99765 & 1.00157 & 0.99812 \\
            &&  & 1 & 0.18867 & 1.03054 & 1.01357 & 0.00188 \\
            \cmidrule{3-8}
            && \multirow{2}{*}{2} & 0 & 0.33148 & 0.96555 & 0.99766 & 0.22642 \\
            &&  & 1 & 0.07594 & 0.97690 & 0.99730 & 0.77358 \\
            \midrule
            \multirow{9.5}{*}{7} &\multirow{9.5}{*}{5.09}  &\multirow{2}{*}{0} & 0 & 0.01069 & 0.81532 & 0.99965 & 0.92015 \\
            &&  & 1 & 0.85634 & 0.86078 & 0.99965 & 0.07985 \\
            \cmidrule{3-8}
            && \multirow{2}{*}{1} & 0 & 0.37517 & 1.00369 & 0.99838 & 0.88685 \\
            &&  & 1 & 0.71151 & 1.00119 & 1.00481 & 0.11315 \\
            \cmidrule{3-8}
            && \multirow{2}{*}{2} & 0 & 0.08475 & 1.04325 & 1.03287 & 0.00052 \\
            & &  & 1 & 0.38954 & 1.00524 & 1.00463 & 0.99948 \\ 
            \cmidrule{3-8}
            && \multirow{2}{*}{3} & 0 & 0.08461 & 0.98373 & 0.98399 & 0.76003 \\
            &&  & 1 & 0.39386 & 1.01515 & 0.97975 & 0.23997 \\
            \midrule
            \multirow{12}{*}{9} & \multirow{12}{*}{3.53} &\multirow{2}{*}{0} & 0 & 0.94960 & 0.82963 & 1.00126 & 0.06572 \\
            &&  & 1 & 0.00362 & 0.82194 & 0.9998 & 0.93428 \\
            \cmidrule{3-8}
            && \multirow{2}{*}{1} & 0 & 0.06822 & 0.87369 & 0.99903 & 0.19003 \\
            &&  & 1 & 0.48656 & 1.01113 & 0.99772 & 0.80995 \\
            \cmidrule{3-8}
            && \multirow{2}{*}{2} & 0 & 0.38262 & 1.02269 & 0.99920 & 0.84123 \\
            &&  & 1 & 0.98681 & 0.99794 & 1.01047 & 0.15877 \\
            \cmidrule{3-8}
            && \multirow{2}{*}{3} & 0 & 0.08146 & 0.99005 & 1.01881 & 0.56715 \\
            &&  & 1 & 0.89689 & 1.01201 & 0.99138 & 0.43285 \\
            \cmidrule{3-8}
            && \multirow{2}{*}{4} & 0 & 0.07455 & 0.96557 & 0.97884 & 0.80133 \\
            &&  & 1 & 0.47558 & 1.09918 & 0.95222 & 0.19867 \\
            \bottomrule
        \end{tabular}
        }
\end{minipage}
\label{tab:epd-plugin-cifar-ffhq}
\end{table*}
\begin{table*}[t!]
\small
\captionsetup[subfloat]{labelformat=simple, labelsep=space}
\caption{Optimized Parameters for \oursplugin~($K=2$) on ImageNet and LSUN Bedroom.}
\begin{minipage}[t]{0.48\textwidth}
    \fontsize{8}{10}\selectfont
         \setlength{\tabcolsep}{4pt} 
        \subfloat[\textbf{ImageNet} $64 \times 64$ \cite{russakovsky2015imagenet}]{
        \centering
        \begin{tabular}{cccccccc}
            \toprule
            Para. NFE &FID& $n$ & $k$ & $r_n^k$ & $s_n^k$ & $\sigma_n^k$ & $\lambda_n^k$ \\
            \midrule
            \multirow{4.5}{*}{3} & \multirow{4.5}{*}{19.89} 
            & \multirow{2}{*}{0} & 0 & 0.01805  & 0.89265  & 0.99984 & 0.81070 \\
            &&  & 1 & 0.59732  & 0.95910 & 0.99862  & 0.18930\\ 
            \cmidrule{3-8}
            && \multirow{2}{*}{1} & 0 & 0.15989 &0.96659  &1.00771  & 0.96197\\
            &&  & 1 &  0.26658 & 0.89747 & 1.04079 & 0.03803\\
            \midrule
            \multirow{7}{*}{5}  & \multirow{7}{*}{8.17} 
            & \multirow{2}{*}{0} & 0 & 0.11246 & 0.82261 & 0.99876 & 0.92199\\
            &&  & 1 & 0.92205  & 0.96191 & 1.01100 & 0.07801 \\
            \cmidrule{3-8}
            && \multirow{2}{*}{1} & 0 & 0.00511  &0.97233  &0.99878 & 0.45635\\
            &&  & 1 & 0.61007 &  0.99912  & 1.00419 & 0.54365\\
            \cmidrule{3-8}
            && \multirow{2}{*}{2} & 0 & 0.35416 & 0.92432 & 0.99057 & 0.04391\\
            &&  & 1 & 0.13234  &  0.96354  &  0.99885 & 0.95609  \\
            \midrule
            \multirow{9.5}{*}{7} & \multirow{9.5}{*}{4.81} 
            & \multirow{2}{*}{0} & 0 & 0.14306  & 0.82532 &0.99963  & 0.99640 \\
            &&  & 1 & 0.02764 & 0.94802  & 0.96580& 0.00360\\
            \cmidrule{3-8}
            && \multirow{2}{*}{1} & 0 & 0.46578 & 0.98602 &1.00224 & 0.99615\\
            &&  & 1 &  0.09086 &  1.08617  &1.02104 & 0.00385\\
            \cmidrule{3-8}
            && \multirow{2}{*}{2} & 0 & 0.04504 & 1.05987 &  1.01408 & 0.00020\\
            &&  & 1  & 0.44154 &  0.99292  &  0.99536 & 0.99980 \\
            \cmidrule{3-8}
            && \multirow{2}{*}{3} & 0 & 0.03175 & 0.90298 & 0.98815  &  0.00276\\
            &&  & 1 & 0.14969 &  0.94543 &1.00853 &  0.99724 \\
            \midrule
            \multirow{12}{*}{9} & \multirow{12}{*}{4.02} & \multirow{2}{*}{0} 
            & 0 & 0.33263 & 0.84332 & 0.99983 & 0.12259\\
            &&  & 1 &  0.13371  & 0.85792 & 0.99931  & 0.87741\\
            \cmidrule{3-8}
            && \multirow{2}{*}{1} & 0&  0.05410 & 0.89662 & 1.00055 &0.24089  \\
            &&  & 1 & 0.54876 & 0.99484  &  0.99886  & 0.75911 \\
            \cmidrule{3-8}
            && \multirow{2}{*}{2} & 0 & 0.37444 & 1.00578  &1.00105 & 0.88450\\
            &&  & 1 &  0.94384 &  1.01652  &0.98910 & 0.11550\\
            \cmidrule{3-8}
            && \multirow{2}{*}{3} & 0 & 0.28771 & 1.00243 &  0.99434 & 0.76097  \\
            &&  & 1 &  0.82883 & 1.00291 &  0.99311 & 0.23903\\
            \cmidrule{3-8}
            && \multirow{2}{*}{4} & 0 &0.11117  & 0.98196 & 1.01350& 0.80293\\
            &&  & 1 & 0.41243 & 0.88880 & 1.08111&  0.19707  \\
            \bottomrule
        \end{tabular}
        }
\end{minipage}\hfill
\begin{minipage}[t]{0.48\textwidth}
    \fontsize{8}{10}\selectfont
     \setlength{\tabcolsep}{4pt} 
    \centering
        \subfloat[ \textbf{LSUN Bedroom} $256 \times 256$~\cite{yu2015lsun}]{
        \centering
        \begin{tabular}{cccccccc}
            \toprule
            Para. NFE &FID& $n$ & $k$ & $r_n^k$ & $s_n^k$ & $\sigma_n^k$ & $\lambda_n^k$ \\
            \midrule
            \multirow{4.5}{*}{3} & \multirow{4.5}{*}{14.12} 
            & \multirow{2}{*}{0} & 0 & 0.78697 & 1.00000 & 1.00375 & 0.10230 \\
            &&  & 1 & 0.02085 & 1.00000 & 0.99945  & 0.89770\\ 
            \cmidrule{3-8}
            && \multirow{2}{*}{1} & 0 & 0.08334 & 1.00000 & 0.96782 &0.18352  \\
            &&  & 1 & 0.23899 &  1.00000 & 0.99524 & 0.81648\\
            \midrule
            \multirow{7}{*}{5}  & \multirow{7}{*}{8.26} 
            & \multirow{2}{*}{0} & 0 & 0.97220  & 0.98923 &1.00016  & 0.07808\\
            &&  & 1 & 0.03306 & 1.00415 & 0.99991 &0.92192  \\
            \cmidrule{3-8}
            && \multirow{2}{*}{1} & 0 & 0.52337 & 0.99607 & 1.00463 & 0.60203 \\
            &&  & 1 & 0.01602   & 1.00079 & 0.99249 &0.39797   \\
            \cmidrule{3-8}
            && \multirow{2}{*}{2} & 0 &  0.12524 & 0.99813  &  0.96174 & 0.49642  \\
            &&  & 1 & 0.29699 & 0.99950  & 1.01130  & 0.50358 \\
            \midrule
            \multirow{9.5}{*}{7} & \multirow{9.5}{*}{5.24} 
            & \multirow{2}{*}{0} & 0 & 0.97094 & 0.98527 & 1.01234 & 0.06101 \\
            &&  & 1 & 0.07156 & 1.00461 & 0.99893 & 0.93899\\
            \cmidrule{3-8}
            && \multirow{2}{*}{1} & 0 & 0.70513 & 0.99016 & 1.01166 & 0.32484 \\
            &&  & 1 & 0.24738 &  0.98946 & 0.99696 & 0.67516 \\
            \cmidrule{3-8}
            && \multirow{2}{*}{2} & 0 &  0.27565 & 1.01344 & 0.97876 & 0.57267 \\
            &&  & 1  &  0.54473 & 1.00123 & 1.00931 & 0.42733 \\
            \cmidrule{3-8}
            && \multirow{2}{*}{3} & 0 & 0.16616 & 0.98549 & 0.96569 & 0.85584 \\
            &&  & 1 & 0.38606 & 0.99734 & 1.02813 & 0.14416 \\
            \midrule
            \multirow{12}{*}{9} & \multirow{12}{*}{4.51} & \multirow{2}{*}{0} 
            & 0 & 0.17020 & 1.01750 & 0.99792 & 0.34563\\
            &&  & 1 & 0.01271 & 0.99479 &1.00060 & 0.65437 \\
            \cmidrule{3-8}
            && \multirow{2}{*}{1} & 0& 0.43953 & 0.98534 & 0.99969 & 0.96036\\
            &&  & 1 & 0.82230 & 0.99246  &0.99977 & 0.03964\\
            \cmidrule{3-8}
            && \multirow{2}{*}{2} & 0 &  0.25682 & 1.00056 &1.00433 &0.30549 \\
            &&  & 1 & 0.50732 & 1.00773 & 0.99838 & 0.69451\\
            \cmidrule{3-8}
            && \multirow{2}{*}{3} & 0 & 0.29627 & 1.01221 & 0.98564 &  0.31065  \\
            &&  & 1 & 0.48616 &  1.01091 & 0.99254 & 0.68935 \\
            \cmidrule{3-8}
            && \multirow{2}{*}{4} & 0 & 0.32949 & 1.00615 &0.98884 & 0.04682\\
            &&  & 1 & 0.19802 &  0.98760 & 0.95685 & 0.95318 \\
            \bottomrule
        \end{tabular}
        }
\end{minipage}
\label{tab:optimized_parameters_imagenet_lsun_plugin}
\end{table*}
\begin{table*}[t!]
\small
\captionsetup[subfloat]{labelformat=simple, labelsep=space}
\caption{Optimized Parameters for \ours~($K = 2$) on SD1.5, SD3-Medium (512 $\times$ 512), SD3-Medium (1024 $\times$ 1024)}
\label{tab:optimized_parameters_sd}

% ----------------------------------------------------------------------
% DATASET 1: SD1.5
% ----------------------------------------------------------------------
\begin{minipage}[t]{0.325\textwidth}
    \centering
    \setlength{\tabcolsep}{8pt} % Adjusted spacing
    \subfloat[\textbf{SD1.5}]{
    \begin{tabular}{cccc}
        \toprule
        $n$ & $k$ & $r_n^k$ & $\lambda_n^k$ \\
        \midrule
        \multirow{2}{*}{0} & 0 & 0.15270 & 0.89191 \\
                           & 1 & 0.61737 & 0.10809 \\
        \cmidrule{2-4}
        \multirow{2}{*}{1} & 0 & 0.31550 & 0.69758 \\
                           & 1 & 0.65520 & 0.30242 \\
        \cmidrule{2-4}
        \multirow{2}{*}{2} & 0 & 0.17886 & 0.63533 \\
                           & 1 & 0.73115 & 0.36467 \\
        \cmidrule{2-4}
        \multirow{2}{*}{3} & 0 & 0.71696 & 0.08420 \\
                           & 1 & 0.86882 & 0.91580 \\
        \cmidrule{2-4}
        \multirow{2}{*}{4} & 0 & 0.70392 & 0.06184 \\
                           & 1 & 0.79816 & 0.93816 \\
        \cmidrule{2-4}
        \multirow{2}{*}{5} & 0 & 0.18810 & 0.56545 \\
                           & 1 & 0.89286 & 0.43455 \\
        \cmidrule{2-4}
        \multirow{2}{*}{6} & 0 & 0.42211 & 0.42102 \\
                           & 1 & 0.77533 & 0.57898 \\
        \cmidrule{2-4}
        \multirow{2}{*}{7} & 0 & 0.29240 & 0.26946 \\
                           & 1 & 0.80529 & 0.73054 \\
        \cmidrule{2-4}
        \multirow{2}{*}{8} & 0 & 0.49096 & 0.13629 \\
                           & 1 & 0.83929 & 0.86371 \\
        \cmidrule{2-4}
        \multirow{2}{*}{9} & 0 & 0.26918 & 0.91604 \\
                           & 1 & 0.55223 & 0.08396 \\
        \bottomrule
    \end{tabular}
    }
\end{minipage}\hfill
% ----------------------------------------------------------------------
% DATASET 2: SD3-Medium (512x512)
% ----------------------------------------------------------------------
\begin{minipage}[t]{0.325\textwidth}
    \centering
    \setlength{\tabcolsep}{8pt}
    \subfloat[\textbf{SD3-Medium} ($512 \times 512$)]{
    \begin{tabular}{cccc}
        \toprule
        $n$ & $k$ & $r_n^k$ & $\lambda_n^k$ \\
        \midrule
        \multirow{2}{*}{0} & 0 & 0.13629 & 0.50848 \\
                           & 1 & 0.79492 & 0.49152 \\
        \cmidrule{2-4}
        \multirow{2}{*}{1} & 0 & 0.13325 & 0.35471 \\
                           & 1 & 0.84627 & 0.64529 \\
        \cmidrule{2-4}
        \multirow{2}{*}{2} & 0 & 0.16606 & 0.08037 \\
                           & 1 & 0.73666 & 0.91963 \\
        \cmidrule{2-4}
        \multirow{2}{*}{3} & 0 & 0.52475 & 0.88174 \\
                           & 1 & 0.93940 & 0.11826 \\
        \cmidrule{2-4}
        \multirow{2}{*}{4} & 0 & 0.73197 & 0.98594 \\
                           & 1 & 0.90343 & 0.01406 \\
        \cmidrule{2-4}
        \multirow{2}{*}{5} & 0 & 0.42911 & 0.46522 \\
                           & 1 & 0.76482 & 0.53478 \\
        \cmidrule{2-4}
        \multirow{2}{*}{6} & 0 & 0.48291 & 0.97987 \\
                           & 1 & 0.67723 & 0.02013 \\
        \cmidrule{2-4}
        \multirow{2}{*}{7} & 0 & 0.48970 & 0.70646 \\
                           & 1 & 0.68562 & 0.29354 \\
        \cmidrule{2-4}
        \multirow{2}{*}{8} & 0 & 0.41043 & 0.88591 \\
                           & 1 & 0.83159 & 0.11409 \\
        \cmidrule{2-4}
        \multirow{2}{*}{9} & 0 & 0.05957 & 0.51526 \\
                           & 1 & 0.15053 & 0.48474 \\
        \bottomrule
    \end{tabular}
    }
\end{minipage}\hfill
% ----------------------------------------------------------------------
% DATASET 3: SD3-Medium (1024x1024)
% ----------------------------------------------------------------------
\begin{minipage}[t]{0.325\textwidth}
    \centering
    \setlength{\tabcolsep}{8pt}
    \subfloat[\textbf{SD3-Medium} ($1024 \times 1024$)]{
    \begin{tabular}{cccc}
        \toprule
        $n$ & $k$ & $r_n^k$ & $\lambda_n^k$ \\
        \midrule
        \multirow{2}{*}{0} & 0 & 0.31763 & 0.43981 \\
                           & 1 & 0.76296 & 0.56019 \\
        \cmidrule{2-4}
        \multirow{2}{*}{1} & 0 & 0.50609 & 0.33906 \\
                           & 1 & 0.73776 & 0.66094 \\
        \cmidrule{2-4}
        \multirow{2}{*}{2} & 0 & 0.42443 & 0.12271 \\
                           & 1 & 0.81546 & 0.87729 \\
        \cmidrule{2-4}
        \multirow{2}{*}{3} & 0 & 0.23347 & 0.06469 \\
                           & 1 & 0.72259 & 0.93531 \\
        \cmidrule{2-4}
        \multirow{2}{*}{4} & 0 & 0.65061 & 0.33202 \\
                           & 1 & 0.82528 & 0.66798 \\
        \cmidrule{2-4}
        \multirow{2}{*}{5} & 0 & 0.32002 & 0.42047 \\
                           & 1 & 0.77396 & 0.57953 \\
        \cmidrule{2-4}
        \multirow{2}{*}{6} & 0 & 0.56253 & 0.64723 \\
                           & 1 & 0.78896 & 0.35277 \\
        \cmidrule{2-4}
        \multirow{2}{*}{7} & 0 & 0.57498 & 0.19995 \\
                           & 1 & 0.73481 & 0.80005 \\
        \cmidrule{2-4}
        \multirow{2}{*}{8} & 0 & 0.22396 & 0.41957 \\
                           & 1 & 0.69669 & 0.58043 \\
        \cmidrule{2-4}
        \multirow{2}{*}{9} & 0 & 0.14939 & 0.91541 \\
                           & 1 & 0.35963 & 0.08459 \\
        \bottomrule
    \end{tabular}
    }
\end{minipage}
\end{table*}
\subsection{Optimized Parameters for \oursplugin}
We provide our optimized parameters of \oursplugin~with $K=2$ for CIFAR10, ImageNet, FFHQ and LSUN Bedroom in~\cref{tab:epd-plugin-cifar-ffhq,tab:optimized_parameters_imagenet_lsun_plugin} with different Para.NFEs. 

\subsection{Additional Qualitative Results}\label{sec:visualization}
\begin{figure*}[t]
    \centering
\includegraphics[width=0.9\textwidth]{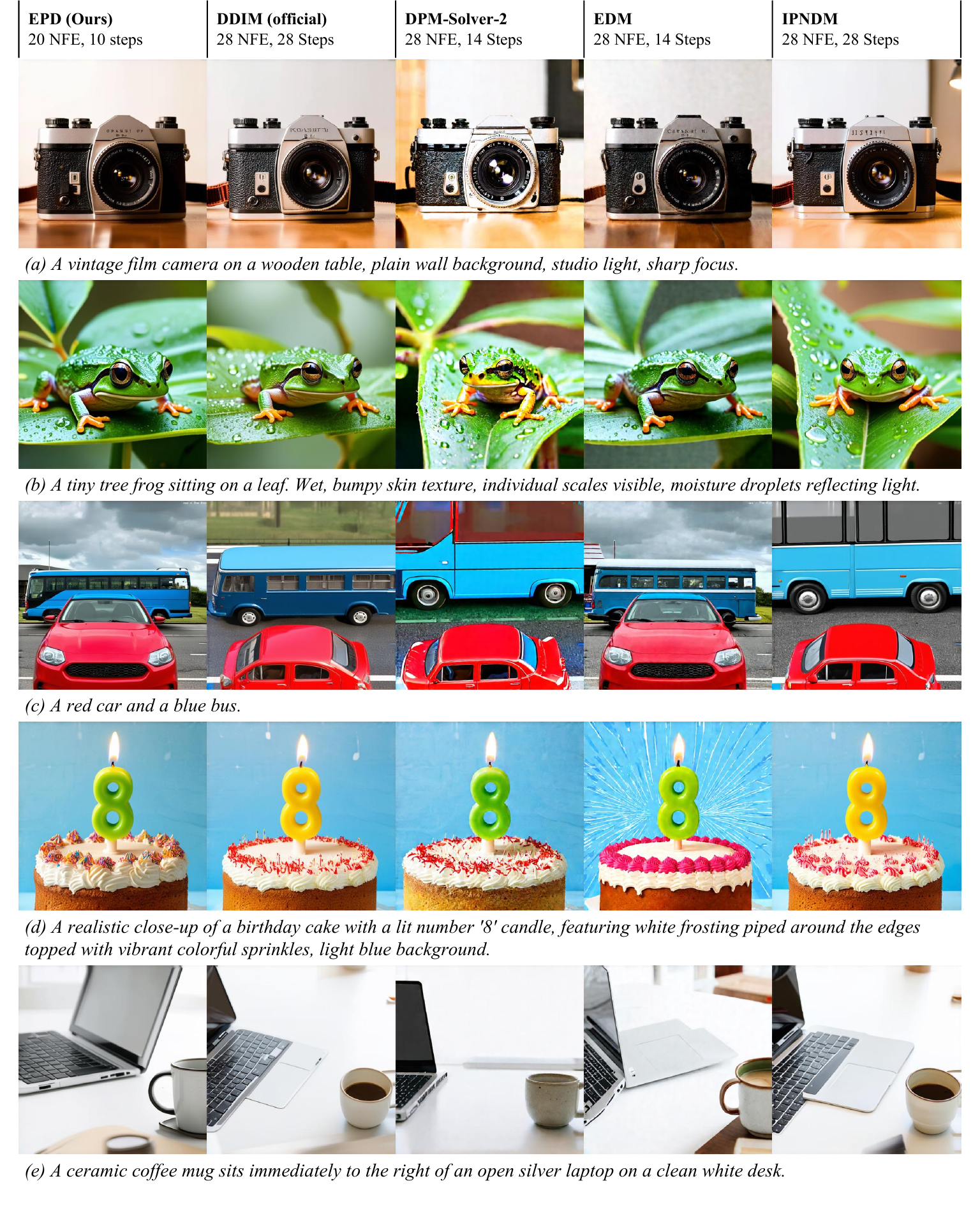}
    \caption{\textbf{Qualitative comparison of text-to-image generation results.} Samples are generated using \textbf{SD3-medium (512 $\times$ 512)}.}
    \label{fig:sd3_512}
        \vspace{-1mm}
\end{figure*}

\begin{figure*}[t]
    \centering
\includegraphics[width=0.9\textwidth]{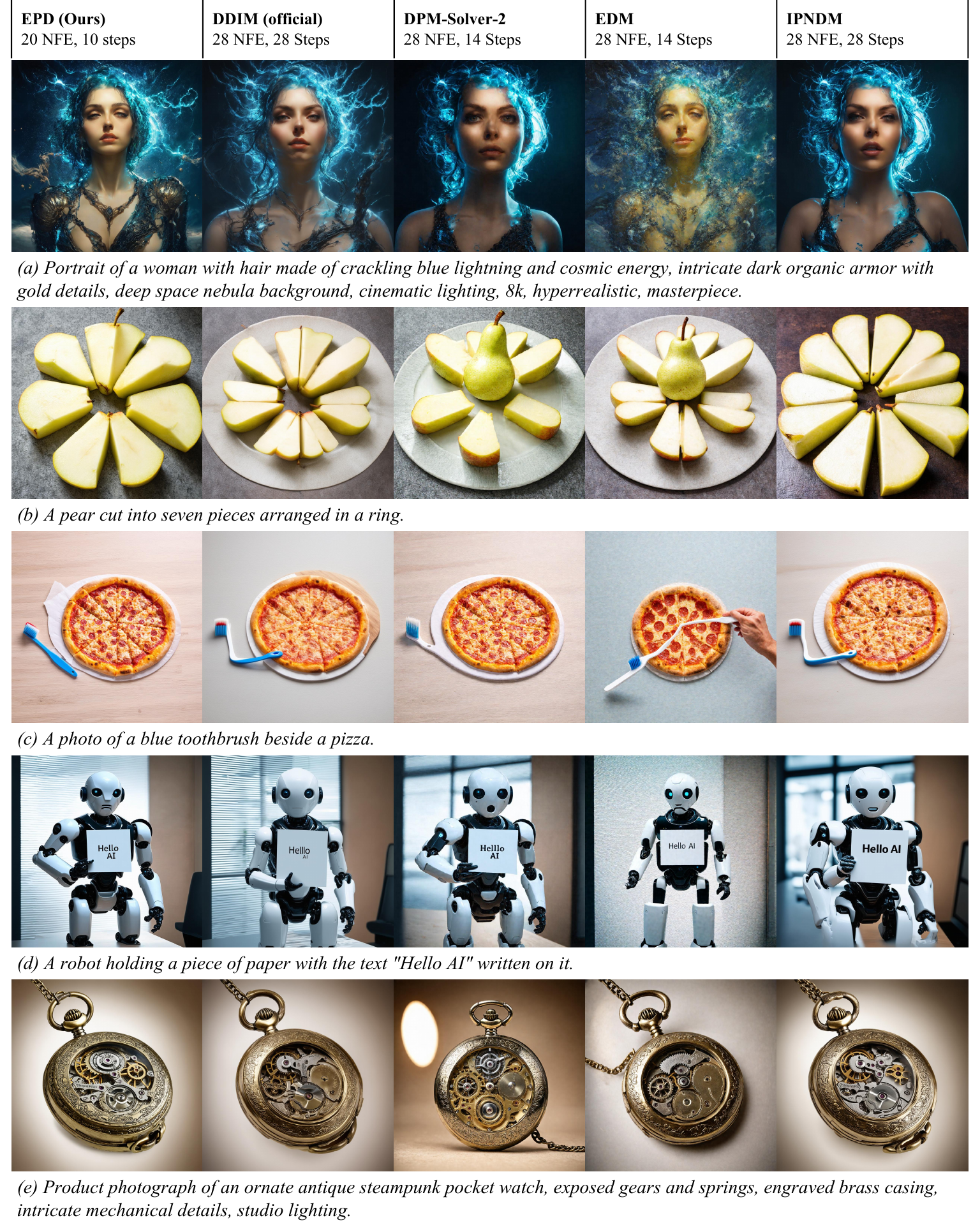}
    \caption{\textbf{Qualitative comparison of text-to-image generation results.} Samples are generated using \textbf{SD3-medium (1024 $\times$ 1024)}.}
    \label{fig:sd3_1024}
        \vspace{-1mm}
\end{figure*}

Here, we show some qualitative results on different datasets in \cref{Qualitative_results,fig:sup_grid_cifar10_3,fig:sup_grid_ffhq_3,fig:sup_grid_img_3,fig:sd3_512,fig:sd3_1024}.

\clearpage

% how to define 

\newcommand{\nfea}{\text{3}}
\newcommand{\nfeb}{\text{9}}

\begin{figure*}[t]
  \centering
  \begin{subfigure}[b]{0.48\linewidth}
      \includegraphics[width=\linewidth]{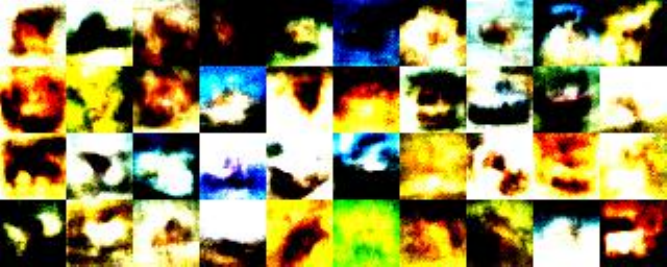}
      \caption{DPM-Solver-2. NFE=\(\nfea\)}
  \end{subfigure}
  \begin{subfigure}[b]{0.48\linewidth}
      \includegraphics[width=\linewidth]{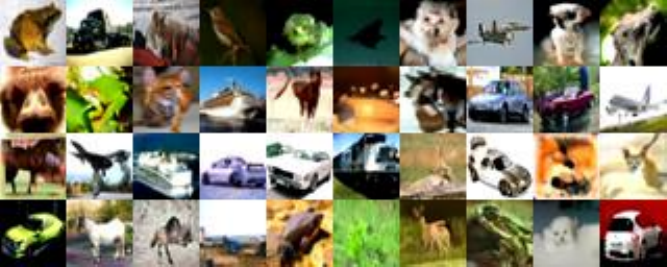}
      \caption{DPM-Solver-2. NFE=\(\nfeb\)}
  \end{subfigure}
  \begin{subfigure}[b]{0.48\linewidth}
      \includegraphics[width=\linewidth]{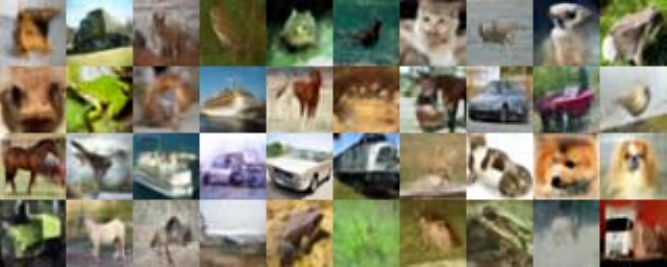}
      \caption{\(\ours\). Para. NFE=\(\nfea\)}
  \end{subfigure}
  \begin{subfigure}[b]{0.48\linewidth}
      \includegraphics[width=\linewidth]{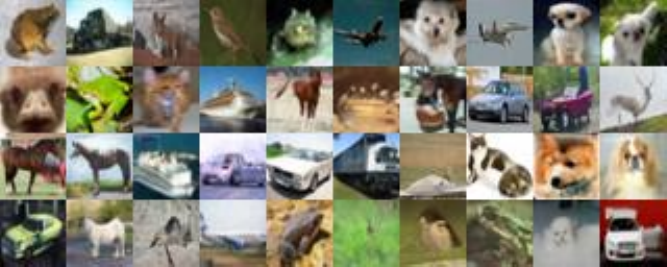}
      \caption{\(\ours\). Para. NFE=\(\nfeb\)}
  \end{subfigure}
  \caption{Qualitative result on CIFAR10 32$\times$32 (\(\nfea\) and \(\nfeb\) NFEs)}
\label{fig:sup_grid_cifar10_3}
\end{figure*}

\begin{figure*}[t]
  \centering
  \begin{subfigure}[b]{0.48\linewidth}
      \includegraphics[width=\linewidth]{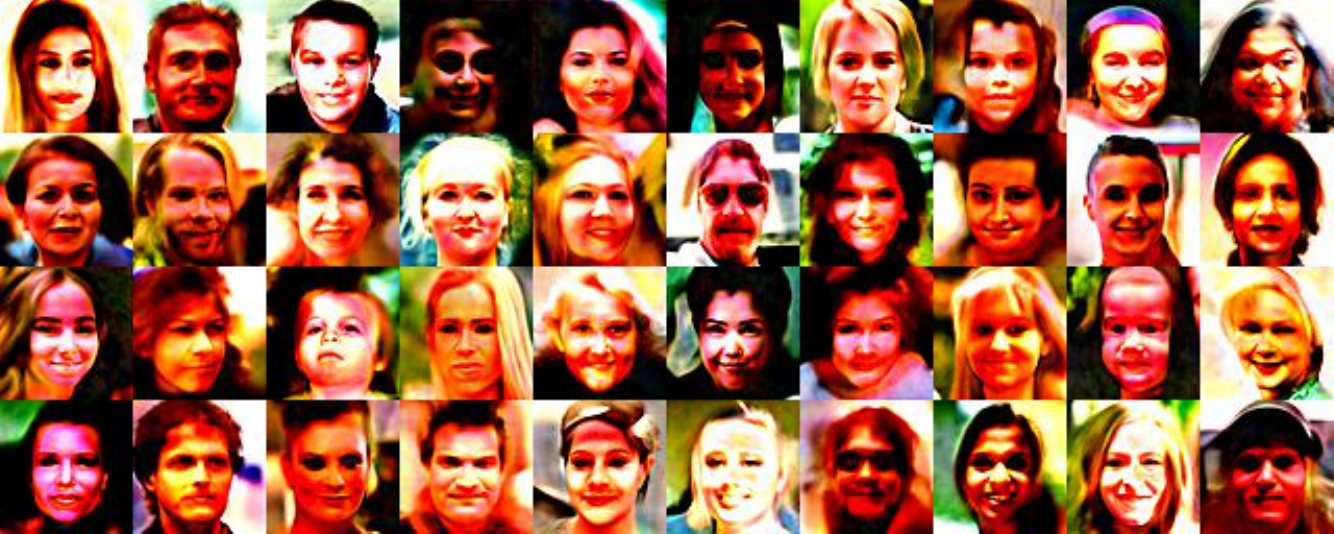}
      \caption{DPM-Solver-2. NFE=\(\nfea\)}
  \end{subfigure}
  \begin{subfigure}[b]{0.48\linewidth}
      \includegraphics[width=\linewidth]{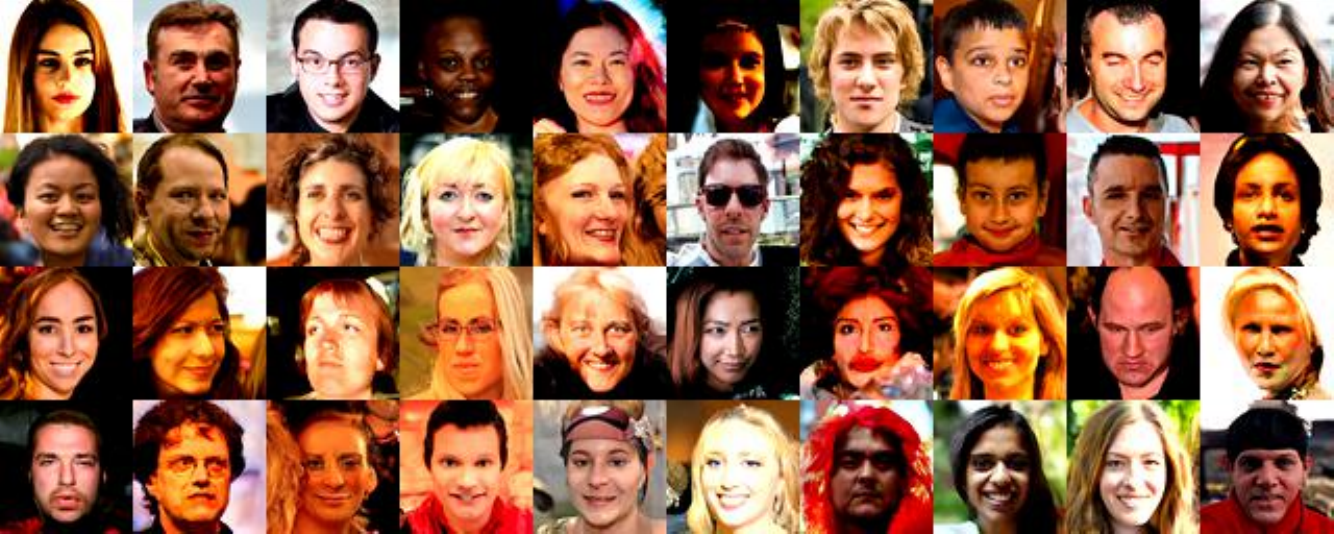}
      \caption{DPM-Solver-2. NFE=\(\nfeb\)}
  \end{subfigure}
  \begin{subfigure}[b]{0.48\linewidth}
      \includegraphics[width=\linewidth]{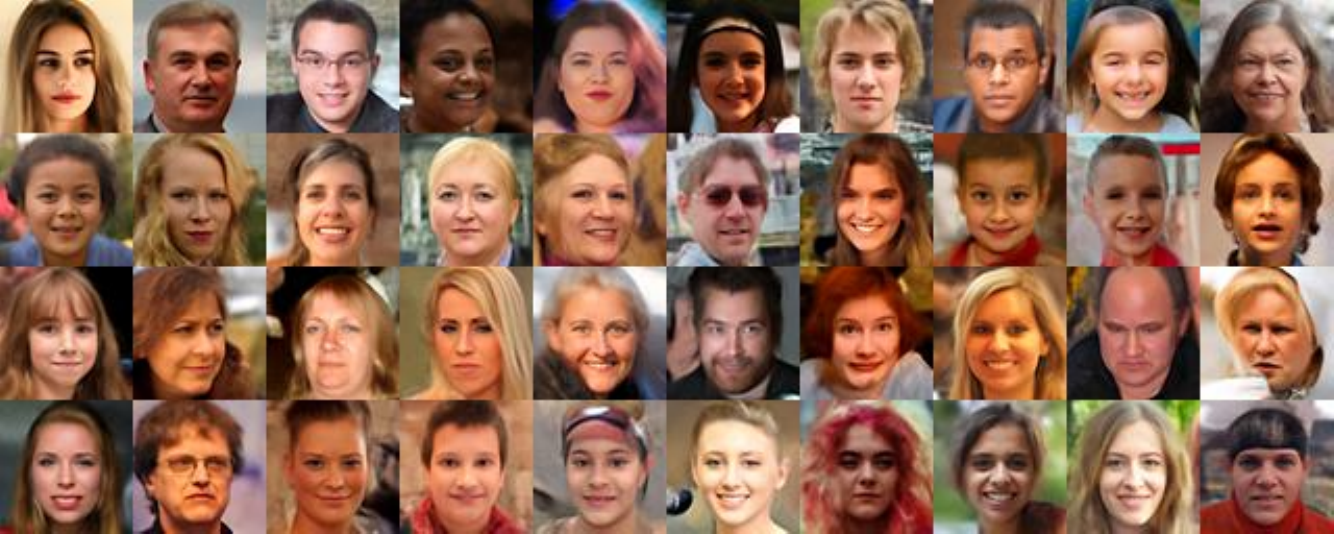}
      \caption{\(\ours\). Para. NFE=\(\nfea\)}
  \end{subfigure}
  \begin{subfigure}[b]{0.48\linewidth}
      \includegraphics[width=\linewidth]{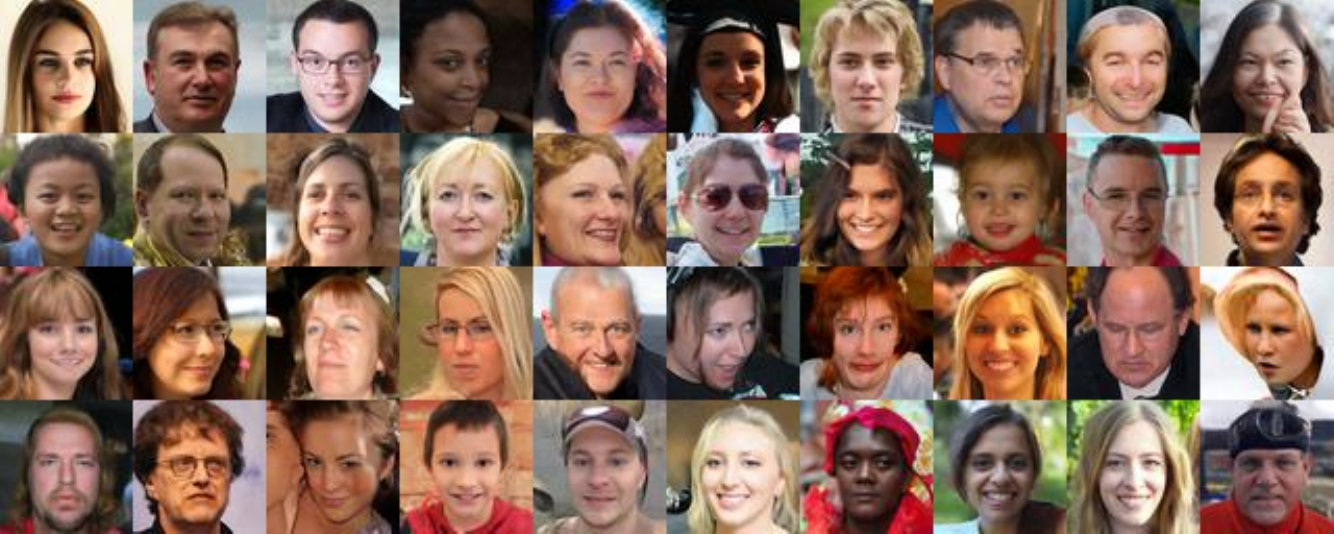}
      \caption{\(\ours\). Para. NFE=\(\nfeb\)}
  \end{subfigure}
  \caption{Qualitative result on FFHQ 64$\times$64 (\(\nfea\) and \(\nfeb\) NFEs)}
  \label{fig:sup_grid_ffhq_3}
\end{figure*}

\begin{figure*}[t]
  \centering
  \begin{subfigure}[b]{0.48\linewidth}
      \includegraphics[width=\linewidth]{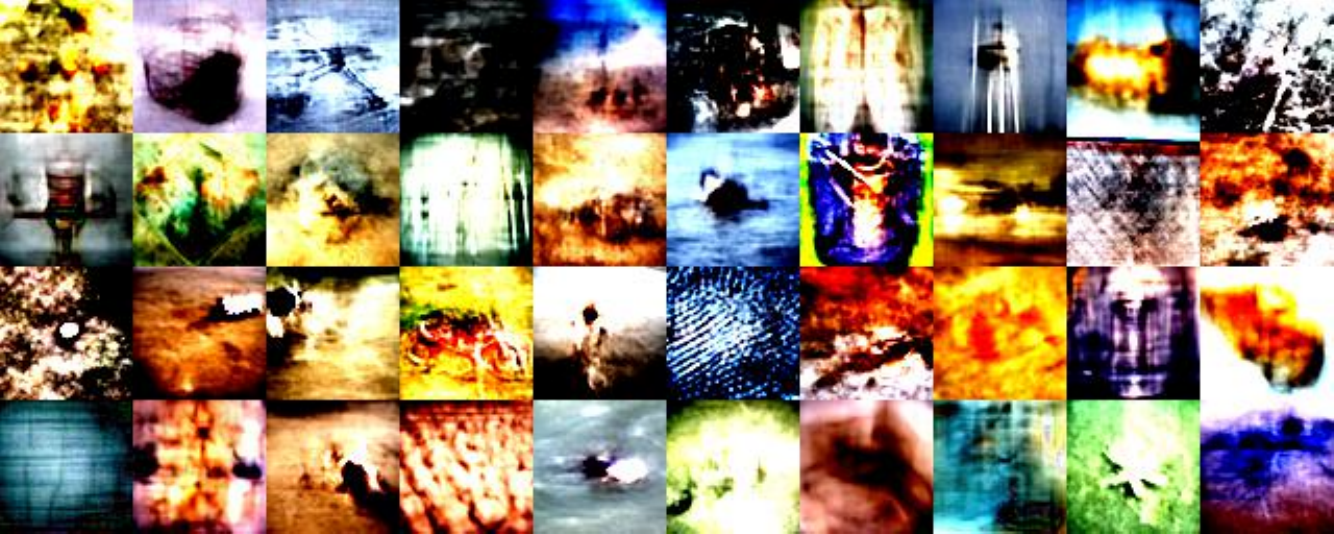}
      \caption{DPM-Solver-2. NFE=\(\nfea\)}
  \end{subfigure}
  \begin{subfigure}[b]{0.48\linewidth}
      \includegraphics[width=\linewidth]{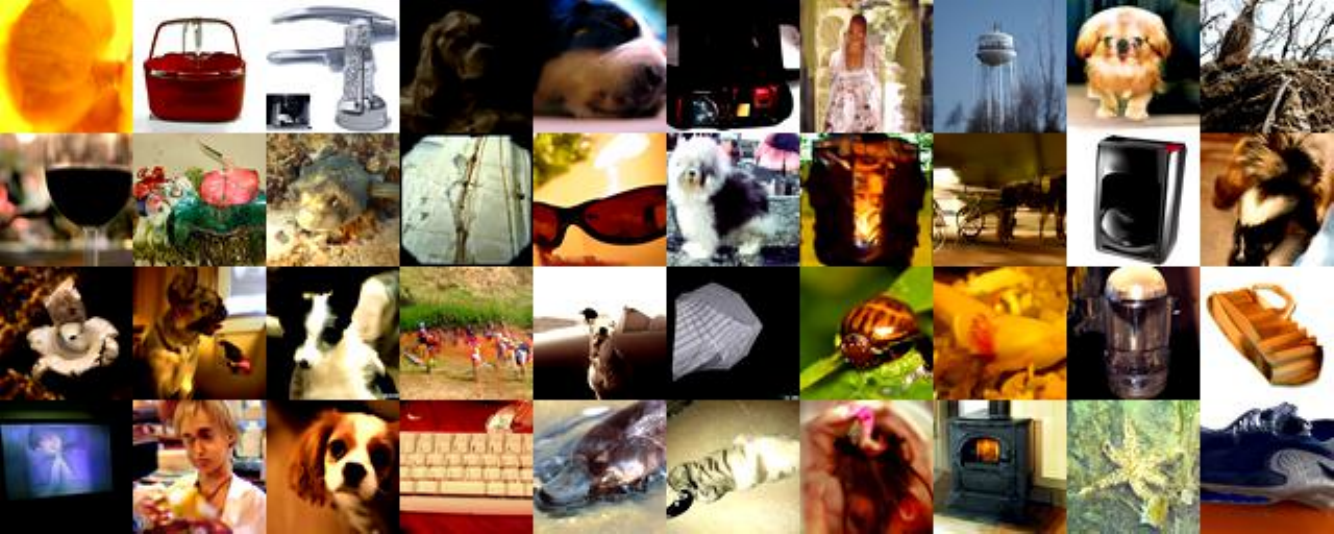}
      \caption{DPM-Solver-2. NFE=\(\nfeb\)}
  \end{subfigure}
  \begin{subfigure}[b]{0.48\linewidth}
      \includegraphics[width=\linewidth]{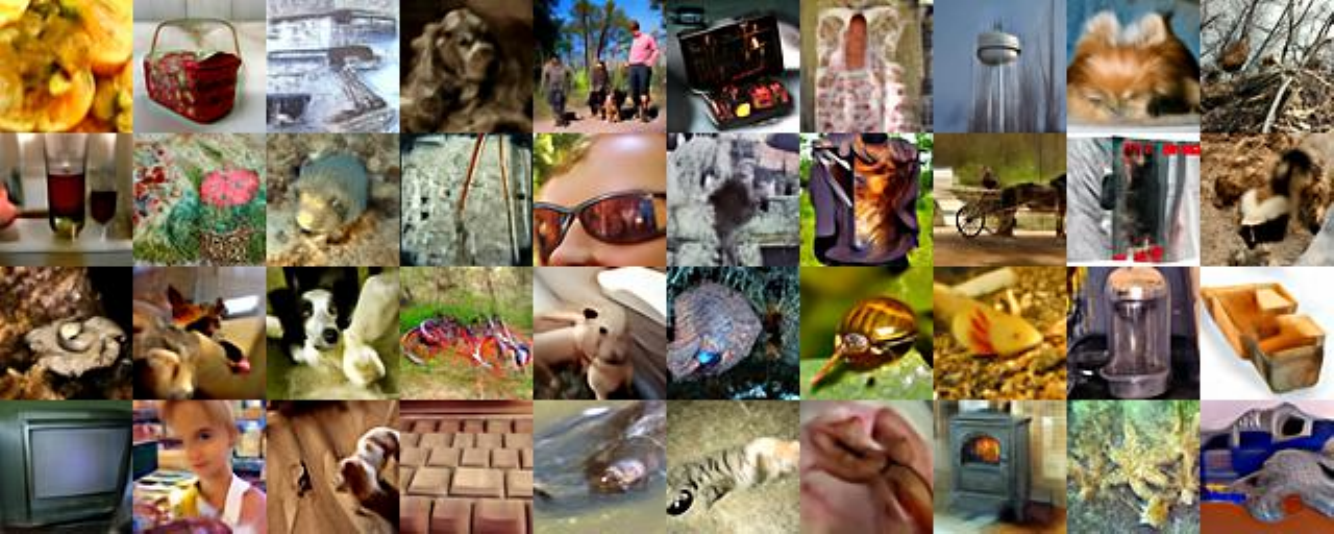}
      \caption{\(\ours\). Para. NFE=\(\nfea\)}
  \end{subfigure}
  \begin{subfigure}[b]{0.48\linewidth}
      \includegraphics[width=\linewidth]{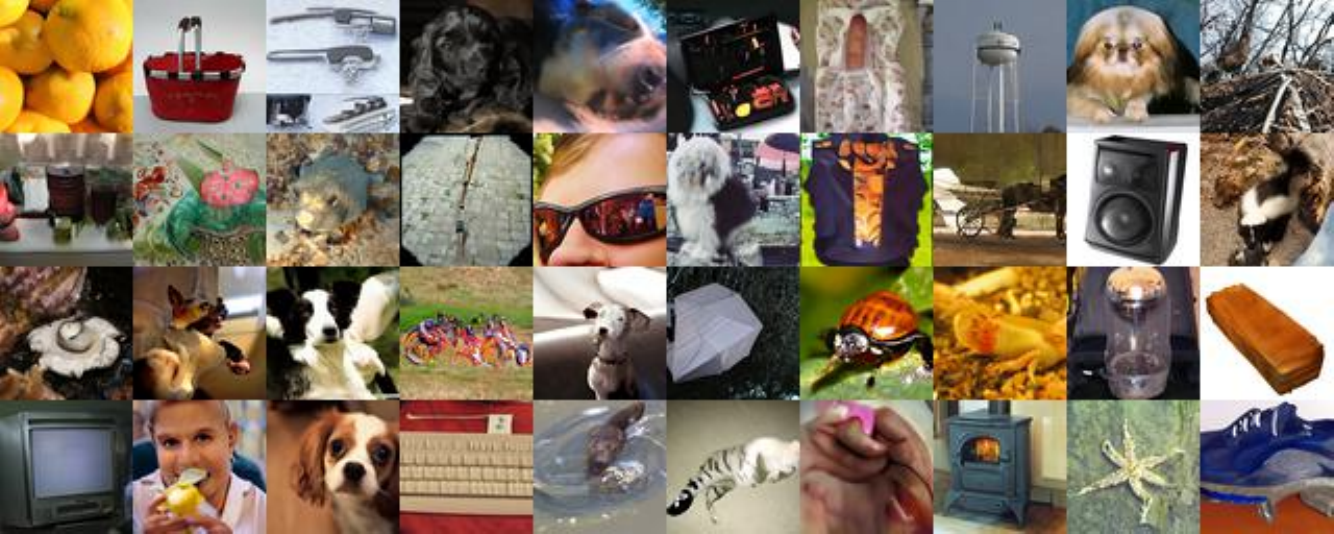}
      \caption{\(\ours\). Para. NFE=\(\nfeb\)}
  \end{subfigure}
  \caption{Qualitative result on ImageNet 64$\times$64 (\(\nfea\) and \(\nfeb\) NFEs)}
  \label{fig:sup_grid_img_3}
\end{figure*}


\vfill

\end{document}